# Domain Adaptation for Statistical Machine Translation

by

## Longyue Wang, Vincent

## Master of Science in Software Engineering

## 2014

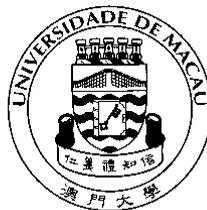

## Faculty of Science and Technology
## University of Macau

Domain Adaptation for Statistical Machine Translation

by

Longyue Wang, Vincent

A thesis submitted in partial fulfillment of the
requirements for the degree of

Master of Science in Software Engineering

Faculty of Science and Technology
University of Macau

2014

Approved by ___________________________________________
Supervisor

___________________________________________

___________________________________________

___________________________________________

Date ___________________________________________

In presenting this thesis in partial fulfillment of the requirements for a Master's degree at the University of Macau, I agree that the Library and the Faculty of Science and Technology shall make its copies freely available for inspection. However, reproduction of this thesis for any purposes or by any means shall not be allowed without my written permission.   Authorization is sought by contacting the author at

Address: S2-4021, Postgraduate Houses, Student Resoureces and Services Section, University of Macau, Av. Padre Tomás Pereira, Taipa, Macau S.A.R., China.

Telephone: +853 83978051
Fax: +853 28838314
E-mail: vincentwang0229@hotmail.com

Signature _______________________

Date ___________________________

University of Macau

Abstract

Domain Adaptation for Statistical Machine Translation

by Longyue Wang, Vincent

Thesis Supervisor: Dr. Sam Chao and Dr. Fai Wong
Master of Science in Software Engineering

Statistical machine translation (SMT) systems perform poorly when it is applied to new target domains. Our goal is to explore domain adaptation approaches and techniques for improving the translation quality of domain-specific SMT systems. However, translating texts from a specific domain (e.g., medicine) is full of challenges. The first challenge is ambiguity. Words or phrases contain different meanings in different contexts. The second one is language style due to the fact that texts from different genres are always presented in different syntax, length and structural organization. The third one is the out-of-vocabulary words (OOVs) problem. In-domain training data are often scarce with low terminology coverage. In this thesis, we explore the state-of-the-art domain adaptation approaches and propose effective solutions to address those problems.

We explore intelligent data selection approaches to optimize models by selecting relevant data from general-domain corpora. As fine-grained selection model has a higher ability of filtering out irrelevant data, we propose a string-difference metric as a new selection criterion. Based on this, we further explore two different approaches, at data level and model level, to combine different type of individual sources to optimize the targeted SMT models. Besides, we deeply analyze their impacts on domain-specific translation quality. We anticipate these approaches can address the ambiguity problem by transferring the data distribution of training corpora to target domain.

In order to make models better learn the language style of sentences, we propose linguistically-augmented data selection approach to enhance perplexity-based models. This method considers various linguistic information, such as part-of-speech (POS), named entity, and so forth, instead of the surface forms. Additionally, we present two methods to combine the different types of linguistic knowledge.

In order to reduce the OOVs, we acquire additional resources to supplement in-domain training data. We apply domain-focused web-crawling methods to obtain in-domain monolingual and parallel sentences from the Internet. To further reduce the irrelevant data, we explore two domain filtering methods for this task. As crawled corpora are usually comparable, we also present an approach to improve the quality of cross-language document alignment.

To prove the robustness and language-independence of our presented methods, all the experiments were conducted on large and multi-lingual corpora. The results show a significant improvement by employing these approaches for SMT domain adaptation. Finally, we develop a domain-specific on-line SMT system named BenTu, which integrates with many useful natural language processing (NLP) toolkits and pipeline the pre-processing, hypotheses decoding and post-processing with an effective multi-tier framework.

# TABLE OF CONTENTS

# LIST OF TABLES

# LIST OF GLOSSARY

**Alignment**: Mapping the text segments (i.e. words, sentences, paragraphs) of a parallel text onto each other.

**Ambiguity**. An attribute of any concept, idea, statement or claim whose meaning, intention or interpretation cannot be definitively resolved according to a rule or process consisting of a finite number of steps.

**Assimilation**. In machine translation, when used for assimilation purposes, the translation should help the reader in understanding texts originally available in a language it does not read.

**Bilingual Corpus**. A collection of text paired with translation into another language.

**Corpus**. A body of linguistic data, usually naturally occurring data in machine readable form, especially one that has been gathered according to some principled sampling method.

**Data Selection**. Selecting data suitable for the domain at hand from large general-domain corpora, under the assumption that a general corpus is broad enough to contain sentences that are similar to those that occur in the domain.

**Dictionary**. A collection of words and phrases with information about them. Traditional dictionaries contain explanations of spelling, pronunciation, inflection, word class, word origin, word meaning and word use. However, they do not provide much information about the relationship between meaning and use. A dictionary for computational purpose rarely says anything about word origin, and may say nothing about meaning or pronunciation either.

**Dissemination**. In machine translation, when used for dissemination purposes, the output is typically post-processed by a human translator in order to obtain high quality translation to be published.

**Domain Adaptation**. There is a mismatch between the domain for which training data are available and the target domain of a machine translation system. Different domains may vary by topic or text style.

**Domain-Specific Statistical Machine Translation**. An SMT translation system that are designed for translating text from a specific domain.

**General-Domain Corpus**. Data in this corpus are mixed by different domain.

**Gold Standard.** For a given task, the set of correct answers as created by one or more humans doing the task, used as a standard for evaluating the success of a computer system doing the task.

**In-Domain Corpus**. Data in this corpus come from the same domain.

**Language Model**. One essential component of any SMT system. It measures how likely it is that a sequence of words would be uttered by a target-language speaker.

**Language Style**. Texts from different genres or topics are always presented in different syntax, length, structural organization etc.

**Monolingual Corpus**. A collection of text in one (mostly are the target side) language.

**Machine Translation.** The use of the computer with or without human assistance to translate texts between natural languages.

**Out-Of-Vocabulary Words**. Some words occur rarely in knowledge base. As a result, their meanings can sometimes be obscure and can only be partly determined through context.

**Parallel corpora.** Two or more corpora in which one corpus contains data produced by native speakers of a language while the other corpus has that original translated into another language.

**Part-of-Speech.** Any of the basic grammatical classes of words, such as noun, verb, adjective, and preposition.

**Statistical Machine Translation**. Translations are generated on the basis of statistical models whose parameters are derived from the analysis of text corpora.

**Segmentation.** Determination of segment boundaries.

**Terminology**. Words or phrases that mainly occur in specific contexts with specific meanings.

# LIST OF ABBREVIATIONS

**ACL**. Association for Computational Linguistics

**CBMT**. Corpus-Based Machine Translation

**CL**. Computational Linguistics

**CLIR**. Cross-Language Information Retrieval

**EM**. Expectation Maximization

**FAHQMT**. Fully Automatic High Quality Machine Translation

**HMT**. Hybrid Machine Translation

**IWSLT**. International Workshop on Spoken Language Translation

**LM**. language model

**MT**. Machine Translation

**MERT**. Minimum Error Rate Training

**ML**. Machine Learning

**MLE**. Maximum likelihood estimation

**NLP**. natural language processing

**NLP2CT**. Natural Language Processing & Portuguese Chinese Machine Translation Laboratory

**OOVs**. Out-of-Vocabulary Words

**POS**. Part-of-Speech

**RBMT**. Rule-Based Machine Translation

**SMT**. Statistical Machine Translation

**TM**. Translation Model

**VSM**. Vector Space Model

# PREFACE

"Both effort and passion are significant to research career." It is not only a motto told by my supervisor Prof. Wong on my first day in NLP$^2$CT Laboratory (Natural Language Processing & Portuguese-Chinese Machine Translation), but also a summary of my life in the past three years: taking more and useful courses, leading teams for various shared tasks, attending research exchange programs and conferences, focusing on reading and writing papers etc.

As I am interested in natural language processing (NLP), I have explored broad fields of NLP such as Chinese word segmentation, named entity, cross-language information retrieval, grammatical error correction, web crawling as well as machine translation. These work involve various nature languages such as Chinese, Japanese, English, French, Czech, Portuguese, German and Spanish etc. Working in these various but closely related fields not only lays a solid foundation for my deeper research, but also inspires me to propose innovative approaches. Besides, statistical machine translation (SMT) domain adaptation is my on-going and key work during master's degree.

I like to attend evaluation tasks, because it is a good way to balance the theoretical approaches and practical applications. Therefore, I have actively attended a lot of evaluation tasks such as SIGHAN2012 Bake-offs, CoNLL-2013, CoNLL-2014, CWMT2013 as well as WMT2014. Though fighting with other outstanding teams in same filed, I not only gain a lot of experience on solving the real-life problems in "big data" environment, but also enhance my abilities of teamwork, leadership, and management in dealing with a large projects and systems. All of these achievements give me confidence to further my research.

Communication with other researchers makes me progress a lot. I found my best ideas are always generated during some discussions; an useful suggestion is often given by kind professors; a good result of campaign benefits from a nice team working. That is why I often attend academic conferences and exchange programs. Though this kind of events, I have cultivated a growing network of research friends from DCU, CMU, PekingU, TuebingenU etc. We often exchange our new ideas, latest work, challenged problems and recommended publications via emails.

I write this thesis to report my work on machine translation in the past three years. I hope it is valuable for others in this field.

Macau, November 11, 2014.

# ACKNOWLEDGEMENTS

I would like to thank all the people, whoever helped me, guided me and supported me. All of you are my gifts in my life. It is because of you, I can walk from my small hometown to this excellent university.

My parents are common workers, but they always try their utmost to give me the best education. They not only bring me up, but also teach me how to be a gentleman. Miss Yang is my primary school teacher, when I was nine-years-old. She is my lifelong teacher. Every time I lost my way, she guided me to make the best decision. Professor Wong and Chao are my supervisors during master degree. They brought me to the wonderful world of natural language processing and machine translation. I would never forget the happy time with the group members in NLP$^2$CT Laboratory.

There is an old saying in China: behind an outstanding gentleman, there must be a nice girl, who always gives him the endless supporting, help, encouragement and kindness. I was so lucky that an angel loved me so much in the past three years. No matter what happened, she was always with me through her laughter and tears. At the end of my master degree time, she made me grew up a lot. I not only have a better understanding on my research work, but also know the most treasure of my life: sincerity, love and responsibility. All the wonderful memories with her are the ideological motivating power to make me keep on fighting, fighting and fighting in my left life. I know this thesis is my last love letter to her: Siyou, you are the best gift in my life, if only…

The goal of our research is communication without boundary. To achieve this wonderful and difficult goal, we struggle all day and night. However, if in the pursuit of the big goal, friends become strangers, hearts become stones, doors become walls, then, all we have done are meaningless. Computers can do everything for us, but cannot give us love and sincerity.

# CHAPTER 1: INTRODUCTION

*When you ask: 'what can we do to improve the quality of the machine translation engine?' I say: 'To many in MT, "more data is better data"; on the whole, that is true.'*

- Tom Hoar, March 2009

Statistical machine translation (SMT) (Brown et al., 1993) is data-driven. SMT systems are trained on a large amount of translated texts. Thus, the performance of SMT systems depends heavily upon the quality and quantity of available training data. However, the point "more data is better data" is not always true when considering the relevance between training data and what we want to translate (test data). That is why SMT systems often perform poorly when applied to new target domains (domain shifting). On the other hand, a translation system trained on the relative domain data could surprisingly perform much better than on a larger amount of irrelevant training data. Thus, domain-specificity of training data with respect to the test data is a significant factor that we cannot ignore.

Here is an interesting example. When we use Google Translator[1] to translate a term "You-Know-Who (is a name of fictional character in J. K. Rowling's Harry Potter series)" from English into Chinese, it is always literally translated as "你知道是谁 (you know who he is)" [2]. Even Google Translator learns from the largest data from the Internet, it still cannot work well when translating a name from fiction domain.

Actually, most real-life SMT challenges are domain specific in nature, but domain specific training data are often sparse or completely unavailable. In such scenarios, domain adaptation techniques are employed to improve domain-specific translation quality by leveraging general-domain data. This thesis explores different approaches to improve translation quality for a given target domain, focusing on adapting translation and language models.

---

[1] https://translate.google.com/.

[2] It should be appropriately translated as "伏地魔 (lord voldemort)" or "神秘人 (mystery man)".

## 1.1 STATISTICAL MACHINE TRANSLATION

Actually, our work relates to many fundamental subjects of computer science, such as NLP, machine learning (ML), machine translation, cross-language information retrieval (CLIR), and computational linguistics (CL). In this section, we will only focus on the definitions, algorithms, techniques etc., which are most related to this thesis topic.

First of all, we give an overview of machine translation, especially for statistical one. As statistical models are built on corpora, we also describe kinds of corpora which are used in our experiments. Then word-based and phrase-based statistical translation models are described in details. Language modeling is a key research direction, which is able to not only smooth SMT outputs, but also measure the information relativity of texts. Thus, we describe it from both SMT and data selection perspectives. Finally, we give details on several evaluation metrics, which are related to our experiments. I hope these background and related knowledge will be helpful to understand my research works in the following contents.

### 1.1.1 OVERVIEW OF MACHINE TRANSLATION

Since machine translation (MT) started in the 1950s, many approaches were explored. Early systems were rule-based machine translation (RBMT), which used large bilingual dictionaries and hand-coded rules for fixing the word order in the final output. It is generated on the basis of morphological, syntactic, and semantic analysis of both the source and the target languages. Starting in the late 1980s, as computational power increased and became less expensive, more interest began to be shown in statistical models for machine translation. Data-driven approaches based on corpora (corpus-based machine translation, CBMT) such as example-based MT and SMT started to be developed. Besides, hybrid machine translation (HMT) is proposed to leverage the strengths of statistical and rule-based translation methodologies. The comparison of different MT approaches is shown in Table 1-1 (Gough and Way, 2004; Way and Gough, 2005; Tian et al., 2013).

Table 1-1: Comparison of Different MT Approaches

| MT Approaches | Periods | Advantages | Disadvantages |
|---|---|---|---|
| **Rule-based MT** | 1950's - 1980's | Deep analysis of linguistic, syntax and semantic information | Endless grammatical rules are required |
| **Example-based MT** | 1990's | Correspondences can be found from raw data, well-structured output | Lack of well aligned bitexts, domain dependent |
| **Statistics-based MT** | Current Trend | Learn from corpus, no manual rules | Domain dependent and sensitive, often effective with large and broad data |
| **Hybrid MT** | Current Trend | Rules post-processed by statistics or statistics guided by rules | More complex and hard to balance, need both statistics and linguistics knowledge |

Until now, researchers have explored the various kinds of SMT systems. According to their natures, we give a general category as shown in Figure 1-1.

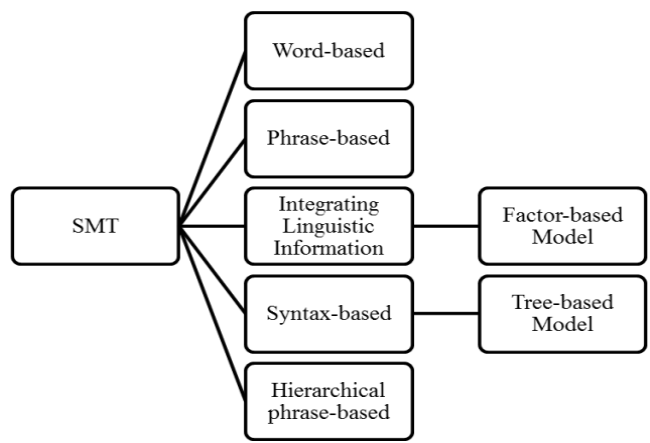

Figure 1-1: SMT Category

In **word-based translation**, the initial statistical models are based on words as atomic units. IBM proposes models (Brown et al., 1988; Brown et al., 1990; Brown et al., 1993) of increasing complexity, that not only take lexical translation into account, but also model recording as well as insertion, deletion, and duplication of words. Besides, a translation model and a language model are combined, under the framework of noisy-channel model (Shannon, 1949), to ensure the fluency of output sentence. In order to reduce the restrictions of word-based translation, the **phrase-based translation** was rooted in work by Och and Weber (1998); Och et al. (1999); Och (2002); Och and Ney (2004) on alignment template models. Translating with the use of phrases in a statistical framework was also proposed by Melamed (1997); Wang and Waibel (1998); Venugopal et al. (2003); Watanabe et al. (2003). Marcu (2001), which propose that the input sentence is broken up into a sequence of phrase (any contiguous of words, not necessarily linguistic entities); these phrases are mapped one-to-one to output phrases, which may be reordered. In addition, a log-linear framework is proposed (Och and Ney, 2002) to integrate different components such as language model, phrase translation model, lexical translation model, or reordering model as feature functions with appropriate weights. An influential description is presented by Koehn et al. (2003), they suggest the use of overlapping phrases. Lopez and Resnik (2006) study the contribution of the different components of a phrase-based model. Some integrates linguistic information such as lemma, part-of-speech (POS) and morphology. **Factored translation models** (Koehn and Hoang, 2007) are an extension of phrase-based models and it allows the integration of syntactic features into the translation and reordering models. Each word is represented as a vector of factors, instead of the simple word surface. Then a phrase mapping is decomposed into several steps that either translate input factors into output factors or transform one factors representation to another. Since modern linguistic theories use tree structure to represent the sentence, another SMT approach is **syntax-based** (Quirk and Menezes, 2006; Knight, 2007). Syntax based models are based on the recursive structure of language, often with the use of synchronous context free grammars. These models may or may not make use of explicit linguistic annotations such as phrase structure constituency labels or labeled syntactic dependencies. **Hierarchical phrase-based translation** (Chiang et al., 2005) combines the strengths of phrase-based and syntax-based translation. It uses phrases (segments or blocks of words) as units for translation and uses synchronous context-free grammars as rules (syntax-based translation). Currently, the most successful approach to SMT machine translation is phrase-based models, thus in this work, we focus on this kind of models.

### 1.1.2 Corpora for Statistical Models

SMT systems are trained by mining the statistical information mainly from the corpora. Generally, there are three kinds of corpora that serve SMT: **parallel corpus**, which is a collection of text paired with translation into another language; **monolingual corpus**, which is a collection of text in one (mostly are the target side) language; **comparable corpus**, which is a collection of similar texts in more than one language or variety but not aligned. All of them are valuable resources for SMT task, which can be used to collect enough statistical evidences for SMT parameter estimation.

In order to make definitions more clear, we take the 2012 International Workshop on Spoken Language Translation (IWSLT2012) English-Chinese training corpus (first six sentence pairs) as an example. As shown in Figure 1-2, each sentence in English and Chinese are paired, thus they are parallel corpus, which can be used to train the translation models. When we only consider the Chinese side in an English-to-Chinese MT system, the Chinese texts are monolingual corpus, which can be used to train the language models. If these two texts are not one-to-one aligned or even just same in meaning, we call them comparable corpus, which can be used to enhance the SMT models only after processing such as alignment, dictionary extraction etc.

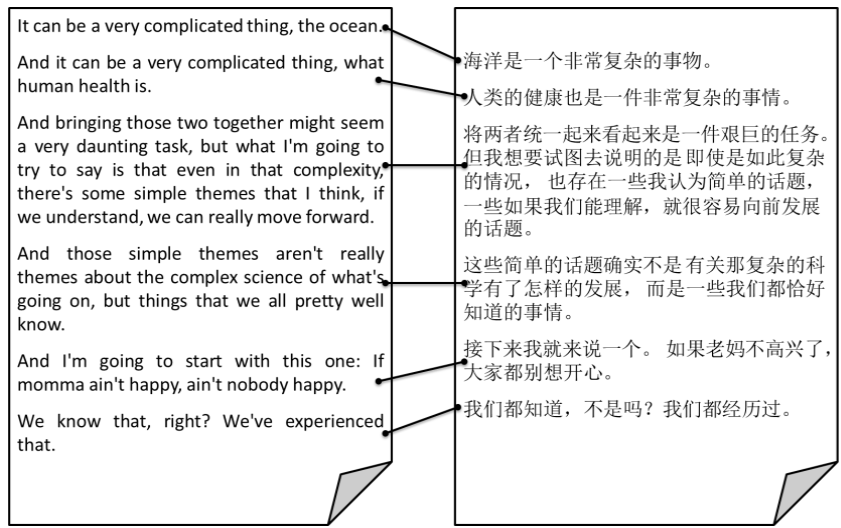

Figure 1-2: Sample of IWSLT2012 Training Corpus

Training data are often manually or automatically collected from different sources. For instance, the IWSLT Corpus is collected from multilingual transcriptions of TED talks, while the European Parliament Corpus is extracted from the conference proceedings of European Parliament. It results in the fact that IWSLT may include a lot of colloquial contents and the European Parliament Corpus would cover many political, economic, and cultural matters. Therefore, corpora can be distinguished according to different styles, topics and genres etc. In Chapter 2, we will use the term "domain" to classify difference kinds of corpora. Furthermore, SMT systems trained on these corpora will also keep the characteristics due to performance basis to some specific domains. In Chapter 3, we will give details on how to best measure domain of sentences and select useful sentences from a larger general-domain corpus to enhance SMT systems.

About collecting corpus for the purpose of SMT, manual work seems impossible due to its high cost. Thus, automatically collecting larger corpora from the Internet is the mainstream work. It consists of techniques such as web crawling, text extraction, formatting, filtering etc. Generally obtaining monolingual and comparable corpora are relatively easy. But if you would like to acquire parallel corpus from Internet, it is more complex and needs more techniques such as cross-lingual documents/sentences alignment. In Chapter 4, it will present detailed methods on how to acquire domain-specific dictionary, monolingual corpus and parallel corpus to enhance SMT systems.

### 1.1.3 WORD-BASED TRANSLATION MODEL

IBM provided models (IBM model 1-5) of increasing complexity to improve the translation quality. The advances of them are:

- IBM Model 1: lexical translation;
- IBM Model 2: adds absolute alignment model;
- IBM Model 3: adds fertility model;
- IBM Model 4: adds relative alignment model;
- IBM Model 5: fixes deficiency.

Although none of the currently competitive machine translation systems are word based models, the principles such as generative modeling and the use of the expectation maximization algorithm are still core methods today. All other IBM models are developed based on IBM Model 1, thus we will discuss the basic and typical one in the rest of this section.

IBM Model 1 is defined by lexical translation probabilities and the notion of alignment. It applies generative modeling, which can generate a number of different translations for a sentence, each with a different probability. We define the translation probability for a foreign sentence $\mathbf{f}=(f_1,\dots,f_{l_f})$ of length $l_f$ to an English sentence $\mathbf{e}=(e_1,\dots,e_{l_e})$ of length $l_e$ with an alignment of each English word $e_j$ to a foreign word $f_i$ according to the alignment function $a: j \rightarrow i$ as follows:

$$p(\mathbf{e},a \mid f) = \frac{\varepsilon}{(l_f+1)^{l_e}} \prod_{j=1}^{l_e} t(e_j \mid f_{a(j)}) \qquad (1\text{-}1)$$

The core is a product over the lexical translation probabilities for all $l_e$ generated output words $e_j$. The fraction before the product is necessary for normalization. Since we include the special NULL token, there are actually $l_f+1$ input words. Hence, there are $(l_f+1)^{l_e}$ different alignments that map $l_f+1$ input words into $l_e$ output words. The parameter $\varepsilon$ is a normalization constant, so that $p(e,a|f)$ is a proper probability distribution, meaning that the probabilities of all possible English translations $e$ and alignments $a$ sum up to one: $\sum_{e,a} p(e,a \mid f) = 1$.

There is a problem in above method: we lack the alignment function $a$, because the training data are only sentence aligned. How to learn from incomplete data is a typical problem in machine learning. Regarding the alignment as a hidden variable in this model, they proposed expectation maximization algorithm (EM) to address the situation of incomplete data. The EM algorithm works as follows (Table 1-2):

Table 1-2: Expectation Maximization Algorithm

| **Algorithm:** EM Steps |
| --- |
| 1. Initialize the model, typically with uniform distribution; |
| 2. Apply the model to the data (expectation step); |
| 3. Learn the model from the data (maximization step); |
| 4. Iterate steps 2 and 3 until convergence. |

First, we initialize the model. Without prior knowledge, uniform probability distributions are a good starting point. For our case of lexical translation this means that each input word $f$ may be translated with equal probability into any output word $e$. Another option would be to start with randomized translation probabilities. In the expectation step, we apply the model to the data. We fill in the gaps in our data with the most likely values. In our case, what is missing is the alignment between words. Therefore, we need to find the most likely alignments. Initially, all alignments are equally likely, but further along, we will prefer alignments where, e.g., the German word *Haus* is aligned to its most likely translation *house*. In the maximization step, we learn the model from the data. The data are now augmented with guesses for the gaps. We may simply consider the best guess according to our model, but it is better to consider all possible guesses and weight them with their corresponding probabilities. Sometimes it is not possible to efficiently compute all possible guesses, so we have to resort to sampling. We learn the model with maximum likelihood estimation, using partial counts collected from the weighted alternatives. We iterate through the two steps until convergence.

### 1.1.4 PHRASED-BASED TRANSLATION MODEL

Words may not be the best candidates for the smallest units for translation. Sometimes, one word in a foreign language translates into two English words, or vice versa. In phrase-based models, the input sentence (source side) is firstly split into phrases. Then, each phrase is translated into an English phrase. Finally, phrases may be reordered. The advantages of this approach are: 1, words may not be the best atomic units for translation, due to frequent one-to-many mappings (and vice versa); 2, translating word groups instead of single words helps to resolve translation ambiguities. 3, if we have large training corpora, we can learn longer and longer useful phrases, sometimes even memorize the translation of entire sentences. 4, the model is conceptually much simpler. We do away with the complex notions of fertility, insertion and deletion of the word-based model.

After applying the Bayes rule to invert the translation direction and integrate a language model. The phrase-based SMT model can be formally described as follows:

$$e_{best} = \arg\max_e \prod_{i=1}^{I} \phi(\bar{f}_i \mid \bar{e}_i) d(start_i - end_{i-1} - 1) \prod_{i=1}^{|e|} P_{LM}(e_i \mid e_1...e_{i-1}) \qquad (1\text{-}2)$$

It consists of three components: the phrase translation table $\phi(\bar{f}_i \mid \bar{e}_i)$, which ensure the foreign phrase to match English words; reordering model $d$, which reorder the phrases appropriately; and language model $P_{LM}(e)$, which ensure the output to be fluent.

For two reasons: 1, the weighting of the different model components may lead to improvement in translation quality; 2, log-linear model structure allows to include additional model components in the form of feature functions, they apply log-linear model structure. The Equation 1-3 can be formed as follows:

$$p(x) = \exp \sum_{i=1}^{n} \lambda_i h_i(x) \qquad (1\text{-}3)$$

- number of feature function $n$=3;
- random variable $x$=($e$, $f$, $start$, $end$);

- feature function $h_1 = \log \phi$;

- feature function $h_2 = \log d$;

- feature function $h_3 = \log P_{LM}$;

Besides, the state-of-the-art SMT toolkit, Moses[3] uses more feature functions such as direct phrase translation probability, inverse phrase translation probability, direct lexical weighting, inverse lexical weighting, phrase penalty, language model, distance penalty, word penalty, distortion weights et al. Feature weights are tuned on development set by Minimum Error Rate Training (MERT) (Och, 2003), using BLEU (detailed in Section 1.1.6) as the objective function.

A general architecture of phrase-based SMT system is illustrated in Figure 1-3. It shows two parts: the training part (static) and translation part (dynamic). Training is the process to build a number of models by mining the statistical information from the training corpora. Translation is also called decoding, which find the best scoring translation from exponential number of candidates given the input sentences. The collections of sentences are called **corpora** (detailed in Section 1.1.2), and for statistical machine translation we are especially interested in parallel corpora, which are texts, paired with a translation in another language. **Translation models** are trained using parallel corpora which provide hidden alignment information. It regards the translation between a sentence pair as a mapping of the words or phrases on either side. **Language models** (detailed in Section 1.1.5) are trained on monolingual corpora in target language, which ensure the fluency of the output and are an essential part of SMT. They influence word choice, reordering and other decisions.

---

[3] Available at http://www.statmt.org/moses/.

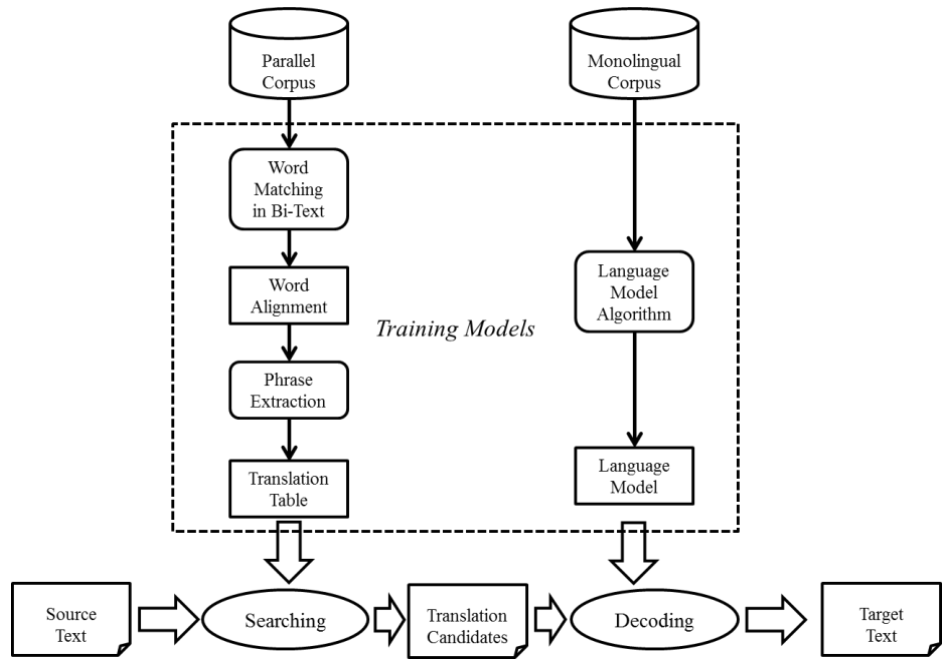

Figure 1-3: SMT Framework

### 1.1.5 LANGUAGE MODEL

One essential component of any SMT system is the language model, which measures how likely it is that a sequence of words would be uttered by a target-language speaker. It is able to not only ensure the system generate fluent sentence, but also aids translation on word choice.

Take a simple sentence *I am going home* for example. It is obviously more native than *home am I going* for an English speaker. Thus, the language model $p_{LM}$ should prefer correct word order to incorrect one as shown in Eq. (1-4).

$$p_{LM}\ (I\ am\ going\ home) > p_{LM}\ (home\ am\ I\ going) \tag{1-4}$$

Besides, a foreign word (for example, the Chinese word 家) has multiple translations such as *house*, *home*, *family*. The lexical translation probability may already give preference to the more common translation (*family*). But in special contexts, other translations may be correct. $p_{LM}$ should give higher probability to the more natural word choice in context as shown in Eq. (1-5).

$$p_{\text{LM}} \ (I \ am \ going \ home) > p_{\text{LM}} \ (I \ am \ going \ family) \hspace{2cm} (1\text{-}5)$$

Currently, the leading method for language models is n-gram language modeling, which are based on statistics of how likely words are to follow each other. Under the Markov assumption that only a limited number of previous words affect the probability of the next word, the language model probability is a product of limited word history probabilities. The score of LM for a sentence $s$ can be computed as follows:

$$p_{LM}(s) = \prod_{i=1}^{l+1} p(w_i \mid w_{i-n+1}^{i-1}) \hspace{2cm} (1\text{-}6)$$

$$p(w_i \mid w_{i-n+1}^{i-1}) = \frac{c(w_{i-n+1}^{i})}{\sum_{w_i} c(w_{i-n+1}^{i})} \hspace{2cm} (1\text{-}7)$$

in which $w_i$ is the $i$th word in $s$ and $w_{i-n+1}^{i-1}$ is the a limited number of previous words of $w_i$. Maximum likelihood estimation (MLE) is used to calculate $p(w_i \mid w_{i-n+1}^{i-1})$ Eq. (1-7).

Moreover, both word-based and phrase-based translation models integrate with language model by the way of noisy-channel model (Shannon, 1949). The translation can be derivated by Bayes rule. As shown in Eq. (1-8), the $P(e)$ here is language model $p_{\text{LM}}(e)$.

$$\arg\max_e p(e \mid f) = \arg\max_e \frac{p(f \mid e) p(e)}{p(f)} \hspace{2cm} (1\text{-}8)$$
$$= \arg\max_e p(f \mid e) p(e)$$

### 1.1.6 EVALUATION

According to our experiments, we will talk about three evaluation metrics to measuring the quality of a language model, sentence selection, cross-language document alignment as well as machine translation.

**Perplexity**

In building LMs, we often use perplexity to measure the model's quality. Perplexity is based on the cross-entropy (Eq. (1-9)), which is the average of the negative logarithm of the word probabilities.

$$H(p,q) = -\sum_{i=1}^{n} p(w_i) \log q(w_i)$$
$$= -\frac{1}{N} \sum_{i=1}^{n} \log q(w_i)$$

(1-9)

where $p$ denotes the empirical distribution of the test sample. $p(x) = n/N$ if $x$ appeared $n$ times in the test sample of size $N$. $q(w_i)$ is the probability of event $w_i$ estimated from the training set (same as $p(w_i \mid w_{i-n+1}^{i-1})$ in Eq. (1-7)). Thus, the perplexity $pp$ can be simply transformed as:

$$pp = b^{H(p,q)}$$

(1-10)

where $b$ is the base with respect to which the cross-entropy is measured (e.g., bits or nats). $H(p, q)$ is the cross-entropy given in Equation 1-10, which is often applied as a cosmetic substitute of perplexity for data selection (Moore and Lewis, 2010; Axelrod et al., 2011).

**F Score**

The most frequent and basic evaluation metrics for information retrieval are precision and recall, which are defined as follows (Manning et al., 2008):

$$P = \frac{\#(relevant \quad items \quad retrieved)}{\#(retrieved \quad items)}$$

(1-11)

$$R = \frac{\#(relevant \quad items \quad retrieved)}{\#(relevant \quad items)}$$

(1-12)

To report the qualities of cross-language document alignment (detailed in Chapter 4), we used the $F1$ score, the recall and the precision values. $F1$-measure ($F$) is formulated by Van Rijsbergen as a combination of recall ($R$) and precision ($P$) with an equal weight in the following form:

$$F = \frac{2PR}{P+R} \qquad\qquad (1\text{-}13)$$

**BLEU Score**

The methods to measure the machine translation quality can be divided into manual evaluation and automatic evaluation. In most experiments, we will use automatic evaluation due to the cost; however, we also do some manual evaluation for some others if necessarily.

The currently most popular automatic evaluation metric, the BLEU (Papineni et al., 2002), is to use a test set that has been translated by humans, and to consider an automatic translation to be good if it is similar to the human reference translations. The BLEU metric is defined as:

$$BLEU - n = \min(1, \frac{Len_{out}}{Len_{ref}}) \prod_{i=1}^{n} \lambda_i \log prescision_i \qquad (1\text{-}14)$$

*Len* is the length (number of tokens) of output or reference. It uses a brevity penalty, $\min(1, \frac{Len_{out}}{Len_{ref}})$ to address the problem of "no penalty for dropping brevity penalty words". The penalty reduces the score if the output is too short. The maximum order *n* for n-grams to be matched is typically set to 4. Moreover, the weights $\lambda_i$ for the different precisions are typically set to 1.

The holy grail of MT is "fully automatic high quality machine translation (FAHQMT)". Given the complexity of language and the many unsolved problems in MT, until today, there is still no system even such as Google[4], Bing[5] or BabelFish[6] can reach this goal (Melby, 1995; Wooten, 2006; Bar-Hillel, 1960). However, this goal has been reached for **limited domain** and **controlled language** applications. About limited domain, take a patent domain system, IPTranslator[7] for instance. The set of possible sentences in patent domain is sufficiently constrained that it is possible sentence to write translation rules or techniques that capture all possibilities. About controlled language, we can consider about the user documentation. Multinational companies have to produce documentation for their products in many languages. One way to achieve that is to author the documentation in a constrained version of English that machine translation systems are able to translate.

---

[4] Available at http://translate.google.cn/.

[5] Available at http://www.bing.com/translator/.

[6] Available at http://www.babelfish.com/.

[7] Available at http://iconictranslation.com/.

Besides, the quality of the SMT output should be judged according to the goal of the output. From the perspective of translation goal, the applications of machine translation fall into three categories: 1, **assimilation**, the translation of foreign material for the purpose of understanding the content; 2, **dissemination**, translating text for publication in other language; and 3, **communication**, such as the translation of emails, chat room discussions, and so on (Koehn, 2012).

Due to the complexity of language and the many unsolved problems in machine translation, it is impossible to ask a SMT system to give fully-automatic high-quality translation for everything, but possible in a specific domain. Thus, we think applying domain-specific SMT systems in various in-domains would be a good way to achieve this goal.

## 1.2 PROBLEMS OF DOMAIN-SPECIFIC SMT

Different from typical SMT (in Section 1.1), translating texts from a specific domain is full of challenge due to specialized vocabulary, distinctive genre style and so forth. There are three major problems that reduce the performance of domain-specific SMT: 1) **ambiguity**, the same word in different domains may have disparate meanings so as to differentiate translations; 2) **language style**, texts from different genres are always written in different structure, length, etc.; 3) **OOVs**, in-domain training data are often scarce that results in a number of unknown words during translation.

In monolingual environment, a word may include several meanings according to different genres, topics, styles, national or ethnic origins, dialects, etc., which could be considered as different domains. For example, the word *Galaxy* is a cell phone product in information technology, may not a collection of star system in astronomy. Even more, the intuition of Macau people to *galaxy* should be a name of a local casino. Even though statistical methods can deal with multi-meaning words by mining their context information, it still cannot work well if the training data are sparse. What is worse, multi-meaning may not coincide in bilingual environment. For instance, both English word *Mouse* and German word *Maus* have two meanings: the animal and the electronic device. This can avoid mistranslations because multi-meaning coincides between languages, otherwise it does not. As shown in Figure 1-4, the English word *Mouse* refers to animal and electronic device. However, the pointing electronic device refers to "*鼠标*" or "*滑鼠*" and the muridae animal refers to "*老鼠*" in Chinese. This example also shows the domain-specificity in different national or ethnic origins. The word "*滑鼠*" is often used in Taiwan, while the word "*鼠标*" is used by mainland people. All these situations may cause potential mistranslations. If we translate the sentence "I want to buy a mouse", three translation candidates are generated (as shown in Figure 1-5). However, systems trained on different kinds of corpora will give different preference. Choosing wrong translation variants is a potential cause for miscomprehension.

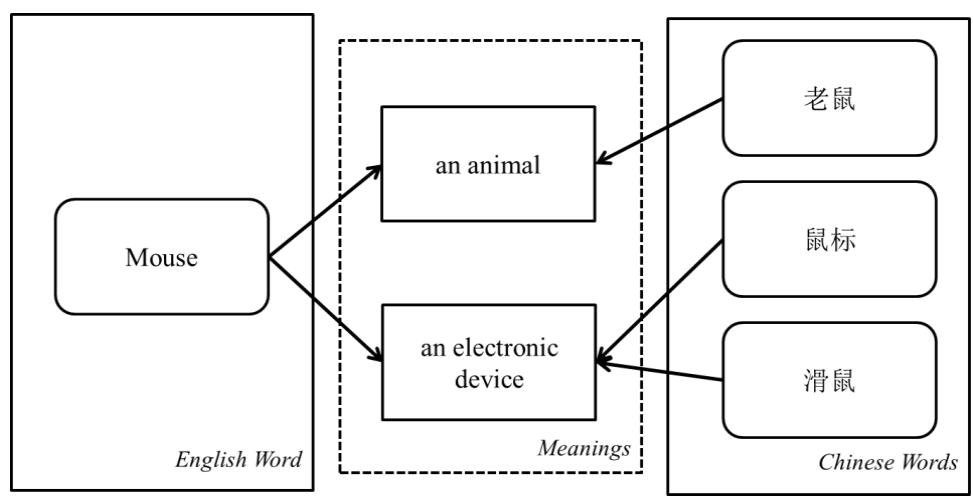

Figure 1-4: Example of Multi-meaning between Languages

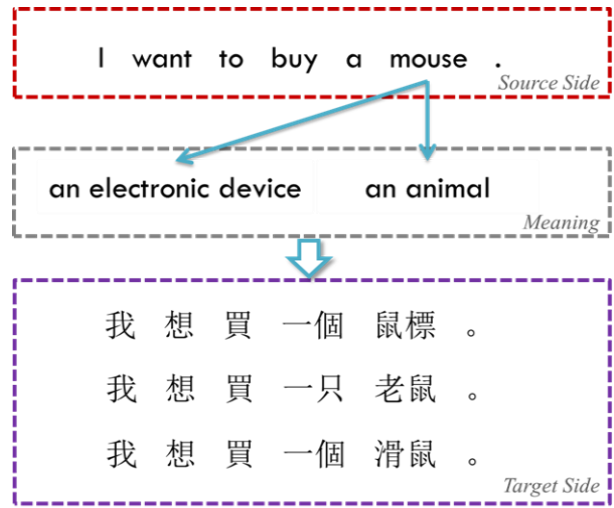

Figure 1-5: Example of Multi-Meaning Translation

Texts in different genres are differed in syntactical structure, parsing prose, adding diction, and organizing figures of thought into usable frameworks. As shown in Figure 1-6, we take texts from legal (upper one) and news (lower one) domain for example. The news article tries to deliver rich information with very economical language, thus structure is so simple that easy to understand. A lot of abbreviation, date, named entitles occur in it. On the contrary, the legal document is longer with duplicated terms. Long subordinate clauses make the structure too complex to understand. It includes high frequency words of shall, may, must, be to, but fewer abbreviations. For translation, it is really a big challenge to capture this high level information. If a news article is translated into legal language style, even all the words are well translated, it is still very funny.

> When an international treaty that relates to a contract and **which** the People's Republic of China has concluded on participated into has provisions of the said treaty shall be applied, but with the exception of clauses to which the People's Republic of China has declared reservation.
> 中华人民共和国缔结或者参加的与合同有关的国际条约同中华人民共和国法律有不同规定的,适用该国际条约的规定。但是,中华人民共和国声明保留的条款除外。

> China's Li Duihong won the women's 25-meter sport pistol Olympic gold with a total of 687.9 points early this morning Beijing time. (Guangming Daily, 1996/07/02)
> 我国女子运动员李对红今天在女子运动手枪决赛中,以687.9环战胜所有对手,并创造新的奥运记录。(《光明日报》1996年7月2日)

Figure 1-6: Example of Language Style in Different Domains

Data-driven SMT learns from a large amount of resources such as monolingual corpus, parallel corpus, and dictionary etc. Fortunately, some official organizations such as the Linguistic Data Consortium (LDC)[8] and European Parliament[9] currently provide rich resources for dozens of languages. Therefore, SMT systems trained on these general-background data could achieve reasonable translation qualities. On the contrary, the training resources in specific domains such as medicine, laws etc. are usually relatively scarce. Take the training data English-French parallel corpora in ACL 2014 Ninth Workshop on Statistical Machine Translation (WMT2014) for example, the sentence number of general-domain corpora is 6.5 times more than that of medical domain corpora. If a medical domain SMT system is trained on such small-scale in-domain corpora, a lot of problems will occur: a) a number of medical terminologies may fail to be translated (OOVs); b) some genre style such as sentence patent will disappear in the output; c) and even worse domain shifting will also result in mistranslations. All above problems will lead to misunderstanding for users.

---

[8]  Available at http://www.ldc.upenn.edu/.

[9]  Available at http://www.statmt.org/europarl/.

## 1.3 RESEARCH OBJECTIVE

This thesis proposes novel approaches to improve domain-specific translation quality focusing on dealing with three problems (discussed in Section 1.2): ambiguity, language style and OOVs.

A multi-meaning word may occur in training corpora with both in-domain and out-of-domain translations. Data selection provides a good way to select sentences that are more close to in-domain but different to out-of-domain. Thus, data distribution of training corpora is able to be moved to target domain. We hope that models trained on the relevant subset could give higher probabilities to correct translations for multi-meaning words.

Similar to ambiguity, we think the language style can also be affected by changing the data distribution. Therefore, we continue to use the principle of data selection to address the problem. To make statistical models better learn structure, sematics, and etc. of a sentence, we mining useful data with the help of linguistic information.

Consider that the Internet is the largest multi-lingual and multi-domain corpus, we prefer mining resources to supplement the limited in-domain data. To effectively and automatically acquire "clean" domain-specific resources from comparable corpora, we will mainly explore filtering irrelevant data from crawled corpus to enhance the existing models.

## 1.4 THESIS LAYOUT AND BRIEF OVERVIEW OF CHAPTERS

In this thesis, we will summarize the related theories, define key concepts, review related work, propose novel approaches, conduct a series of experiments and practice SMT in real-life environment. Thus, the thesis is divided into 6 chapters and it is organized as follows:

*Chapter 1. Introduction*

This chapter gives the background of statistical machine translation, which is the core of our work. It includes the key concepts of corpora, translation models, language models and evaluation metrics. We try to mine some interesting and related knowledge by deeply analyzing, summarizing and comparing each of them. Besides, it introduces three big challenges of domain-specific SMT. Following it, solutions are discussed and highlighted to overcome the mentioned limitations.

*Chapter 2. SMT Domain Adaptation*

By analyzing the linguistic phenomena of corpora in different domains, we give key definitions of domain in SMT. Also, we describe related work of domain adaptation in details. Moreover, an overview of the proposed approaches is given.

*Chapter 3. Intelligent Data Selection for SMT Domain Adaptation*

It presents a new citation to select more relevant sentences for domain-specific SMT. Then a systematic comparison of different data selection approaches is conducted. In addition, linguistic information is further used to improve the perplexity-based data selection methods.

*Chapter 4. Domain-focused Web-Crawling for SMT Domain Adaptation*

In this chapter, we detail the web-crawling method for domain adaptation. We present two methods in both cross-language document alignment part and data filtering part.

*Chapter 5. Domain-Specific SMT Online System*

This chapter concludes the thesis by pointing out the contributions of this research in domain-specific machine translation system. We combine various domain adaptation approaches and detailed techniques for medical domain text translation. Finally, the framework of developed system is given.

*Chapter 6. Conclusion*

It draws the thesis conclusion and outlines some future works.

CHAPTER 2: SMT DOMAIN ADAPTATION

*Domain (software engineering), a field of study that defines a set of common requirements, terminology, and functionality for any software program constructed to solve a problem in that field.*

<div align="right">-- Wikipedia</div>

From the historical perspective on MT, applications were often developed around a number of specific domains. In 1950s, MT systems focused on short sentences in military domain. From 1960s to 1970s, MT systems intended to translate everything like humans. Then in 1980s, systems based on rules in Europe (i.e., Systran[10], Eurotra [11] ), mainly use within European Parliament (domain). Since 1990s, domain-specific MT became a new direction.

In this chapter, we firstly discuss the definition of domain, which is the key point in domain adaptation. By analyzing the phenomena from linguistics and statistics perspectives, we give the definitions in different ways. After reviewing the related work of domain adaptation, we discussed more about data selection and web-crawling.

## 2.1 DEFINITION OF DOMAIN

What is domain? Actually, there is no uniform definition in natural language processing (NLP) or SMT research community yet. From the perspective of linguistics, different genres, topics, styles of language may be considered as different domains.

---

[10] http://www.systranet.com/translate.

[11] http://www.eurotra.eu/.

- **Genre**: A literary genre is a category of literary composition. Genres may be determined by literary technique, tone, content, or even (as in the case of fiction) length[12].

- **Topic/Theme**: In linguistics, the topic, or theme, of a sentence is what is being talked about, and the comment (rheme or focus) is what is being said about the topic. That the information structure of a clause is divided in this way is generally agreed on, but the boundary between topic/theme depends on grammatical theory[13].

- **Language Styles**: for example, the word "*mouse*" refers to "*老鼠*" in animal topic, but to "*鼠標*" in computer topic. Considering the styles of language, Chinese can be divided into simplified Chinese (used by mainland people) and traditional Chinese (used by Taiwan, Macau, Hong Kong and Singapore people). If you translate the *mouse* under the computer topic in Taiwan, it should be "*滑鼠*" instead of "*鼠標*".[14]

- Even in the same language, different national or ethnic origins, dialects could be considered as different domains too. The Chinese word "*橡皮*" refers to "rubber" in British English and "eraser" in American English. In China, all students study British English at school. In a classroom of an American university, a Chinese student who just arrived in the U.S. wants to use the eraser of his American classmate, and asks: "May I use your rubber?" However, rubber is "condom" in American colloquial.

Based on the above discussions, we find domain is really complex because it is affected by various factors. In our work, domain is used to indicate a particular combination of all these factors: genres, topics/themes, language styles, origins/dialects. From the perspective of computing linguistics (as shown in Figure 2-1), factors can be divided into four parts: lexicon, which mainly considers the word sense in difference domains; text type, which indicates the genre, topic and theme; language style, which is nature of a language of itself; and terminology, which includes frequency of specific lexical items.

---

[12] http://en.wikipedia.org/wiki/Genres.

[13] http://en.wikipedia.org/wiki/Topic%E2%80%93comment.

[14] In thesis, we consider mandarin and Cantonese are two different languages due to their different writing methods, words.

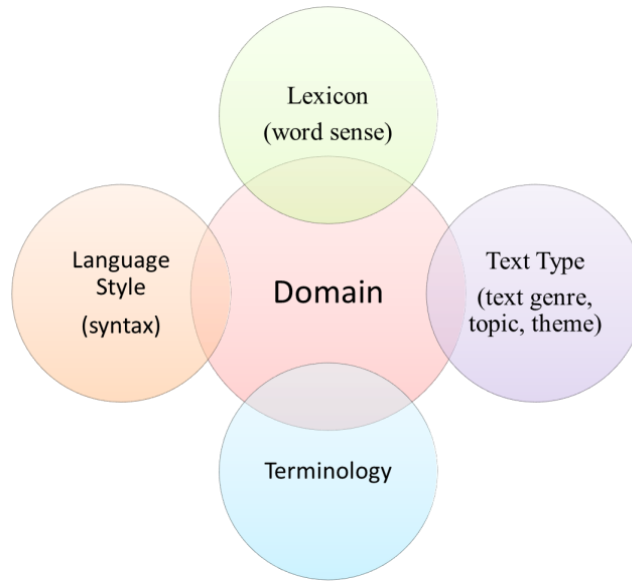

Figure 2-1: Domain Factors

Here, we try to give two definitions from perspectives of translation goal and resources. From the perspective of translation goal, domain is the field of activity. The domain could be defined according to what kind of task the SMT system works for. For instance, if the system is designed for translating patent documents, then the domain is patent, which may include mathematics, chemistry, biology and medicine et al. Although each sub-domain can also be regarded as an isolate domain, we combine them together as a single domain because documents in these sub-domains may occur in this patent translation task. Therefore, this is a task-orientated definition.

Another definition can be generated according to the corpus itself. All corpus-based SMT systems are trained on corpora. As corpora are collected from different genres, each corpus may perform well in some specific domains. For example, the most famous one, European Parliament Proceedings Parallel Corpus come from their conference proceedings. Also the official corpora of International Workshop on Spoken Language Translation (IWSLT) are mainly dialogue. Each corpus can be regarded as a specific domain.

## 2.2 SMT DOMAIN ADAPTATION

The quality of translations provided by statistical machine translation (SMT) systems depend heavily on the quantity of available parallel training data as well as the domain-specificity of the test data with respect to the training data. Most real-life SMT challenges are domain specific in nature, but domain specific parallel training data is often sparse or completely unavailable. Therefore, domain adaptation in SMT becomes an active topic, which is employed to improve domain-specific translation quality by leveraging parallel out-of-domain data. It aims at moving the probability distribution towards the target domain translations in case of ambiguity (Sennrish, 2013) and OOVs.

### 2.2.1 OVERVIEW AND RELATED WORK

Various domain adaptation approaches have been proposed, which can be divided into different kinds of category from different perspectives (as shown in Figure 2-2). Considering supervision, domain adaptation approaches can be decided into supervised, semi-supervised and unsupervised. From the training resources perspective, domain adaptation can use monolingual corpora, parallel corpora, comparable corpora, dictionaries and web-crawled data. Besides, domain adaptation can be employed in different components: word-alignment model, language model, translation model and reordering model. Besides, Wang et al., (2013) classify these approaches according to different component levels: word level such as mining unknown words from comparable corpora (Daumé III and Jagarlamudi, 2011), phrase level such as weighted phrase extraction (Mansour and Ney, 2012), sentence level such as selecting relevant sentences from larger corpus (Moore and Lewis, 2010) and model level such as mixing multiple models (Civera and Juan, 2007; Foster and Kuhn, 2007; Eidelman et al., 2012).

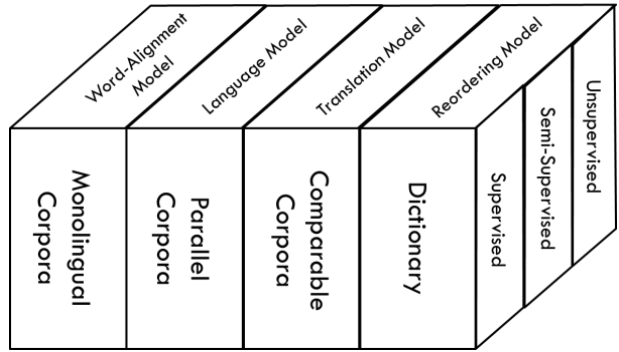

Figure 2-2: Domain Adaptation for SMT

Comprehensively considering the category work of Wang et al., (2013), Sennrich (2013), and Chen et al. (2013), we re-classify the domain adaptation approaches into: mixture models, transductive learning, data selection, instance weighting, and web-crawling. The related works are detailed as follows:

- Research on mixture models has considered both linear and log-linear mixtures. Both were studied in (Foster and Kuhn, 2007), which concluded that the best approach was to combine sub-models of the same type (for instance, several different TMs or several different LMs) linearly, while combining models of different types (for instance, a mixture TM with a mixture LM) log-linearly. (Koehn and Schroeder, 2007), instead, opted for combining the sub-models directly in the SMT log-linear framework.

- In transductive learning, an MT system trained on general domain data is used to translate in-domain monolingual data. The resulting bilingual sentence pairs are then used as additional training data (Ueffing et al., 2007; Chen et al., 2008; Schwenk, 2008; Bertoldi and Federico, 2009).

- Data selection approaches retrieve sentence or sentence pairs that are similar to the in-domain data, and then use them to adapt models into target domain. Researchers explored it by the means of information retrieval (IR) techniques (Zhao et al., 2004; Hildebrand et al., 2005; Lü et al., 2007) and language modeling approaches (Lin et al., 1997; Gao et al., 2002; Moore and Lewis, 2010; Axelrod et al., 2011).

- Instance weighting approaches (Matsoukas et al., 2009; Foster et al., 2010; Huang and Xiang, 2010; Phillips and Brown, 2011; Sennrich, 2012) typically use a rich feature set to decide on weights for the training data, at the sentence or phrase pair level. For example, a sentence from a subcorpus whose domain is far from that of the dev set would typically receive a low weight, but sentences in this subcorpus that appear to be of a general nature might receive higher weights.

- Considering the Web as a parallel corpus, Resnik and Smith (2003) proposed the STRAND system, in which they used Altavista to search for multilingual websites and examined the similarity of the HTML structures of the fetched web pages in order to identify pairs of potentially parallel pages. Similarly, Esplà-Gomis and Forcada (2010) proposed Bitextor, a system that exploits shallow features (file size, text length, tag structure, and list of numbers in a web page) to mine parallel documents from multilingual web sites. Besides structure similarity, other systems either filter fetched web pages by keeping only those containing language markers in their URLs (Désilets et al., 2008), or employ a predefined bilingual wordlist (Chen et al., 2004), or a naive aligner (Zhang et al., 2006) in order to estimate the content similarity of candidate parallel web pages. More recently, some work (Pecina et al., 2011; Pecina et al., 2012; Pecina et al., 2014) explore using domain focused web-crawled resources (e.g., monolingual, parallel, comparable corpora and dictionaries) to adapt language model and translation model.

Among above approaches, we focus on two of them: data selection, which solves the ambiguity problems by adjusting the data distribution of training corpora; domain focused web-crawling, which reduces the OOVs by mining domain-specific dictionary, parallel and monolingual sentences from comparable corpora. In the following two sub-sections, existing models regarding these directions are reviewed.

### 2.2.2 DATA SELECTION FOR SMT DOMAIN ADAPTATION

In SMT, one of the most dominant approaches involve selecting data suitable for the domain at hand from large general-domain corpora, the assumption being that if a general corpus is broad enough it will contain sentences that are similar to those that occur in the specific domain. It aims at finding such appropriate data from large general-domain corpora are called supplementary data selection approaches.

Data selection can be used for language model adaptation and translation model adaptation by applying methods on monolingual and parallel corpora, respectively. As shown in Figure 2-3, lager general-domain and small in-domain corpus are at hand. We select a set of relative data from general-domain corpus by measuring the similarity between candidates and in-domain data. The selected subsets are called pseudo in-domain corpus, which are used to train adapted language model or translation model. Finally, we linearly or log-linearly integrate the adapted models with existing in-domain models. The final models outperform the ones trained on all of data due to adjusting the data distribution to target domain.

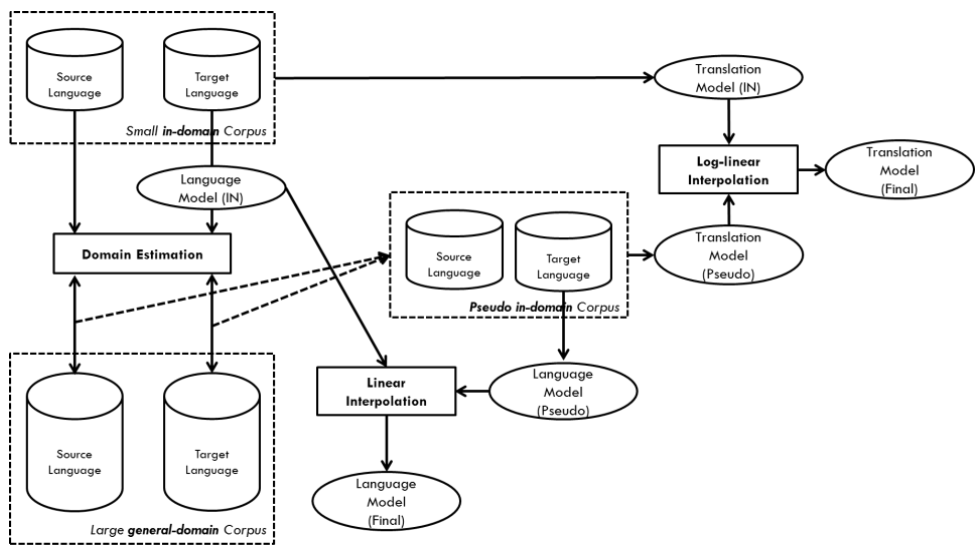

Figure 2-3: Data Selection Framework

Formally, data selection for SMT can be factored into three processing stages:

**Scoring**: given two parallel corpora: a general-domain corpus $G$ and an in-domain corpus $R$, each sentence pair in $G$ can be scored as follows:

$$Score_{<S_i,T_i>} \rightarrow Sim(V_i, M_R) \qquad (2\text{-}1)$$

where $S_i$ and $T_i$ are the source and target side of the $i$-th sentence pair. The source sentences $<S_i>$ and target sentences $<T_i>$ can be scored individually, or both sides $<S_i,T_i>$ can be used to measure similarity. We define the set $\{<S_i>, <T_i>, <S_i,T_i>\}$ as $V_i$. $M_R$ is an abstract model representing the target domain.

**Resampling**: two ways are used to resample data. One is ranking the scored sentences and selecting top K ($0 < K <$ Size of general-domain corpus) of them as sampled data. The other is setting a score value $N$ to classify sentence into in-domain and non-in-domain according to a filter function:

$$Filter_i = \begin{cases} 1, & Score_{<S_i,T_i>} > \theta \\ 0, & Others \end{cases} \qquad (2\text{-}2)$$

in which $Filter_i$ is 1 for the sentence pair $<S_i,T_i>$ when the similarity score is higher than a tunable threshold $\theta$, and 0 otherwise. A pseudo in-domain sub-corpus is then built by bootstrapping using sentences from the general-domain corpus.

**Translation**: adapted translation models (in Section 1.1.4) or language models (in Section 1.1.5) can be obtained by training on these pseudo in-domain sub-corpora.

The core of data selection is similarity matric (in Eq. (2-1)) used to measure domain relevance of each sentence. According to commonly-used similarity functions, they can be divided into two categories: 1) vector space model (VSM), which converts sentences into a term-weighted vector and then applies a similarity function to rank them (Zhao et al., 2004; Hildebrand et al., 2005; Lü et al., 2007); 2) perplexity-based model, which employs a $n$-grams domain-specific language model to score the perplexity of each sentence (Lin et al., 1997; Gao et al., 2002; Moore and Lewis, 2010; Axelrod et al., 2011).

Vector space model (VSM) converts sentences into a term-weighted vector and then applies a vector similarity function to rank them. The sentence $S_i$ is represented as a vector:

$$S_i = \langle w_{i1}, w_{i2}, ..., w_{in} \rangle \tag{2-3}$$

in which $n$ is the size of the vocabulary and $w_{ij}$ is standard *tf-idf* weight:

$$w_{ij} = tf_{ij} \times \log(idf_j) \tag{2-4}$$

in which $tf_{ij}$ is the term frequency (TF) of the $j$-th word in the vocabulary in the document $D_i$, and $idf_j$ is the inverse document frequency (IDF) of the $j$-th word calculated. VSM uses the similarity score between the vector representing the in-domain sentences and the vector representing each sentence in general-domain corpus. There are many similarity functions we could have employed for this purpose (Cha, 2007). A simple but effective one is cosine measure, which is defined as:

$$\cos \theta = \frac{S_{Gen} \cap S_{IN}}{\|S_{Gen}\| \|S_{IN}\|} \tag{2-5}$$

where $S_{Gen} \cap S_{IN}$ is the intersection (i.e. the dot product) of the sentence vector in general-domain and the one in in-domain, $\|S_i\|$ is the norm of vector $S_i$.

Zhao et al. (2004) firstly use this information retrieval technique to retrieve relative sentences from monolingual corpus to build a LM, and then interpolate it with general-background LM for LM adaptation. Hildebrand et al. (2005) extended it to sentence pairs, which are used to train a domain-specific TM. They concluded that it is possible to adapt this method to improve the translation performance especially in the LM adaptation. Similar to the experiments described in this paper, Lü et al. (2007) proposed re-sampling and re-weighting methods for online and offline TM optimization, which are closer to a real-life SMT system. Furthermore, their results indicated that duplicated sentences can affect the translations. They obtained about 1 BLEU point improvement using 60% of total data.

Perplexity-based approaches employ *n*-gram domain-specific language models to score the perplexity of each sentence in general-domain corpus. As perplexity (as shown in Eq. (1-9)) and cross-entropy (as shown in Eq. (1-10)) are monotonically related, both are used to measure domain relevance. Until now, there are three perplexity-based variants. The first is called basic cross-entropy given by:

$$H_{I-src}(x) \tag{2-6}$$

where $H_{I-src}(x)$ is the cross-entropy of string *x* according to language model trained on the source side (*src*) of in-domain (*I*) corpus. The second one is Moore-Lewis cross-entropy difference (Moore and Lewis, 2010):

$$H_{I-src}(x) - H_{O-src}(x) \tag{2-7}$$

which tries to select the sentences that are more similar to *I* but different to non-in-domain data (*O*). A LM is built on a random subset (equal in size to corpus *I*) of the general-domain corpus (*G*). All above two methods only consider the information in source language. Furthermore, Axelrod et al. (2011) proposed modified Moore-Lewis that sums cross-entropy difference over both source side (*src*) and target side (*tgt*):

$$\begin{aligned} &\left[ H_{I-src}(x) - H_{O-src}(x) \right] \\ &+ \left[ H_{I-tgt}(x) - H_{O-tgt}(x) \right] \end{aligned} \tag{2-8}$$

Perplexity-based methods have been adapted by Lin et al. (1997) and Gao et al. (2002), in which perplexity is used to score text segments according to an in-domain LM. More recently, Moore and Lewis (2010) derived the cross-entropy difference metric from a simple variant of Bayes rule. However, this is a preliminary study that did not yet show an improvement for MT task. The method was further developed by Axelrod et al. (2011) for SMT adaptation. They also presented a novel bilingual method and compared it with other variants. The experimental results show that the fast and simple technique allows to discard over 99% of the general corpus resulted in an increase of 1.8 BLEU points.

### 2.2.3 DOMAIN FOCUSED WEB-CRAWLING

Parallel corpus is a valuable resource for cross-language information retrieval and data-driven natural language processing systems, especially for Statistical Machine Translation (SMT). Therefore, people proposed many approaches from different perspectives to mine useful parallel information for machine translation.

The biggest and most heterogeneous text corpus in the world is the Word Wide Web (Philip and Noah, 2003). It is known that many websites are available in multiple languages, which means some of them can be paired into bitexts. Based on this point, different systems have been developed to harvest bitexts from the Internet. Resnik and Smith (2003) proposed the STRAND system, in which they used Altavista to search for multilingual websites and examined the similarity of the HTML structures of the fetched web pages in order to identify pairs of potentially parallel pages. Similarly, Esplà-Gomis and Forcada (2010) proposed Bitextor, a system that exploits shallow features (file size, text length, tag structure, and list of numbers in a web page) to mine parallel documents from multilingual web sites. Besides structure similarity, other systems either filter fetched web pages by keeping only those containing language markers in their URLs (Désilets et al., 2008), or employ a predefined bilingual wordlist (Chen et al., 2004), or a naive aligner (Zhang et al., 2006) in order to estimate the content similarity of candidate parallel web pages.

The crawled corpora are often comparable, which is not able to be used directly for SMT task. Thus, some work on sentence or document alignment from comparable corpus (Koehn, 2005; Tiedemann, 2009; 2010; 2011; 2012). Some well-designed algorithms and tools can be used in practice, such as the work of Patry and Langlais (2005) in document alignment (with a precision of 99%), the works of Koehn (2005) and Gillick (2009) in sentence boundary detection (error rates on test news data are less than 0.25%), and the work of Moore (2002) in sentence alignment (it achieves 99.34% in precision). The standard parallel corpus construction follows the process as illustrated in Figure 2-4. The overall construction process is divided into five major steps. The initial step is to identify the appropriate sources of the websites that contain the data and crawl the documents which are bilingual ready. In the second step of content extraction, the HTML files are parsed by discarding all the HTML tags profits from the function of NekoHTML[15] and XPath[16]. At the same time, the type of documents is analyzed for categorizing the text domain and topics in the subsequent stage. Documents are aligned in bilingual correspondence. Information together with the texts is stored in some unified formats. A key bridge between aligned documents and aligned sentences is the sentence boundary detection process. Different detection results will affect the alignment relation of sentence pairs, i.e. one to one or many to one alignment relationship. So far, the processing flow is automatically done. The final result is verified by human to get rid of the noisy texts, in particular the low quality translations.

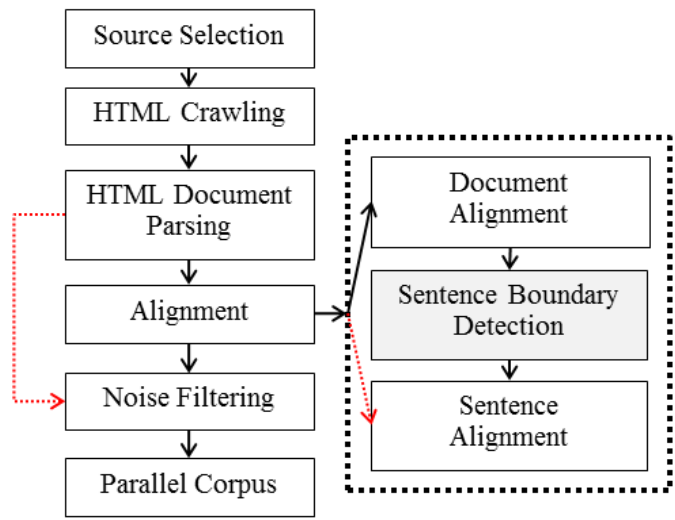

Figure 2-4: Web-Crawling Framework

In our thesis, we mainly focus on cross-language document alignment and domain focused web-crawling.

**Cross-Language Document Retrieval**

The issues of CLIR have been discussed from different perspectives for several decades. In this section, we briefly describe some related methods. From a statistical perspective, the CLIR problem can be treated as document alignment. Given a set of parallel documents, the alignment that maximizes the probability over all possible alignments is retrieved (Gale and Church, 1991) as follows:

$$\arg\max \Pr(A \mid D_s, D_t) \approx \arg\max_{A} \prod_{(L_s \Leftrightarrow L_t) \in A} \Pr(L_s \leftrightarrow L_t \mid L_s L_t) \qquad (2\text{-}9)$$

where $A$ is an alignment, $D_s$ and $D_t$ are the source and target documents, respectively $L_1$ and $L_2$ are the documents of two languages, $L_s \leftrightarrow L_t$ is an individual aligned pairs, an alignment $A$ is a set consisting of $L_s \leftrightarrow L_t$ pairs.

---

[15]  http://nekohtml.sourceforge.net/.

[16]  http://www.w3schools.com/xpath/default.asp.

On the matching strategies for CLIR, query translation is most widely used method due to its tractability (Gao et al., 2001). However, it is relatively difficult to resolve the problem of term ambiguity because "queries are often short and short queries provide little context for disambiguation" (Oard and Diekema, 1998). Hence, some researchers have used document translation method as the opposite strategies to improve translation quality, since more varied context within each document is available for translation (Braschler and Schauble, 2001; Franz et al., 1999).

However, another problem introduced based on this approach is word (term) disambiguation, because a word may have multiple possible translations (Oard and Diekema, 1998). Significant efforts have been devoted to this problem. Davis and Ogden (1997) applied a part-of-speech (POS) method which requires POS tagging software for both languages. Marcello et al. presented a novel statistical method to score and rank the target documents by integrating probabilities computed by query-translation model and query-document model (Federico and Bertoldi, 2002). However, this approach cannot aim at describing how users actually create queries which have a key effect on the retrieval performance. Due to the availability of parallel corpora in multiple languages, some authors have tried to extract beneficial information for CLIR by using SMT techniques. Sánchez-Martínez et al. (Sánchez-Martínez and Carrasco, 2011) applied SMT technology to generate and translate queries in order to retrieve long documents.

Some researchers like Marcello, Sánchez-Martínez et al. have attempted to estimate translation probability from a parallel corpus according to a well-known algorithm developed by IBM (Brown et al., 1993). The algorithm can automatically generate a bilingual term list with a set of probabilities that a term is translated into equivalents in another language from a set of sentence alignments included in a parallel corpus. The IBM Model 1 is the simplest among the five models and often used for CLIR. The fundamental idea of the Model 1 is to estimate each translation probability so that the probability represented is maximized

$$P(t \mid s) = \frac{\varepsilon}{(l+1)^m} \prod_{j=1}^{m} \sum_{i=0}^{l} P(t_j \mid s_i) \qquad (2\text{-}10)$$

where $t$ is a sequence of terms $t_1, \ldots, t_m$ in the target language, $s$ is a sequence of terms $s_1, \ldots, s_l$ in the source language, $P(t_j/s_i)$ is the translation probability, and $\varepsilon$ is a parameter ($\varepsilon = P(m|e)$), where $e$ is target language and $m$ is the length of source language). Eq. (2-10) tries to balance the probability of translation, and the query selection, in which problem still exists: it tends to select the terms consisting of more words as query because of its less frequency, while cutting the length of terms may affect the quality of translation. Besides, the IBM model 1 only proposes translations word-by-word and ignores the context words in the query. This observation suggests that a disambiguation process can be added to select the correct translation words (Oard and Diekema, 1998). However, in our method, the conflict can be resolved through contexts.

If translated sentences share cognates, then the character lengths of those cognates are correlated (Yang and Li, 2004). Brown, et al. (1991) and Gale and Church (1991) have developed the models based on relationship between the lengths of sentences that are mutual translations. Although it has been suggested that length-based methods are language-independent (Gale and Church, 1991), they really rely on length correlations arising from the historical relationships of the languages being aligned.

The length-based model assumes that each term in $L_s$ is responsible for generating some number of terms in $L_t$. This leads to a further approximation that encapsulates the dependence to a single parameter $\delta$. $\delta(l_s, l_t)$ is function of $l_s$ and $l_t$, which can be designed according to different language pairs. The length-based method is developed based on the following approximation to Eq. (2-11):

$$\Pr(L_s \leftrightarrow L_t \mid L_s, L_t) \approx \Pr(L_s \leftrightarrow L_t \mid \delta(l_s, l_t)) \qquad (2\text{-}11)$$

**Domain Focused Web-Crawling**

As data in specific domain are usually relatively scarce, the use of web resources to complement the training resources provides an effective way to enhance the SMT systems. More recently, some work (Pecina et al., 2011; Pecina et al., 2012; Pecina et al., 2014) explore using domain focused web-crawled resources (e.g., monolingual, parallel, comparable corpora and dictionaries) to adapt language model and translation model.

A key challenge for a focused crawler that aspires to build domain-specific web collections is the prioritisation of the links to follow. Several algorithms have been exploited for selecting the most promising links. The Best-First algorithm (Cho et al., 1998) sorts the links with respect to their relevance scores and selects a predefined amount of them as the seeds for the next crawling cycle. Menczer and Belew (2000) proposed an adaptive population of agents, called InfoSpiders, and searched for pages relevant to a domain using evolving query vectors and Neural Networks to decide which links to follow. Hybrid models and modifications of these crawling strategies have also been proposed (Gao et al., 2010) with the aim of reaching relevant pages rapidly.

Apart from the crawling algorithm, classification of web content as relevant to a domain or not also affects the acquisition of domain-specific resources, on the assumption that relevant pages are more likely to contain links to more pages in the same domain. Qi and Davison (2009) review features and algorithms used in web page classification. In most of the algorithms reviewed, on-page features (i.e. textual content and HTML tags) are used to construct a corresponding feature vector and then, several machine-learning approaches, such as SVMs, Decision Trees, and Neural Networks, are employed (Yu et al., 2004).

CHAPTER 3: INTELLIGENT DATA SELECTION FOR SMT DOMAIN
ADAPTATION

*Data selection is to use an in-domain translation model, combined with sentence pairs from out-of-domain that are similar to the in-domain text.*

-- Rico Sennrich, 2013

In this chapter, we firstly present a new similarity metric as selection criterion to select better data to enhance the exsiting models. Experimental results show that this high-contrined similariy measure help to retrive more reletive data than other commomly-used ones (Cosine tf-idf and cross entropy methods). To further improve the performance, we combine three different individual selection models at both corpus level and model level. Then we systematally compare and analysis on different data seletion models. Finally, we present a novel perplexity-based methed by considering the linguitic information.

## 3.1 EDIT DISTANCE: A NEW DATA SELECTION CRITERION

This section aims at effective use of training data by extracting sentences from large general-domain corpora to adapt statistical machine translation systems to domain-specific data. We regard this task as a problem of scoring training sentences with respect to the target domain via different similarity metrics. Thus, we explore which data selection model can best benefit the in-domain translation. Comparing the VSM-based and perplexity-based methods (according to the description in in Section 2.2.2), we found that VSM-based methods have weakness at filtering irrelevant data due its simple single word matching algorithm. However, perplexity-based approaches show better filtering ability with considering the n-gram words correlation. Thus, we found that the more information considered during measuring, the better ability of filtering obtained. Based on this point, we propose a string-difference metric as data selection criterion, which comprehensively considers the word matching, correlation and position. We hypothesize that the string-difference based method is a viable alternative (Wang et al., 2013).

One of string-difference metrics, edit distance is a widely used similarity measure, known as Levenshtein distance (Levenshtein, 1966). Koehn and Senellart (2010) applied this algorithm in translation memory work. Leveling et al. (2012) investigated different approximated sentence retrieval approaches for example-based MT. Both of them gave the formula for fuzzy matching. This inspires us to regard this metric as a new data selection criterion for SMT domain adaptation task. Good performance could be expected under the assumption that the general corpus is big enough to cover the very similar sentences with respect to the test data.

To evaluate this proposal, we compare it with other two state-of-the-art methods on a large dataset. Comparative experiments are conducted on Chinese-English travel domain and the results indicate that the proposed approach achieves consistent and significant improvement over the baseline system (+4.36 BLEU) as well as the best rival model (+1.23 BLEU) using a much smaller subset. This study has a profound implication for mining very large corpora in a computationally-limited environment.

### 3.1.1 METHODOLOGY

In this section, we mainly describe our proposed and comparative selection models. Then we use the selected subsets (called pseudo in-domain corpus) to train translation models (described in Section 1.1.4) and language models (described in Section 1.1.5) for SMT task.

**Proposed Model**

Given a sentence $s_G$ from general corpus and a sentence $s_R$ from the test set or in-domain corpus, the edit distance for these two sequences is defined as the minimum number of edits, i.e. symbol insertions, deletions and substitutions, needed to transform $s_G$ into $s_R$. There are several different implementations of the edit-distance-based retrieval model. We used the normalized Levenshtein similarity score (fuzzy matching score, FMS) proposed by Koehn and Senellart (2010):

$$FMS(s_G, s_R) = 1 - \frac{LED_{word}(s_G, s_R)}{Max(|s_G|, |s_R|)} \tag{3-1}$$

in which $ED(s_G, s_R)$ is a distance function and $|s|$ is the number of tokens of sentence *s*. In this study, we only employed a word-based Levenshtein edit distance function ($LED_{word}$) instead of additionally using letter-based $ED$. If there is a sentence of which score exceeds a threshold, we will further penalize it according to space and punctuations edit differences.

In practice, we apply this function (Eq. (3-2)) as similarity function. Therefore, each sentence $S_G$ in general-domain corpus can be scored as:

$$Score(s_G) = \frac{1}{N} \sum_{i}^{N} FMS(s_G, s_{I_i}) \tag{3-2}$$

in which *N* is the size of in-domain corpus and $s_{I_i}$ is the *i*th sentence in in-domain corpus. FMS algorithm is given in Eq. (3-1). Then we select the *K%* sentences with higher score for model training, where is *K* is a tunable threshold.

**Comparative Models**

We compare the proposed method with other two typical data selection methods: VSM-based and perplexity-based (detailed in Section 2.2.2).

For VSM-based data selection, the selection method is similar to edit-distance based one. The only difference is that we apply cosine *tf-idf* (Eq. (2-3), (2-5) and (2-6)) as similarity function.

For perplexity-based models, we implement all three variants: cross-entropy (Eq. (2-6)), Moore-Lewis (Eq. (2-7)) and modified Moore-Lewis (Eq. (2-8)). We use them to directly score each sentence in general-domain corpus. Then we select the *K%* sentences with higher score for model training.

### 3.1.2 EXPERIMENTAL SETUP

Two corpora are needed for the domain adaptation task. Our general corpus includes 5 million English-Chinese parallel sentences comparing a various genres such as movie subtitles, law literature, news and novels. The in-domain corpus and test set are randomly selected from the IWSLT2010 (International Workshop on Spoken Language Translation) Chinese-English Dialog task[17], consisting of transcriptions of conversational speech in a travel setting. All of them were identically segmented[18] (Zhang, 2003) and tokenized[19] (Koehn, 2005). The sizes of the test set, in-domain corpus and general corpus we used are summarized in Table 3-1.

Table 3-1: Corpora statistics

| Data Set | Sentences | Tokens | Ave. Len. |
|---|---|---|---|
| Test Set | 3,500 | 34,382 | 9.60 |
| In-domain Training Corpus | 17,975 | 151,797 | 9.45 |
| General-domain Training Corpus | 5,211,281 | 53,650,998 | 12.93 |

All experiments presented in this paper are carried out with the Moses toolkit (Koehn et al., 2007), a state-of-the-art open-source phrase-based SMT system. The translation and the re-ordering model relied on "*grow-diag-final*" symmetrized word-to-word alignments built using GIZA++ (Och and Ney, 2003) and the training script of Moses. A 5-gram language model was trained on the target side of the training parallel corpus using the IRSTLM toolkit (Federico et al., 2008), exploiting improved Modified Kneser-Ney smoothing, and quantizing both probabilities and back-off weights.

As described in Section 3.1.1, a number of SMT systems are trained on pseudo in-domain corpus obtained by different selection models. Totally five systems are built:

- **Baseline**, translation models and language models are train on entire general corpus.

---

[17] Available at http://iwslt2010.fbk.eu/node/33.

[18] IC-TCLAS2013 is available at http://ictclas.nlpir.org/.

[19] Scripts are available at http://www.statmt.org/europarl/.

- **IR**, translation models and language models are train on $K$% general corpus ranked by cosine *tf-idf*.

- **CE**, translation models and language models are train on $K$% general corpus ranked by cross-entropy.

- **CED**, translation models and language models are train on $K$% general corpus ranked by Moore-Lewis.

- **B-CED**, translation models and language models are train on $K$% general corpus ranked by modified Moore-Lewis.

- **FMS$_{ours}$,** translation models and language models are train on $K$% general corpus ranked by edit-distance.

### 3.1.3 Results and Discussions

Considering that the best result of each method may depend on the size of the selected data, we investigate each of selected corpora in a step of 2x starting from using 0.25% of general corpus (0.29%, 0.52%, 1.00%, 2.30%, 4.25% and 12.5%) where $K$% means $K$ percentage of general corpus are selected as a subset.

The baseline consisted of a SMT system trained with toolkits and settings as described above on general corpus and the BLEU is **29.34** points. The baseline is a bit lower, because general corpus does not consist of enough sentences on domain of travel and the out-of-domain data can be treated as noises for in-domain set.

Firstly, we evaluated IR which improves by at most 1.03 BLEU points when using 4.25% data of the general corpus as shown in Figure 3-1. Then the performance begins to drop when the size threshold is more than 4.25%. The results show that keywords overlap plays a significant role in retrieving sentences in similar domains. However, it still needs a large amount of selected data to obtain an ideal performance due to its weakness in filtering noises.

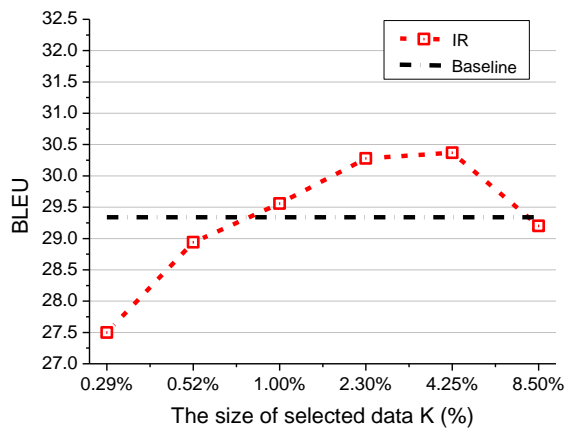

Figure 3-1: Translation Results Using Subset of General Corpus Selected by Standard IR Model

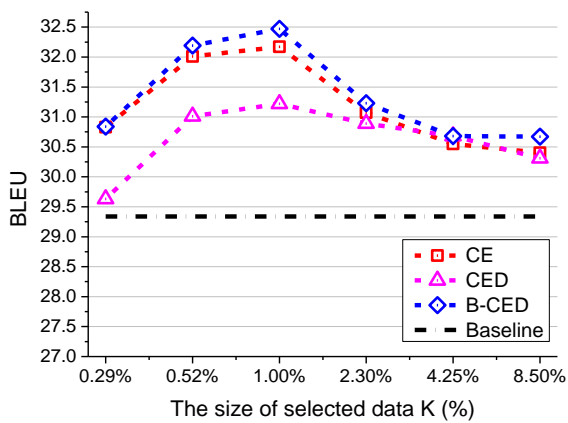

Figure 3-2: Translation Results Using Subset of General Corpus Selected by Three Perplexity-Based Variants

Secondly, we compared three perplexity-based methods. As illustrated in Figure 3-2, all of them were able to significantly outperform the baseline system using only 1% of entire training data. The size threshold is much smaller than the one of IR when obtaining the equivalent performance. Besides, the curve drops slowly and always over the baseline. This shows a better ability of filtering noises. Among the perplexity-based variants, the B-CED works best, which is similar to the conclusion drawn by Axelrod et al. (2011). It proves that bilingual resources are helpful to balance OOVs and noises. Next we will use B-CED to stand for perplexity-based methods and compare with other selection criteria.

Finally, we evaluated FMS and compared it with IR, B-CED and the baseline system, which are shown in Figure 3-3. FMS seems to give an outstanding performance on most size thresholds. It always outperforms B-CED over at least 1 point under the same settings. Even using only 0.29% data, the BLEU is still higher than baseline over 0.66 points. In addition, FMS is able to conduct a better in-domain SMT system using less data than other selection methods. This indicates that it is stronger to filter noises and keep in-domain data when considering more constrain factors for similarity measuring.

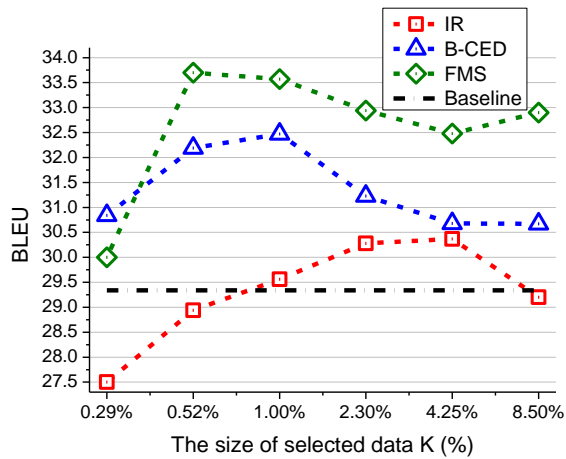

Figure 3-3: Translation Results Using Subset of General Corpus Selected by Different Methods

Table 3-2: Best Result of Each Method with Corresponding Size of Subset

| Corpus | Size (%) | BLEU |
|--------|----------|------|
| **Baseline** | 100 | 29.34 |
| **IR** | 4.25 | 30.37 (+1.03) |
| **CE** | 1.00 | 32.17 (+2.83) |
| **CED** | 1.00 | 31.22 (+1.88) |
| **B-CED** | 1.00 | 32.47 (+3.13) |
| **FMS$_{ours}$** | **0.52** | 33.70 (**+4.36**) |

To give a better numerical comparison, Table 3-2 lists the best result of each method. As expected, FMS could use the smallest data (0.52%) to achieve the best performance. It outperforms baseline system trained on entire dataset over 4.36 BLEU points and B-CED over 1.23 points.

3.1.4 SECTION SUMMARY

In this section, we regard data selection as a problem of scoring the sentences in general corpus via different similarity metrics. After revisiting the state-of-the-art data selection methods for SMT adaptation, we make edit distance as a new selection criterion for this topic. In order to evaluate the proposed method, we compare it with four other related methods on a large data set. The methods we implemented are standard information retrieval model, source-side cross-entropy, source-side cross-entropy difference, bilingual cross-entropy difference as well as a baseline system. We can analyze the results from two different aspects:

- **Translation Quality**: The results show a significant performance of the proposed method with increasing 4.36 BLEU points than the baseline system. And it also outperforms other four methods over 1-3 points.

- **Filtering Noises**: Fuzzy matching could discard about 99.5% data of the general corpus without reducing translation quality. However, other methods will drop their performance when using the same size of data. The proposed metric has a very strong ability to filter noises in general corpus.

Finally, we can draw a composite conclusion that edit distance is a more suitable similarity model for SMT domain adaptation.

## 3.2 A HYBRID DATA SELECTION MODEL

Until now, three state-of-the-art selection criteria have been discussed (in Section 3.1). The analysis shows that each individual retrieval model has its own advantages and disadvantages, which result in their performance either unclear or unstable. Instead of exploring any single individual model, we propose a hybrid data selection model named *i*CPE, which combines three state-of-the-art similarity metrics: **C**osine *tf-idf*, **P**erplexity and **E**dit distance at both corpus level and model levels: i) *corpus level* where joining the sub-corpora obtained via a different individual model; and ii) *model level* where interpolating multiple TMs or LMs together.

To compare the proposed model with the presented individual models, we conduct comparative experiments on a large Chinese-English general corpus to adapt to in-domain sentences on Hong Kong law. Using BLEU (Papineni et al., 2002) as an evaluation metric, results indicate this simple and effective hybrid model performs better over the baseline system trained on entire data as well as the best rival method. This consistently boosting the performance of the proposed approach has a profound implication for mining very large corpora in a computationally-limited environment.

### 3.2.1 METHODOLOGY

We use different pseudo in-domain corpora retrieved by different individual and combined methods to train translation models (described in Section 1.1.4) and language models (described in Section 1.1.5) for SMT domain adaptation. For comparison, apply all individual models, which described in Section 3.1.1.

The existing domain adaptation methods can be summarized into two broad categories: i) *corpus level* by selecting, joining, or weighting the datasets upon which the models are trained; and ii) *model level* by combining multiple models together in a weighted manner.

For corpus level combination, we weight the sub-corpora retrieved by different methods by modifying the frequencies of the sentence in the GIZA++ file (Lü et al., 2007) and then join them together. It can be formally stated as follows:

$$
\begin{aligned}
iCPE_{(S_i, T_i)} &= \alpha CosIR(S_x, T_x) \\
&+ \beta PPBased(S_y, T_y) \\
&+ \lambda EDBased(S_z, T_z)
\end{aligned}
\tag{3-3}
$$

where $\alpha$, $\beta$ and $\lambda$ are the weights for different criteria. $(S_x, T_x)$, $(S_y, T_y)$ and $(S_z, T_z)$ are the sentence pairs respectively selected by cosine *tf-idf* (*CosIR*), perplexity-based (*PPBased*) and edit-distance based (*EDBased*).

For model level combination, we perform linear interpolation on the models trained with the sub-corpora retrieved by different data selection methods. The phrase translation probability $\phi(\bar{f}|\bar{e})$ and the lexical weight $p_w(\bar{f}|\bar{e},a)$ are estimated using Equation 3-4 and Equation 3-5, respectively.

$$\phi(\bar{f}|\bar{e}) = \sum_{i=0}^{n} \alpha_i \phi_i(\bar{f}|\bar{e}) \qquad (3\text{-}4)$$

$$p_w(\bar{f}|\bar{e},a) = \sum_{i=0}^{n} \beta_i p_{w,i}(\bar{f}|\bar{e},a) \qquad (3\text{-}5)$$

where $i = 1, 2, 3$ denote phrase translation probability and lexical weight trained with the sub-corpora retrieved by *CosIR*, *PPBased* and *EDBased*. $\alpha_i$ and $\beta_i$ are the interpolation weights.

### 3.2.2 EXPERIMENTAL SETUP

Our general-domain corpus includes more than 1 million parallel sentences comprising various genres such as newswires (LDC2005T10), sample sentences from dictionaries, law literature and other crawled sentences. The distribution of domains and sentence length of the general corpus are shown in Table 3-3 and Figure 3-4, respectively. The in-domain corpus and test set are randomly selected that are disjoined from the LDC corpus (LDC2004T08), consisting of texts of Hong Kong law. All of them were segmented (with the same segmentation scheme)[20] (Zhang et al., 2003; Wang et al., 2012) and tokenized[21] (Koehn, 2005). In the preprocessing, we also removed the sentences with length more than 80. To evaluate the methods for both LM and TM, we used the target side sentences of the corpora to train all the LMs for translation. The sizes of the test set, in-domain corpus and general corpus we used are summarized in Table 3-4.

Table 3-3: Domain proportions in general corpus

| Statistics | Domains | | | | Total |
|---|---|---|---|---|---|
| | *News* | *Novel* | *Law*[b] | *Miscellaneous*[a] | |
| **Sentence Number** (#) | 279,962 | 304,932 | 48,754 | 504,396 | 1,138,044 |
| **Percentage** (%) | 24.60 | 26.79 | 4.28 | 44.33 | 100.00 |

---

[20] IC-TCLAS2013 is available at http://ictclas.nlpir.org/.

[21] The scripts are available at http://www.statmt.org/europarl/.

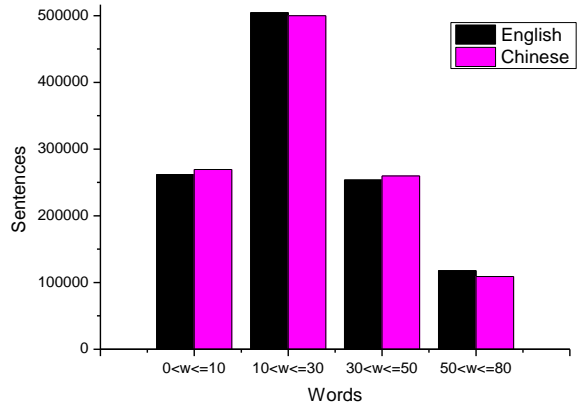

Figure 3-4: Distributions of Sentences (length) of General Corpus

Table 3-4: Statistics Summary of Used Corpora

| Data Set | Language | Sentences | Tokens | Ave. Len. |
|---|---|---|---|---|
| **Test Set** | English | 2,050 | 60,399 | 29.46 |
| | Chinese | | 59,628 | 29.09 |
| **In-domain Training Corpus** | English | 45,621 | 1,330,464 | 29.16 |
| | Chinese | | 1,321,655 | 28.97 |
| **In-domain Training Corpus** | English | 1,138,044 | 28,626,367 | 25.15 |
| | Chinese | | 28,239,747 | 24.81 |

The experiments presented in this paper were carried out with the Moses toolkit (Koehn et al., 2007), a state-of-the-art open-source phrase-based SMT system. The translation and the re-ordering model relied on "*grow-diag-final*" symmetrized word-to-word alignments built using GIZA++ (Och and Ney, 2003) and the training script of Moses. A 5-gram language model was trained using the IRSTLM toolkit (Federico et al., 2008), exploiting improved Modified Kneser-Ney smoothing, and quantizing both, probabilities and back-off weights.

In previous work, cosine *tf-idf* method often selected data using test set as reference set (Hildebrand et al., 2005; Lü et al., 2007), which limits the practical applicability of the method in a real-life SMT system. For perplexity-based approaches, an in-domain corpus which is identical to the test sentences is employed for data selection (Moore and Lewis, 2010; Axelrod et al., 2011). To compare the different methods fairly, we propose two strategies: one is *offline strategy* where we use test set to find similar sentences in general corpus; the other one is called *online strategy* where an additional in-domain corpus is used to select useful data.

In order to evaluate the performance of our model, we compare it with other five individual systems and baseline:

- **Baseline**, translation models and language models are trained on an entire general corpus.
- **Cos-IR,** translation models and language models are trained on $K$% general corpus ranked by cosine *tf-idf*.
- **B-CED**, translation models and language models are trained on $K$% general corpus ranked by modified Moore-Lewis.
- *i***CPE-C**, translation models and language models are trained on $K$% general corpus combined by proposed corpus-level method.
- *i***CPE-M**, translation models and language models are trained on $K$% general corpus combined by proposed model-level method.

### 3.2.3 RESULTS AND DISCUSSIONS

For each method, we selected the top $N=\{80K, 160K, 320K\}$ sentence pairs out of the $1.1M$ in the general corpus[22]. Table 3-5 contains BLEU scores of the systems trained on subsets selected via different models.

Table 3-5: Translation Results via Different Methods

| Method | Sentences | BLEU (Offline) | BLEU (Online) |
|---|---|---|---|
| **GC-Baseline** | $1.1M$ | **39.15** | |
| **IC-Baseline** | $1.1M$ | 36.30 | |
| **Cos-IR** | $80K$ | 39.04 | 37.53 |
| | $160K$ | 39.85 | 39.45 |

---

[22]  Roughly 7.0%, 14.0%, 28.0% of general-domain corpus. Besides, $K$ is short for thousand and M is short for million.

|        | 320$K$ | **40.17** | **40.03** |
|--------|--------|-----------|-----------|
| **B-CED** | 80$K$  | 40.91     | 35.50     |
|        | 160$K$ | **41.12** | 39.47     |
|        | 320$K$ | 40.02     | **40.98** |
| **FMS**   | 80$K$  | 37.42     | 36.22     |
|        | 160$K$ | 37.90     | 36.71     |
|        | 320$K$ | 38.15     | 38.00     |
| **$i$CPE-C** | 80$K$  | 42.25     | 39.39     |
|        | 160$K$ | **43.04** | **41.87** |
|        | 320$K$ | 42.42     | 40.44     |
| **$i$CPE-M** | 80$K$  | 42.93     | 40.57     |
|        | 160$K$ | 43.65     | 41.95     |
|        | 320$K$ | **43.97** | **42.21** |

All the methods but FMS could be used to train a state-of-the-art SMT system. Cos-IR improves by at most 1.02 (offline) and 0.88 (online) BLEU points using 28.12% of the general corpus. This shows that keywords overlap plays a significant role in finding sentences in similar domains. Besides, Cos-IR has a strong robustness because the selection with online strategy still works well. However, it needs a large amount of selected data (28.0%) to obtain an ideal performance. The main reason is that the sentences including same keywords still may be irrelevant. For instance, there are two sentences including the same phrase "*according to the article*", but one may be in the domain of law and other one may be from news.

Perplexity-based variant B-CED works very well with the offline strategy. It achieves 41.12 (using 7.0% data) and 40.98 (using 14.0% data) BLEU with offline and online strategies. This indicates that bilingual resources are very useful to build a stable in-domain model. When using an in-domain corpus as the reference set, B-CED should enlarge the size of selected data to obtain an ideal BLEU. It has a good but unstable performance with different strategies. The main reason is that considering the word order may be helpful to filter the noise, but it depends heavily upon the in-domain LMs.

FMS fails to outperform the baseline system even it is much stricter than other criteria. When adding word position factor into similarity measuring, FMS tries to find nearly the same sentences on length, collocation and semantics. But our general corpus seems not large enough to cover a certain amount of FMS-similar sentences. With increasing the size of general or in-domain corpus, we believe FMS may work better.

We combined Cos-IR, FMS and B-CED (which is the best one among PPBased criteria) and gave equal weights (set $\alpha = \beta = \lambda = 1$ in Equation 3-4 and $\alpha_i = \beta_i = 1/3$ in Equation 3-5 and 3-6) to each component at two combination levels. At both levels, *i*CPE performs much better than other methods as well as the baseline systems. This shows a strong ability to balance the OOV and noise problems. On the one hand, filtering too much unmatched words may not sufficiently address the data sparsity issue of the SMT model; on the other hand, adding too much of the selected data may lead to the dilution of the in-domain characteristics of the SMT model. However, it seems to succeed the advantage of each individual model when combining them together. For instance, the performance of *i*CPE does not drop sharply (like PPBased approaches) when using an in-domain corpus as reference set. This not only shows its stronger robustness for building a real-life SMT system, but also proves that combination method works better than any single individual approach.

Furthermore, *i*CPE has achieved at most 3.89 (offline) and 2.72 (online) improvements over the baseline system at corpus level combination. Besides, the result is still higher than the best individual model (B-CED) by 1.92 (offline) and 0.91 (online). The performance can be further improved by interpolating at the model level. It works better (obtained around 1 BLEU point improvement) than the corpus combination method in the same settings.

### 3.2.4 SECTION SUMMARY

In this section, we regard data selection as a problem of measuring similarities via different criteria. This is the first time to systematically compare the state-of-the-art data selection methods for SMT adaptation. We not only explore edit-distance based method for this task for the first time, but also present offline and online strategies for fair comparison. We further integrate the presented individual data selection model at both corpus and model levels. It achieves a good performance in terms of its robustness and effectiveness. In order to evaluate the proposed data selection model on a large general corpus, we compare it with three other related methods: Cos-IR, B-CED, FMS as well as two baseline systems. We can analyze the results from three different aspects:

- **Translation Quality**. The results show a significant performance of the most methods in particular the proposed iCPE. It suggests better to use bilingual resources in similarity measuring.

- **Noise Filtering**. iCPE could discard about 93% data of the general corpus with a better translation quality. While other models perform either badly or unsteadily.

- **Robustness**. To build a real-life system, in-domain data set is preferable (online strategy). However, only iCPE gives a consistently boosting performance.

## 3.3 A SYSTEMATIC COMPARISON AND ANALYSIS ON DIFFERENT DATA SELECTION APPROACHES

Until now, we have explored various data selection methods for SMT domain adaptation. We divided some commonly-used approaches into three categories: 1) vector space model (VSM), which converts sentences into a term-weighted vector and then applies a similarity function to rank them; 2) perplexity-based model, which employs a n-grams domain-specific language model to score the perplexity of each sentence; 3) string-and-string difference, which consider the same or different terms between any two strings. In this section, we will compare all above methods with three novelties:

- **Large Corpora**. We evaluate these methods in large data environment. We hope it can show more real-life results.

- **Mixture Modeling**. As a small in-domain corpus is available, we log-linearly integrate the adapted models with the existing in-domain models (as described in Section 3.2.2). We hope to further improve the adapted models.

- **Deeply Analyzing**. We analyze three potential indicators such as vocabulary size, out-of-vocabulary words (OOVs) and overlapping. We anticipate an in-depth analysis of these typical methods from three different categories could be valuable to other work on this filed.

### 3.3.1 EXPERIMENTAL SETUP

Two corpora are needed for the domain adaptation task. The general corpus includes more than 1 million parallel sentences comprising varieties of genres such as newswires (LDC2005T10), translation example from dictionaries, law statements and sentences from online sources. The distribution of text genres of the general corpus is shown in Table 3-6. The miscellaneous part includes the sentences crawled from various materials and the law portion includes the articles collected from Chinese mainland, Hong Kong and Macau.

Table 3-6: Proportions of Different Text Domains of The English and Chinese General Corpus

| Domain | Sentence Number | Percentage (%) |
|---|---|---|
| News | 279,962 | 24.60 |
| Novel | 304,932 | 26.79 |
| Law | 48,754 | 4.28 |
| Miscellaneous | 504,396 | 44.33 |
| Total | 1,138,044 | 100.00 |

The in-domain corpus and test data are randomly selected that are disjoined from the Hong Kong law corpus (LDC2004T08). All of them were identically segmented[23] (Zhang et al., 2003) and tokenized[24] (Koehn, 2005). In the preprocessing, we removed the sentences of which length is more than 80. To evaluate each method on both TM adaptation and LM adaptation, we simply used the target side of parallel corpora to train the LMs during SMT training process. Thus each data selection step can optimize the data for both TM and LM.   The size of the test set, in-domain corpus and general corpus we used is summarized in Table 3-7.

---

[23] IC-TCLAS2013 is available at http://ictclas.nlpir.org/.

[24] Scripts are available athttp://www.statmt.org/europarl/.

Table 3-7: Detailed Statistics of Used Corpora

| Data Set | Language | Sentences | Words | Vocabulary | Ratio (V/W) |
|---|---|---|---|---|---|
| Test Set | EN | 2,050 | 60,399 | 5,510 | 0.09123 |
| | ZH | | 59,628 | 4,984 | 0.08358 |
| Dev Set | EN | 2,000 | 59,732 | 5,017 | 0.08399 |
| | ZH | | 59,064 | 4,854 | 0.08218 |
| In-domain Corpus | EN | 43,621 | 1,330,464 | 22,864 | 0.01718 |
| | ZH | | 1,321,655 | 18,446 | 0.01396 |
| General-domain Corpus | EN | 1,138,044 | 28,626,367 | 469,950 | 0.01642 |
| | ZH | | 28,239,747 | 278,206 | 0.00985 |

The experiments presented in this paper were carried out with the Moses toolkit (Koehn P. et al., 2007), a state-of-the-art open-source phrase-based SMT system. The translation and the re-ordering model relied on "grow-diag-final" symmetrized word-to-word alignments built using GIZA++ (Och and Ney, 2003) and the training script of Moses. A 5-gram language model was trained using the IRSTLM toolkit (Federico et al., 2008), exploiting improved modified Kneser-Ney smoothing, and quantizing both, probabilities and back-off weights.

For the comparison, totally five existing representative data selection models, three baseline systems and the proposed model were selected. The corresponding settings of the above models are as follows:

- **Baseline**: the in-domain baseline (IC-Baseline) and general-domain baseline (GC-Baseline) were respectively trained on in-domain corpus and general corpus. Then a combined baseline system (GI-Baseline) was created by passing the above two phrase tables to the decoder and using them in parallel.

- **Individual Model**: as described in Section 3.1, 3.2 and 3.3, the individual models are Cosine *tf-idf* (Cos-IR), fuzzy matching scorer (FMS) which is an edit-distance based (ED-Based) instance as well as three perplexity-based (PP-Based) variants: cross-entropy (CE), cross-entropy difference (CED), bilingual cross-entropy difference (B-CED).

- **Proposed Model**: as described in Section 3.2.1, we combined Cos-IR, PP-Based and ED-Based at corpus level (named *i*TPB-C) and model level (named *i*TPB-M).

### 3.3.2 RESULTS

We report our results in terms of the BLEU obtained by each of the models. For each method, we used the $N$ percent of the ranked general-domain data, where $N = \{15\%, 30\%, 45\%, 60\%, 75\%, 90\%$ and $100\%\}$.

Figure 3-5 shows the translation performance of each translation system using different data selection methods. We only plot the GI-Baseline (41.06) and G-Baseline (39.15) in the figure, because the BLEU of I-Baseline is only 36.30 and all the comparative systems can do much better than it. By observing the trends regarding the different models, we find that all the individual systems can outperform the GI-Baseline. In another words, all the presented data selection methods can be used to train adapted SMT systems. Perplexity-based models perform better than the ones in other categories. They show the powerful selection ability for this task. Especially for PP-MML, it can achieve the best performance among the individual models. On the constant, ED-LED performs poorly and its trend seems unstable. The main reason is that ED-LED only measures the difference between each two strings, instead of considering the global information in the whole data set (e.g., term distribution). About the selection size, perplexity-based methods, especially for PP-C and PP-MML can discard the more 50% of general-domain data to achieve similar or better performance. Although VSM-Cos and PP-ML have similar best BLEU points, they peak at different size thresholds. VSM-Cos has to use about 75% of general-domain data, which is much larger than PP-ML.

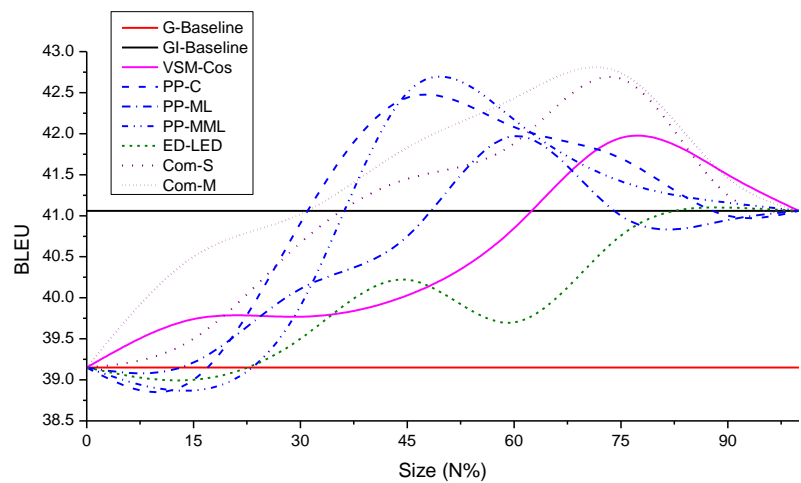

Figure 3-5: BLEU Scores Obtained by Different Data Selection Methods

Table 3-8 shows the BLEU scores by each model for the threshold for which the model obtains its best results (75% for VSM-Cos, 45% for PP-C, 60% for PP-ML, 45% for PP-MML, 90% for ED-LED, 75% for Com-S and 75% for Com-M). We show both the absolute BLEU points and the relative values compared to the baseline.

VSM-Cos and PP-NL can improve by nearly 1 points than the GI-Baseline. However, the PP-C and PP-MML can do better than them. They can achieve 42.44 (+1.38) and 42.50 (+1.44) BLEU scores. The combined models work best and outperform the baseline by more than 1.6 points. In addition, Com-M is slightly better than Com-S (+0.07 BLEU), which indicated that linear interpolation method may have similar performance with simple combination method.

Table 3-8: The Best BLEU of Each Comparative Data Selection Model

|  | GI-Baseline | VSM-Cos | PP-C | PP-ML | PP-MML | ED-LED | Com-S | Com-M |
|---|---|---|---|---|---|---|---|---|
| **BLEU** | 41.06 | 41.95 | 42.44 | 41.97 | 42.50 | 41.10 | 42.67 | 42.74 |
| **Diff.** |  | +0.89 | +1.38 | +0.91 | +1.44 | +0.04 | +1.61 | +1.68 |

### 3.3.3 ANALYSIS

Considering that only using the final translation output to evaluate each data selection method may miss some deeper factors between them, we also report the vocabulary size, OOVs and overlapping of each pseudo in-domain subset obtained by different methods[25].

**Vocabulary Size**

We report the vocabulary size of pseudo in-domain subcorpora obtained by different data selection models. Table 3-9 not only shows the absolute values of count, but also the relative values of difference between subcorpora and entire general-domain corpus. Note that, both Com-S and Com-M contain the same sentence pairs (described in Section 3). Thus, we just use *Com* to stand for both them.

Table 3-9: Vocabulary Size of Pseudo In-Domain Subcorpora Obtained by Different Data Selection Methods

| V. Size | General-domain | VSM-Cos | PP-C | PP-ML | PP-MML | ED-LED | Com |
|---------|---------------|---------|------|-------|--------|--------|-----|
| Count (V) | 469,950 | 167,613 | 110,049 | 124,381 | 80,200 | 235,065 | 127,472 |
| Diff.% | | -64.33% | -76.58% | -73.53% | -82.93% | -49.98% | -72.88% |

As shown in Table 3, all the data selection methods result in substantial reductions (from -49.98% to -82.93%) of the vocabulary size. Among them, PP-MML results in the highest reduction (-82.93%). In other words, PP-MML can discard the most general-domain data than any other models. The reason may be that PP-MML considers not only the in-domain and general-domain term distribution but also the data in both languages. Global information and bilingual information are helpful in selecting in-domain data. On the contrary, ED-LED results in the lowest reduction (-49.98%), which shows its weakness in discarding the unrelated data. Because string-and-string methods can only reflect the relatedness between two sentences and are poor in predicting the domain- specificity in a large corpus. PP-C and PP-ML have similar reduction but higher than VSM-Cos. Although vector-based data selection is global[26], it only considers the co-occurrence of single words instead of n-grams like perplexity. We also found that the reductions of combination models are between PP-ML/MML and VSM-Cos. It shows that combing at sentence/model level may have no help in filtering. But some sentences in different subsets may be duplicated, which give high weights for related terms.

---

[25] As the statistics on both sides of a parallel corpus are monotonically related, we use the English side for analyzing.

**OOVs**

There are two main factors limit the performance of domain-specific SMT systems. One is unknown words, which depends upon the knowledge size of entire training data. The other one is ambiguity problem, which directly affects the quality of word alignment. Data selection is not a method to expend the knowledge of training data, but is a soft way to weight data with respect to the target domain. On the one hand, discarding parallel training data would push the probability distribution towards the target domain in case of ambiguity; on the other hand, losing these parallel data will increase the unknown word problem. In this section, we show how each model balance this double-edged-sword issue.

Taking the entire general-domain corpus as a baseline, we compare the OOV ratio of each model with it. Table 3-10 shows the OOV ratio of the test set with respect to the responding training data sets.

Table 3-10: OOVs of Pseudo In-Domain Subcorpora Obtained by Different Data Selection Methods

| OOVs | General-domain | VSM-Cos | PP-C | PP-ML | PP-MML | ED-LED | Com |
|------|---------------|---------|------|-------|--------|--------|-----|
| **OOV Ratio** | 0.01003 | 0.01465 | 0.01826 | 0.01465 | 0.02448 | 0.05519 | 0.02465 |
| **Diff.%** | | +46.06% | +82.05% | +46.06% | +144.07% | +450.25% | +145.76% |

As previously hypothesised, all the selection methods result in increases of OOVs (from +46.06% to +450.25%). ED-LED results in the highest increase and its OOVs are 5 times more than that of baseline. Although the threshold for ED-LED is 90%, this method still results in OOV problem. After analysis the sentences in its subset, we find that 1) many related sentences are cut down from the general-domain corpus; 2) there are a large amount of duplicated sentences but most of them are not domain related. VSM-Cos and PP-ML result in the lowest increase, however, PP-C and PP-MML have more OOVs than them. Combing the VSM-Cos, PP-MML and ED-LED together really help in reduce the OOVs.

---

[26] It makes the entire vocabularies as terms of vector.

**Overlapping**

If a data selection model brings only very little information that has not been selected by the baseline, its impact would be limited. Table 3-11 shows for each model the percentage of sentences selected by the model that are overlapping/unique to the other models. Overlapping: sentences occur in every subset obtained by different models. Unique: sentences do not occur in the data set obtained by any other of the models.

Table 3-11: Overlapping of Pseudo In-Domain Subcorpora Obtained by Different Data Selection Methods

| O/U | VSM-Cos | PP-C | PP-ML | PP-MML | ED-LED | Com |
|---|---|---|---|---|---|---|
| **Overlap** | 6.83% | 11.20% | 7.00% | 14.00% | 12.53% | 39.33% |
| **Unique** | 2.81% | 3.32% | 3.76% | 3.75% | 1.56% | 4.09% |

3.3.4 SECTION SUMMARY

In this section, we analyze the impacts of different data selection criteria on SMT domain adaptation. Empirical results reveal that the proposed model achieves a good performance in terms of robustness and effectiveness. We analyze the results from three different aspects:

- **Translation quality**: the results show a significant performance of the most methods especially for *i*TPB. Under the current size of datasets, considering more factors in similarity measuring may not benefit the translation quality.

- **Noises and OOVs**: it is a big challenge to balance them for single individual data selection model. However bilingual resources and combination methods are helpful to deal with this problem.

- **Robustness and effectiveness**: a real-life system should achieve a robust and effective performance with online strategy. Only *i*TPB obtained a consistently boosting performance.

Finally, we can draw a composite conclusion that (*a* > *b* means *a* is better than *b*):

$$iTPB > PPBased > Cos\text{-}IR > Baseline > FMS$$

## 3.4 LINGUISTICALLY-AUGMENTED DATA SELECTION

After investigating the data selection methods, we found that they all rely solely on the use of surface forms. The rationale being that the fact that these languages have a larger set of different words leads to sparsity problems, if the methods applied rely solely on surface forms. By reviewing some work in language modeling, researchers have looked at using linguistic information such as classes (Whittaker and Woodland, 1998), part-of-speech (PoS) tags (Heeman, 1999), stems and endings (Maučec et al., 2004). LMs built on different types of information (e.g. word and class-based) can then be interpolated to reduce perplexity (Maltese et al., 2001), especially for dealing with highly inflected languages. Therefore, we anticipate that this type of information could be useful as well for data selection.

This section explores the use of linguistic information for the selection of data to train language models. We depart from the state-of-the-art method in perplexity based data selection and extend it in order to use word-level linguistic units (i.e. lemmas, named entity categories and part-of-speech tags) instead of surface forms. We then present two methods that combine the different types of linguistic knowledge as well as the surface forms: 1) naïve selection of the top ranked sentences selected by each method; 2) linear interpolation of the datasets selected by the different methods. The following contents present detailed results and analysis for four languages with different levels of morphologic complexity (English, Spanish, Czech and Chinese). The interpolation-based combination outperforms the purely statistical baseline in all the scenarios, resulting in language models with lower perplexity. In relative terms the improvements are similar regardless of the language, with perplexity reductions achieved in the range 7.72% to 13.02%. In absolute terms the reduction is higher for languages with high type-token ratio (Chinese, 202.16) or rich morphology (Czech, 81.53) and lower for the remaining languages, Spanish (55.2) and English (34.43 on the English side of the same parallel dataset as for Czech and 61.90 on the same parallel dataset as for Spanish).

Furthermore, we apply this approach to select sentence pairs from large general-domain corpus to adapt translation models to target domain. We conduct experiments for English-Chinese language pairs. Although Chinese is non-highly inflected language, the results still show great improvement on translation quality (Toral et al., 2015).

### 3.4.1 METHODOLOGY

We use surface forms and different types of linguistic information at the word level in perplexity-based data selection. Our hypothesis is that ranking by perplexity on $n$-grams that represent linguistic patterns (rather than $n$-grams that represent surface forms, as done in previous approaches, Section 3.1 and 3.2) captures additional information, leading to better generalization and the ability to combat data sparseness, and thus may select valuable data that is not selected according solely to surface forms.

**Linguistic Information**

Specifically, we explore the use of three types of linguistic information at word level: lemmas, NE categories and PoS tags. All these three types of information group different surface forms into classes, and thus they reduce data sparsity and vocabulary size. They differ with respect to which surface forms are grouped together and the degree of vocabulary reduction that can be attained.

NE categories group together proper nouns that belong to the same semantic class (e.g. person, location, organization). The distributional properties of NEs (Toral and Way, 2014) (a huge amount of different instances and a very low number of occurrences per instance) lead to sparsity if surface forms that hold NEs were to be used for selection.

Lemmas group together word forms that share the same root. We hypothesis that the use of lemmas is especially useful for highly inflected languages, as in these languages the ratio of surface forms to lemmas is particularly high, and thus by grouping together different surface forms that share the same lemma, we are effectively reducing the sparsity.

Finally, PoS tags group together words that share the same grammatical function (e.g. adjectives, nouns, verbs). While PoS tags have weak predictive power as a result of the lack of lexical information, and thus they are expected to perform poorly on their own, they have been reported to be useful when used in combination with lexical models (Cussens et al., 2000).

By taking into account these types of information, we experiment with the following models:

- Forms (hereafter f) use surface forms. This model replicates the Moore-Lewis approach and provides the baseline in this study.
- Forms and NEs (hereafter fn) use surface forms, with the exception of any word (or word sequence) detected as a NE, which is substituted by its category (e.g. person, location, organization).
- Lemmas (hereafter l) use lemmas.
- Lemmas and NEs (hereafter ln) use lemmas, with the exception of any word (or word sequence) detected as a NE, which is substituted by its category.
- Tags (hereafter t) use PoS tags.
- Tags and NEs (hereafter tn) use PoS tags, with the exception of any word (or word sequence) detected as a NE, which is substituted by its category.

A sample sentence, according to each of these models, is shown in Figure 3-6. In this example the PoS tagset comes from Penn Treebank[27] while the NE tagset comes from Freeling[28] and is based on the EAGLES annotation guidelines[29]. We use these models to perform data selection both individually and combined. The following subsections detail both procedures, respectively.

---

f: America 's appallingly low savings rate .

fn: NP00G00 's appallingly low savings rate .

l: america 's appallingly low saving rate .

ln: NP00G00 's appallingly low saving rate .

t: NNP POS RB JJ NNS NN Fp

tn: NP00G00 POS RB JJ NNS NN Fp

Figure 3-6: Sample Sentence According to Each of The Models

**Data Selection Model**

For language model adaptation, the model is Moore-Lewis Eq. (2-7) on monolingual corpus. For SMT translation model adaptation, the model is modified Moore-Lewis Eq. (2-8) on parallel corpus. The difference of our approach with respect to cross-entropy on surface forms is that in our case the in-domain and general-domain corpora are pre-processed according to the linguistic model that we use (Figure 3-6). The LMs are built on these pre-processed versions of the corpora. We also use the pre-processed version of the general-domain corpus for the scoring phase. Once the sentences have been scored they are replaced with the corresponding sentences in the original corpus, keeping the ranking order. This allows the evaluation phase to be performed on subsets of the original corpus, even if they have been ranked according to a linguistically-motivated model.

**Combination of Models**

We also investigate the combination of the different individual models. We propose two combination methods, which we will refer to as naïve and advanced. The naïve combination (noted as c in the results) proceeds as follows. Given the sentences selected by all the individual models considered for a given threshold, we iterate through them following the ranking order (i.e. we traverse the first ranked sentence by each of the methods, then we proceed to the set of second best ranked sentences, and so forth). As we iterate through the sentences we keep a sentence if it has not been seen before, i.e. we keep all the distinct sentences. We stop the procedure when we have obtained a set of sentences whose size is that indicated by the threshold, i.e. the size of the set of sentences in the combination is the same as the size of the set of sentences produced by any of the methods.

The advanced combination (noted as ci in the results) proceeds similarly, the difference being that sentences are not kept in one unique set. Conversely, we consider as many sets as there are individual models. As we iterate through the sentences these are kept in the sets that correspond to their provenance model. As with the naïve approach, the procedure stops when the number of distinct sentences kept across the sets is the same as the number of sentences produced by any of the models. Finally, we build LMs for the sentences contained in each of these sets and perform linear interpolation on these LMs with a development set.

For a given threshold and set of individual models, both combination models contain the same sentences, the difference being that, while in the naïve combination these sentences are concatenated, in the advanced combination (multiple instances) of these sentences are placed in different LMs (according to the linguistic preprocessing model), and these LMs are given weights according to the interpolation.

### 3.4.2 EXPERIMENTAL SETUP

**Language Model Adaptation**

We carry out experiments for the following four languages: English, Spanish, Czech and Chinese. Although the experiments are run on each language independently from the others, we have used parallel corpora for some of these language pairs (English-Spanish and English-Czech). By running experiments for two different languages using parallel corpus data, we can extract more meaningful conclusions from the comparison of the results. All the corpora used in this study are de-duplicated at sentence level.

For both English-Spanish and English-Czech we use corpora from the WMT translation task series[30]. For both language pairs the in-domain corpus is News Commentary version 8 (hereafter NC), while the general-domain is United Nations[31] (Eisele and Chen, 2010) (hereafter UN) for English-Spanish and CzEng 1.0[32] (Bojar et al., 2012) for English-Czech. For both language pairs we use newstest2012 (test set for WMT 2012) as the development set and newstest2013 (test set for WMT 2013) as the test set.

For Chinese the in-domain corpus is the Chinese side of the News Magazine Corpus (LDC2005T10)[33], while the general-domain data is collected from the UM-Corpus[34] (Tian et al., 2014), CWMT News and the Sci-Tech corpus. Two random sets of 2,000 sentences each are taken out of the in-domain data to be used as development and test sets.

Table 3-12 details the general-domain (referred to as out) and in-domain (referred to as in) corpora used for each language, including the number of sentences and words, the vocabulary size and the type-token ratio. Type-token ratios are similar for English and Spanish both for in-domain (.01441 and .01714) and general-domain corpora (.00193 and .00173). As expected due to its highly inflected nature, ratios are higher for Czech when compared to the equivalent data in English, both for in-domain (.04372 vs. .01775) and general-domain corpora (.01114 vs. .00849). Finally, the ratios for Chinese are rather high at .02161 and .04079 for in-domain and general-domain corpora, respectively. This has to do with the considerably larger set of characters of this logogram-based language when compared to the other languages of this study, whose writing systems are based on alphabets.

---

[30] http://www.statmt.org/wmt14/translation-task.html.

[31] http://www.uncorpora.org/.

[32] http://ufal.mff.cuni.cz/czeng/.

[33] http://catalog.ldc.upenn.edu/LDC2005T10.

[34] http://nlp2ct.cis.umac.mo/um-corpus/.

Table 3-12: Detailed Information About The Corpora Used. Languages Are Referred to As EN (English), ES (Spanish), CS (Czech) and ZH (Chinese).

| Lang, corpus | Sentences | Words | Vocabulary | Ratio |
|---|---|---|---|---|
| **EN, out** | 11,196,913 | 320,065,223 | 618,775 | .00193 |
| **ES, out** | | 366,174,710 | 631,959 | .00173 |
| **EN, in** | 173,950 | 4,515,562 | 65,064 | .01441 |
| **ES, in** | | 5,112,490 | 87,622 | .01714 |
| **EN, out** | 10,276,812 | 164,622,981 | 1,398,519 | .00849 |
| **CS, out** | | 147,061,482 | 1,638,842 | .01114 |
| **EN, in** | 139,325 | 3,435,449 | 60,983 | .01775 |
| **CS, in** | | 3,190,502 | 139,480 | .04372 |
| **ZH, out** | 3,422,788 | 56,986,145 | 2,324,258 | .04079 |
| **ZH, in** | 270,623 | 9,753,911 | 210,820 | .02161 |

In order to perform data selection using the linguistically-augmented models (in Section 3.4.1), these corpora have been processed with the following NLP tools:

- For the English-Spanish data, we have used Freeling 3.0 (Padró and Stanilovsky, 2012) to perform lemmatization, PoS tagging and NE recognition. These corpora are tokenized and truecased using the corresponding scripts from the Moses toolkit (Koehn, 2007).

- The English-Czech parallel data has been processed by the TectoMT framework (Popel and Žabokrtský, 2010) using the following pipeline. The Czech language side was tokenized by the Czech TectoMT tokenizer and PoS-tagged and lemmatized (technical sues of the lemmas produced by the tagger were omitted) by the Featurama tagger[35]. The Czech NEs were labelled by the TectoMT component based on the NE recognizer of Strakova (Straková et al., 2013). Truecasing was done by changing the case of the first character of each word to correspond with the case of the first character of its lemma. The English language side was tokenized by the English TectoMT tokenizer, PoS-tagged by the Morce tagger (Hajič et al., 2007) and lemmatized using the rule-based lemmatize by Popel (Popel, 2009). The English NEs were labelled

---

[35] http://featurama.sourceforge.net/.

with the Stanford NE recognizer[36]. Truecasing was done the same way as on the Czech side.

- The Chinese corpora have been processed (word segmentation, PoS tagging and NE recognition) with the Stanford CoreNLP toolkit[37]. From this toolkit we have used a CRF-based word segmenter (Tseng, 2005; Chang et al., 2008), a maximum entropy PoS tagger (Toutanova et al., 2003) and a CRF-based NE recognizer (Finkel et al., 2005) with built-in Chinese models.

Due to the different nature of the languages considered and to the processing tools used, not all the linguistic models that have been introduced (in Section 3.4.1) have been used for all the four languages considered. Table 3-13 shows the individual models that have been used for each of the languages. The model that uses PoS tags only (t) is not used for Spanish-English nor for Czech-English as the corpora processing pipeline contains NE tags already integrated with the PoS-tagged output. Models that use lemmas (l and ln) are not used for Chinese as this linguistic concept does not apply to this language.

All the LMs used in the experiments are built with IRSTLM 5.80.01 (Federico et al., 2008), they consider n-grams up to order 4 and they are smoothed using a simplified version of the modified Kneser-Ney method (Chen and Goodman, 1996). IRSTLM is also used to compute perplexities. Linear interpolation of LMs is carried out with SRILM (Stolcke, 2002) via the Moses toolkit.[38]

Table 3-13: Individual Models Used for Each Language

| Language | Models | | | | | |
|---|---|---|---|---|---|---|
| | f | fn | l | ln | t | tn |
| ES-EN | √ | √ | √ | √ | | √ |
| CS-EN | √ | √ | √ | √ | | √ |
| ZH | √ | √ | | | √ | √ |

---

[36] http://nlp.stanford.edu/ner/.

[37] http://nlp.stanford.edu/software/.

[38] https://github.com/moses-smt/mosesdecoder/blob/RELEASE-2.1/scripts/ems/support/interpolate-lm.perl

**Translation Model Adaptation for SMT**

We also explore this method on parallel corpus to adapt Chinese-English SMT systems to target domain.

As shown in Table 3-14, two corpora are needed for the domain adaptation task. General-domain parallel corpus combined with various general-domain corpora: CWMT2013[39], UM-Corpus[40], News Magazine (LDC2005T10)[41] etc. In-domain parallel corpus, dev set, test set are the official corpus of IWSLT2014 TED Talk task[42]. We use parallel corpora for TM training and the target side for LM training.

Table 3-14: Corpora Statistics

| Data Set | Sentences | Ave. Len. |
|---|---|---|
| Test Set | 1,570 | 26.54/23.41 |
| Dev Set | 887 | 26.47/23.24 |
| In-domain | 177,477 | 26.47/23.58 |
| Training Set | 10,021,162 | 23.02/21.36 |

All processes are same to language model adaptation. We apply both individual and combined methods for this task.

### 3.4.3 RESULTS AND DISCUSSIONS

We report our results in terms of the perplexities obtained on a test set by LMs built on different subsets of the data selected by each of the models. These subsets correspond to different thresholds, i.e. percentages of sentences selected from the general-domain corpus. These are the first $\frac{1}{32}$ ranked sentences, $\frac{1}{16}$, $\frac{1}{8}$, $\frac{1}{4}$, $\frac{1}{2}$ and 1.

---

[39] Available at http://www.liip.cn/cwmt2013/.

[40] Available at http://nlp2ct.cis.umac.mo/um-corpus/.

[41] Available at https://catalog.ldc.upenn.edu/LDC2005T10.

[42] Available at http://workshop2014.iwslt.org/.

Figures 3-7, 3-8, 3-9, 3-10 and 3-11 show test data perplexities obtained by LMs built on data selected by each model on different subsets of the general-domain corpus, for English (English-Spanish dataset), Spanish, English (English-Czech dataset), Czech and Chinese, respectively. In each figure, the x-axis indicates the percentage of the data selected (as $\frac{1}{x}$) while the y-axis indicates the perplexity value.

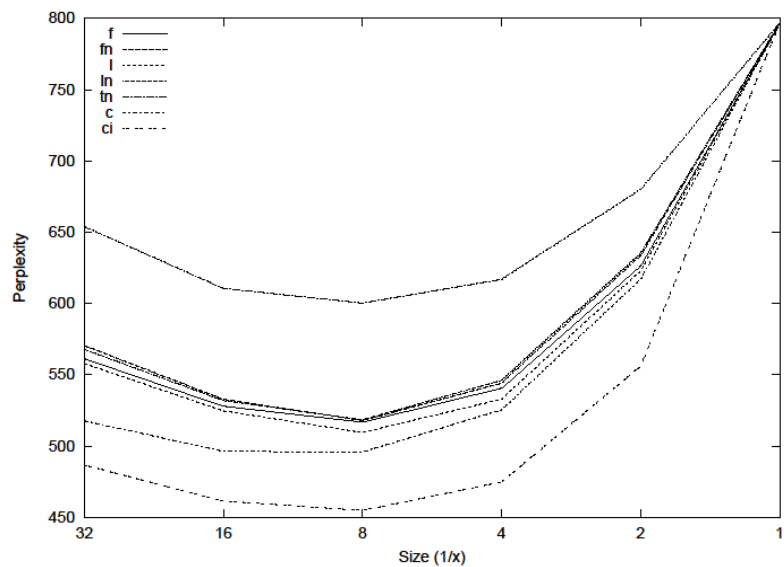

Figure 3-7: Perplexities Obtained by the Different Models, English (English-Spanish)

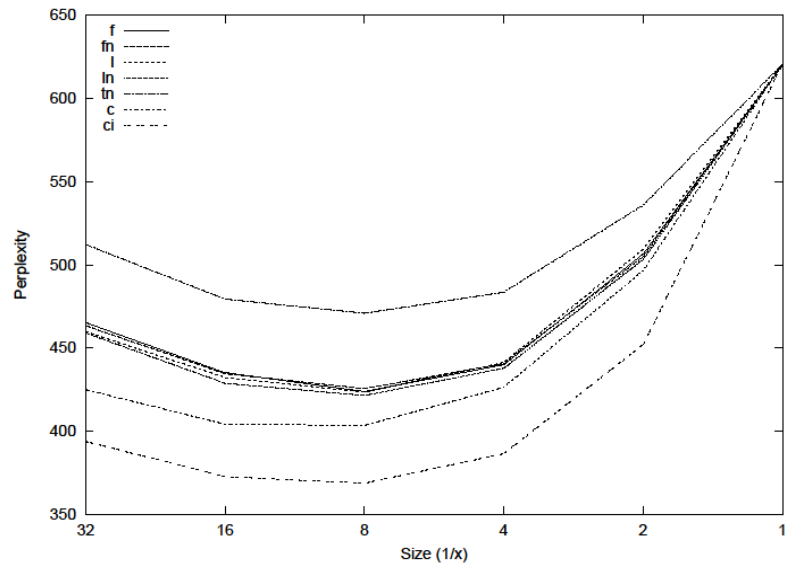

Figure 3-8: Perplexities Obtained by the Different Models, Spanish (English-Spanish)

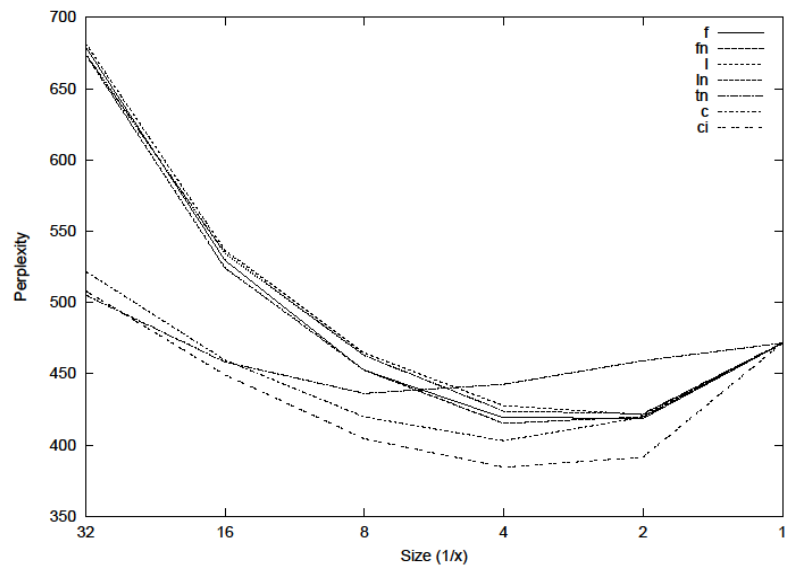

Figure 3-9: Perplexities Obtained by the Different Models, English (English-Czech)

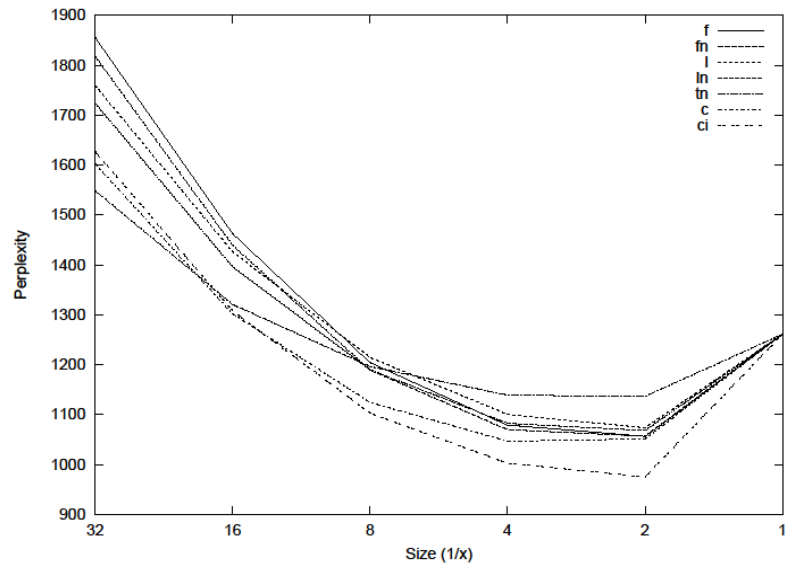

Figure 3-10: Perplexities Obtained by The Different Models, Czech (English-Czech)

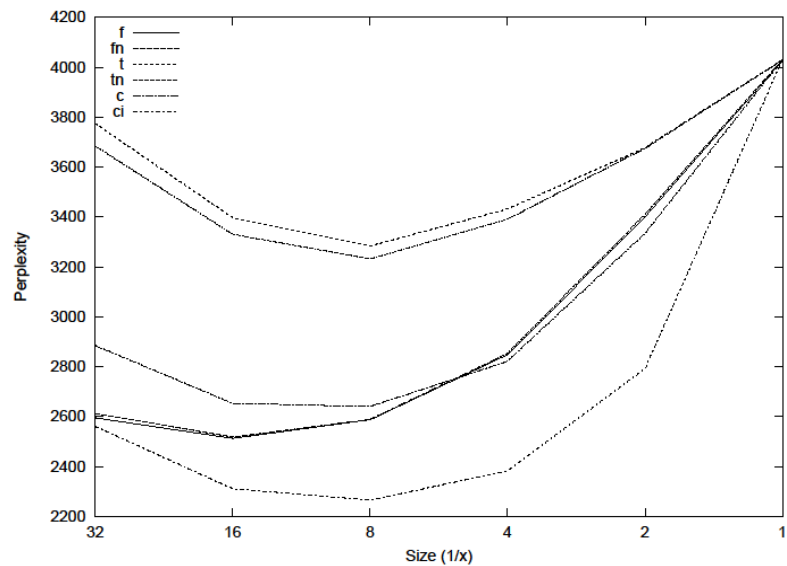

Figure 3-11: Perplexities Obtained by The Different Models, Chinese

The trends observed regarding the different models are common across all the figures. All the individual linguistic models, except for the ones that use PoS tags, perform similarly to the baseline. The models that use tags perform slightly worse, as expected, due to their lack of lexical information. Both combination models outperform the baseline and the individual linguistic models, the only exception to this being the naïve combination performing worse than individual models f and fn for Chinese. The advanced combination (linear interpolation) outperforms the naïve combination in all the scenarios.

Table 3-15 gives a detailed account of the results obtained by each individual model for the different languages considered. We show the scores by each model for the threshold for which the baseline obtains its best result ($\frac{1}{8}$ for English-Spanish, $\frac{1}{2}$ for English-Czech and $\frac{1}{16}$ for Chinese). For the baseline we show the absolute perplexity, while for the linguistic models we show relative values compared to the baseline (as percentages).

Table 3-15: Results for The Different Individual Models. The Model That Obtains The Lowest Perplexity For Each Language Is Shown in Bold

| Language | Models | | | | | |
|---|---|---|---|---|---|---|
| | f | fn | l | ln | t | tn |
| EN | 516.68 | 0.24% | **-1.41%** | 0.33% | | 16.18% |
| ES | 423.88 | -0.39% | -0.50% | **-0.97%** | | 10.65% |
| EN | 418.84 | -0.23% | **0.35%** | 0.49% | | 3.90% |
| CS | 1056.18 | -0.03% | **1.60%** | 1.06% | | 7.52% |
| ZH | 2512.45 | **0.20%** | | | 35.21% | 32.54% |

As previously seen in Figures, the baselines and the individual models that use lexical information (fn, l and ln) obtain very similar scores, while models that use tags (t and tn) lag behind. Different individual models get the best result for different languages (l for English in English-Spanish, ln for Spanish, fn for Czech and English in English-Czech and f for Chinese), although the differences being so small they may be considered non-significant.

Table 3-16 presents the results for the combination models and compares them to the baseline. The naïve combination outperforms the baseline in all the scenarios (4.05% and 4.85% lower perplexities in English-Spanish and 3.78% and 0.49% in English-Czech) except for Chinese (5.38% higher perplexity). The advanced

combination outperforms the baselines in all the scenarios, the relative improvements being in the range 7.72% to 13.02% depending on the language. In absolute terms the reduction is higher for languages with high type-token ratio (Chinese, 202.16) and rich morphology (Czech, 81.53) and lower for Spanish (55.2) and English (34.43 on the same dataset as Czech and 61.90 on the same dataset as Spanish).

Table 3-16: Results for The Different Combination Models

| Language | f | c | ci |
|----------|---------|--------|---------|
| EN | 516.68 | -4.05% | -11.98% |
| ES | 423.88 | -4.85% | -13.02% |
| EN | 418.84 | -3.78% | -8.22% |
| CS | 1056.18 | -0.49% | -7.72% |
| ZH | 2512.45 | 5.60% | -8.05% |

For translation model adaptation, all systems are log-linearly interpolated with the in-domain model to further improve the adapted model. To English-Chinese corpus, we have to use the linguistic information they both contain: surface, POS and NER.

- **Baseline**: the in-domain baseline (IC-Baseline) and general-domain baseline (GC-Baseline) are respectively trained on in-domain corpus and general corpus. GI-Baseline is trained on all above data.

- **Individual Model**: surface form based (f), POS based (t), surface and named entity based (fn), surface and POS (ft).

- **Combined Model**: corpus level (Comb-C) and model level (Comb-M).

We investigate $K$={25, 50, 75}% of ranked general-domain data as pseudo in-domain corpus for SMT training. The results are shown in Table 3-17. Although Chinese is inflected-poor language, fn and ft still can improve GI baseline by nearly 1 BLEU. t perform poorly due to lack of lexical information. f is the famous MML method only considering surface form. ft does slightly better than f by 0.44 point, which indicates replacing some non-NN and non-VV word by its POS tags can reduce the sparsity and keep the language style of the in-domain sentence. fn performs no better than f: although NER tags can reduce the surface variants, but these name words (location, person, organization) are usually very important to define the domain. All combination methods do better than an individual model (from +0.64 to +0.11 BLEU), because they select the top sentences retrieved by each approach. Combination method may success the advantages of linguistic information (reducing sparsity and learn language style). Model combination is better than simply combine different data, because surface-based and POS-based sub-corpora may need different weights. Western languages such as English, French and German may have better performance on our method due to the high-flatted.

Table 3-17: Translation Model Adaptation Results for The Different Models

| System | 25% | 50% | 75% |
|---|---|---|---|
| **GC-Baseline** | 39.52 | | |
| **IC-Baseline** | 10.40 | | |
| **GI-Baseline** | **40.20** | | |
| **f** | 31.91 (-8.29) | 38.83 (-1.37) | 41.37 (+1.17) |
| **t** | 21.20 (-19.00) | 27.90 (-12.30) | 27.90 (-12.30) |
| **fn** | 31.93 (-8.27) | 37.86 (-2.34) | 40.93 (+0.73) |
| **ft** | 30.00 (-10.20) | 38.74 (-1.46) | 41.81 (+1.61) |
| **Comb-C** | 33.01 (-7.19) | 39.07 (-1.13) | 41.92 (+1.72) |
| **Comb-M** | 32.74 (-7.46) | 38.95 (-1.25) | 42.01 (+1.81) |

### 3.4.4 SECTION SUMMARY

This section explores the use of different types of linguistic information at the word level (lemmas, NEs and PoS tags) for the task of training data selection for LMs following the perplexity-based approach. By using these types of information, we have introduced five linguistically motivated models. We have also presented two methods to combine the individual linguistic models as well as the baseline (surface forms), a simple selection of top ranked sentences selected by each method and a linear interpolation of LMs built on the data selected by the different methods.

The experiments are carried out on four languages with different levels of morphological complexity (English, Spanish, Czech and Chinese). Our combination model based on linear interpolation outperforms the purely statistical baseline in all the scenarios, resulting in language models with lower perplexity. In relative terms the improvements are similar regardless of the language, with perplexity reductions achieved in the range 7.72% to 13.02%. In absolute terms the reduction is higher for languages with high type-token ratio (Chinese, 202.16) or rich morphology (Czech, 81.53) and lower for the remaining 405 languages, Spanish (55.2) and English (34.43 on the same dataset as Czech and 61.90 on the same dataset as Spanish).

# CHAPTER 4: DOMAIN FOCUSED WEB-CRAWLING FOR SMT DOMAIN ADAPTATION

*The web is immense, free and available by mouse-click. It contains hundreds of billions of words of text and can be used for all manner of language research.*

-- Adam Kilgarriff and Gregory Grefenstette, 2003

In order to reduce the OOVs, we explore to acquire additional resources to fix the scarcity of in-domain corpora. We use domain-focused web-crawling methods to obtain in-domain monolingual/parallel data from the Internet and then supplement and adjust data distribution in training corpora. We firstly present a combination method named TQDL to improve the cross-language document alignment performance. Then, we present a perplexity-based filtering method to further reduce the noise in acquired corpus. In order to explore the best way of utilizing web-crawled data, we combine them with existing data at corpus level, alignment level and model level.

## 4.1 INTEGRATED MODELS FOR CROSS-LANGUAGE DOCUMENT RETRIEVAL

This section proposed an integrated approach for Cross-Language Information Retrieval (CLIR), which integrated with four statistical models: **T**ranslation model, **Q**uery generation model, **D**ocument retrieval model and **L**ength Filter model. Given a certain document in the source language, it will be translated into the target language of the statistical machine translation model. The query generation model then selects the most relevant words in the translated version of the document as a query. Instead of retrieving all the target documents with the query, the length-based model can help to filter out a large amount of irrelevant candidates according to their length information. Finally, the left documents in the target language are scored by the document searching model, which mainly computes the similarities between query and document.

Different from the traditional parallel corpora-based model which relies on IBM algorithm, we divided our CLIR model into four independent parts, but all work together to deal with the term disambiguation, query generation and document retrieval. Besides, the TQDL method can efficiently solve the problem of translation ambiguity and query expansion for disambiguation, which are the big issues in Cross-Language Information Retrieval. Another contribution is the length filter, which are trained from a parallel corpus according to the ratio of length between two languages. This cannot only improve the recall value due to filtering out lots of useless documents dynamically, but also increase the efficiency in a smaller search space. Therefore, the precision can be improved but not at the cost of recall.

In order to evaluate the retrieval performance of the proposed model on cross-languages document retrieval, a number of experiments have been conducted on different settings. Firstly, the Europarl corpus which is the collection of parallel texts in 11 languages from the proceedings of the European Parliament was used for evaluation. And we tested the models extensively to the case that: the lengths of texts are uneven and some of them may have similar contents under the same topic, because it is hard to be distinguished and make full use of the resources.

After comparing different strategies, the experimental results show a significant performance of the method. The precision is normally above 90% by using a larger query size. The length-based filter plays a very important role in improving the F-measure and optimizing efficiency.

This fully illustrates the discrimination power of the proposed method. It is of a great significance to both cross-language searching on the Internet and the parallel corpus producing for statistical machine translation systems. In the future work, the TQDL system will be evaluated for Chinese language, which is a big changing and more meaningful to CLIR.

### 4.1.1 PROPOSED INTEGRATED MODELS

The approach relies on four models: translation model which generates the most probable translation of source documents; query generation model which determines what words in a document might be more favorable to use in a query; length filter model dynamically creates a subset of candidates for retrieval according to the length information; and document searching model, which evaluates the similarity between a given query and each document in the target document set. The workflow of the approach for CLIR is shown in Figure 4-1.

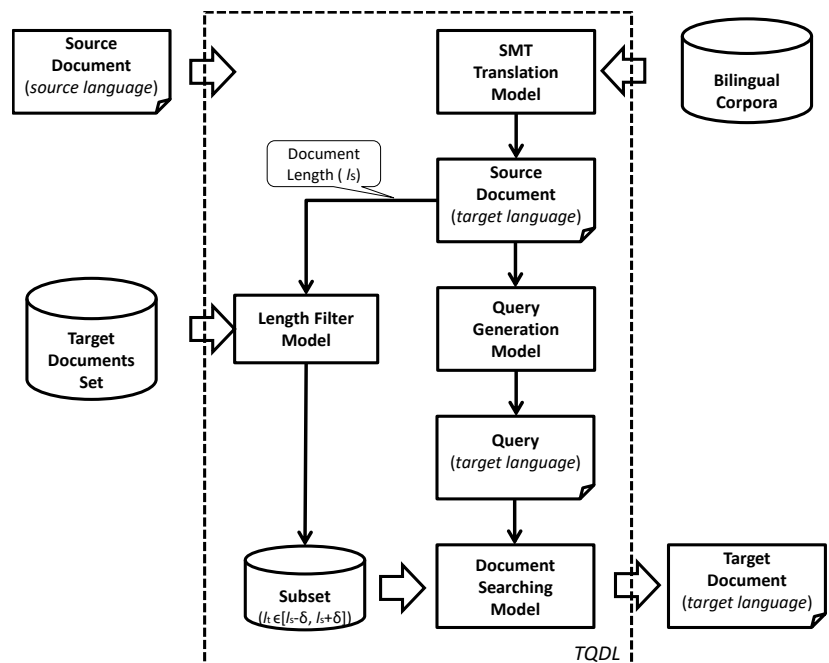

Figure 4-1: The Proposed Approach for CLIR

**Translation Model**

Currently, the good performing statistical machine translation systems are based on phrase-based models which translate small word sequences at a time. Generally speaking, translation model is common for contiguous sequences of words to translate as a whole. Phrasal translation is certainly significant for CLIR (Ballesteros and Croft, 1997). It can do a good job in dealing with term disambiguation.

In this work, documents are translated using the translation model provided by Moses, where the log-linear model is considered for training the phrase-based system models (Och and Ney, 2002), and is represented as:

$$p(e_1^I \mid f_1^J) = \frac{\exp(\sum_{m=1}^{M} \lambda_m h_m(e_1^I, f_1^J))}{\sum_{e_1'^I} \exp(\sum_{m=1}^{M} \lambda_m h_m(e_1'^I, f_1^J))} \qquad (4\text{-}1)$$

where $h_m$ indicates a set of different models, $\lambda_m$ means the scaling factors, and the denominator can be ignored during the maximization process. The most important models in Eq. (4-1) normally are phrase-based models which are carried out at source to target and target to source directions. The source document will maximize the equation to generate the translation, including the words most likely to occur in the target document set.

**Query Generation Model**

After translating the source document into the target language of the translation model, the system should select a certain amount of words as a query for searching instead of using the whole translated text. It is for two reasons, one is computational cost, and the other is that the unimportant words will degrade the similarity score. This is also the reason why it often responses nothing from the search engines on the Internet when we choose a whole text as a query.

In this section, we apply a classical algorithm which is commonly used by the search engines as a central tool in scoring and ranking relevance of a document given a user query. Term Frequency–Inverse Document Frequency (TF-IDF) calculates the values for each word in a document through an inverse proportion of the frequency of the word in a particular document to the percentage of documents where the word appears (Ramos, 2003). Given a document collection $D$, a word $w$, and an individual document $d \in D$, we calculate

$$P(w,d) = f(w,d) \times \log \frac{|D|}{f(w,D)} \qquad (4\text{-}2)$$

where $f(w, d)$ denotes the number of times $w$ that appears in $d$, $|D|$ is the size of the corpus, and $f(w,D)$ indicates the number of documents in which $w$ appears in $D$ (Berger et al., 2000).

In implementation, if $w$ is an Out-of-Vocabulary term (OOV), the denominator $f(w,D)$ becomes zero, and will be problematic (divided by zero). Thus, our model makes *log* $(|D|/ f(w,D))=1$ (*IDF*=1) when this situation occurs. Additionally, a list of stop-words in the target language is also used in query generation to remove the words which are high frequency but less discrimination power. Numbers are also treated as useful terms in our model, which also play an important role in distinguishing the documents. Finally, after evaluating and ranking all the words in a document by their scores, we take a portion of the (*n*-best) words for constructing the query and are guided by:

$$Size_q = [\lambda_{percent} \times Len_d] \qquad (4\text{-}3)$$

$Size_q$ is the number of terms. $\lambda_{percent}$ is the percentage and is manually defined, which determines the $Size_q$ according to $Len_d$, the length of the document. The model uses the first $Size_q$-th words as the query. In another word, the larger document, the more words are selected as the query.

**Document Retrieval Model**

In order to use the generated query for retrieving documents, the core algorithm of the document retrieval model is derived from the Vector Space Model (VSM). Our system takes this model to calculate the similarity of each indexed document according to the input query. The final scoring formula is given by:

$$Score(q,d) = coord(q,d) \sum_{t \, in \, q} tf(t,d) \times idf(t) \times bst \times norm(t,d) \qquad (4\text{-}4)$$

where $tf(t,d)$ is the term frequency factor for term $t$ in document $d$, $idf(t)$ is the inverse document frequency of term $t$, while $coord(q,d)$ is frequency of all the terms in query occur in a document. *bst* is a weight for each term in the query. *Norm(t,d)* encapsulates a few (indexing time) boost and length factors, for instance, weights for each document and field. As a summary, many factors that could affect the overall score are taken into account in this model.

**Length Filter Model**

In order to obtain a suitable filter, we firstly analyzed the golden data[43] of ACL Workshop on SMT 2011, which includes Spanish, English, and French, German and Czech 5 languages and 10 language pairs. English-Spanish language pair was used for analyzing and the data of the corpus are summarizes in Table 4-1.

Table 4-1: Analytical Data of Corpus of ACL Workshop on SMT 2011

| Dataset | Size of corpus | | |
|---|---|---|---|
| | No. of Sentences | No. of Characters | Ave. No. Characters |
| **English** | 3,003 | 74,753 | 25 |
| **Spanish** | 3,003 | 79,426 | 26 |

Figure 4-2 plots the distribution of word number in each aligned sentences. $l_t$ is the length of English sentence while $l_s$ is the length of sentence in Spanish. So the expectation is c= $E$ ($l_t/l_s$) =1.0073, with the correlation $R^2 = 0.9157$. This shows that the data points are not substantially scatter in the plot and many data points are along with the regression line. Therefore, it is suitable to design a filter based on length ratio.

---

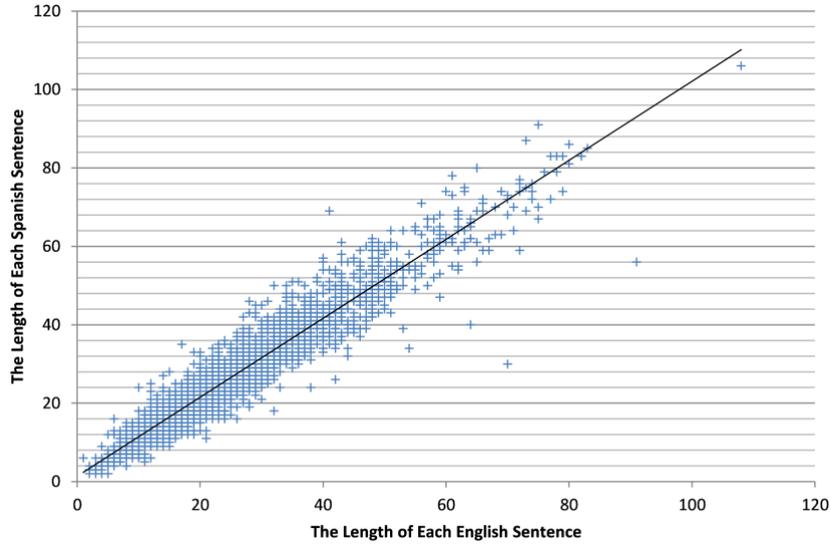

Figure 4-2: The Length Ratio of Spanish-English Sentences

To obtain an estimated length-threshold ($\delta$) for filter model, the function $\delta$ ($l_s$, $l_t$) can be designed as follows:

$$\delta(l_s, l_t) = \frac{|\, l_t - l_s \,|}{l_s} \qquad (4\text{-}5)$$

where $l_s$ and $l_t$ respectively stand for the length of a certain aligned sentence in the corpus we used. Finally, we got the average $\delta$ of around **0.15**. In implementation, we choose *4δ* instead of *δ* to avoid some abnormal cases, where the right document would be discarded by the filter.

Filter $F$ describes the relation between bilingual sentences based on the length ratio. Since western languages are similar in terms of word representation, the length ratio can be simply estimated as a 1:1. Given a certain document in source language, $F$ can collect a subset for retrieval according to the average length ratio. So $F$ is designed as follows:

$$F = \begin{cases} 1, length_t \in C \\ 0, length_t \notin C \end{cases}, C = [length_s - \delta, length_s + \delta] \qquad (4\text{-}6)$$

where $length_s$ is the length of source document, and $length_t$ is the length of target document. $\delta$ is an average threshold obtained through Eq. (4-5), $C$ is a confidence interval. If $length_t$ is included in $C$, $F$ is 1, which has a chance to be retrieved, otherwise set as 0, which will be skipped during searching.

### 4.1.2 EXPERIMENTAL SETUP

In order to evaluate the retrieval performance of the proposed model on text of cross languages, we use the Europarl corpus[44] which is the collection of parallel texts in 11 languages from the proceedings of the European Parliament (Koehn, 2005). The corpus is commonly used for the construction and evaluation of statistical machine translation. The corpus consists of spoken records held at the European Parliament and are labeled with corresponding IDs (e.g. <CHAPTER $id$>, <SPEAKER $id$>). The corpus is quite suitable for use in training the proposed probabilistic models between different language pairs (e.g. English-Spanish, English-French, English-German, etc.), as well as for evaluating retrieval performance of the system.

The datasets (training and test set) are collected for this evaluation. The chapters from April 1998 to October 2006 were used as a training set for model construction, both for training the Language Model (LM) and Translation Model (TM). While the chapters from April 1996 to March 1998 were considered as the testing set for evaluating the performance of the model. Besides, each paragraph (split by <SPEAKER $id$> label) is treated as a document, for dealing with the low discrimination power. The analytical data of the corpus are presented in Table 4-2. The Test Set contains 23,342 documents, of which length is 309 in average. Actually 30% of documents are much more or less than the average number. Table 4-1 summarizes the number of documents, sentences, words and the average word number of each document.

---

[44] Available online at http://www.statmt.org/europarl/.

Table 4-2: Analytical Data of Corpus

| Dataset | Size of corpus | | | |
|---|---|---|---|---|
| | Documents | Sentences | Words | Ave. words in document |
| Training Set | 2,900 | 1,902,050 | 23,411,545 | 50 |
| Test Set | 23,342 | 80,000 | 7,217,827 | 309 |

The most frequent and basic evaluation metrics for information retrieval are precision and recall, which are defined as Eq. (1-11), (1-12) and (1-13).

The probabilistic LMs are constructed on monolingual corpora by using the SRILM (Stolcke et al., 2002). We use GIZA++ (Och and Ney, 2003) to train the word alignment models for different pairs of languages of the Europarl corpus, and the phrase pairs that are consistent with the word alignment are extracted. For constructing the phrase-based statistical machine translation model, we use the open source Moses (Koehn et al., 2007) toolkit, and the translation model is trained based on the log-linear model, as given in Equation 2-3. A 5-gram LM is trained on Spanish data with the SRILM toolkits.

Once LM and TM have been obtained, we evaluate the proposed method with the following steps:

- The source documents are first translated into target language using the constructed translation model.
- The words candidates are computed and ranked based on a TF-IDF algorithm and the n-best words candidates then are selected to form the query based on Equation 5-2 and 5-3.
- All the target documents are stored and indexed using Apache Lucene[45]   as our default search engine.
- In retrieval, target documents are scored and ranked by using the document retrieval model to return the list of most related documents with Equation 4-4.

---

[45]  Available at http://lucene.apache.org.

A number of experiments have been performed to investigate our proposed method on different settings. In order to evaluate the performance of the three independent models, we firstly conducted experiments to test them respectively before whole the TQDL platform. The performance of the method is evaluated in terms of the ***average precision***, that is, how often the target document is included within the first N-best candidate documents when retrieved.

**Monolingual Environment Information Retrieval**

In this experiment, we want to evaluate the performance of the proposed system to retrieve documents (monolingual environment) given the query. It supposes that the translations of source documents are available, and the step to obtain the translation for the input document can therefore be neglected. Under such assumptions, the CLIR problem can be treated as normal IR in monolingual environment. In conducting the experiment, we used all of the source documents of Test Set. The empirical results based on different configurations are presented in Table 4-3, where the first column gives the number of documents returned against the number of words/terms used as the query.

Table 4-3: The Average Precision in Monolingual Environment

| Retrieved Documents (N-Best) | Query Size ($Size_q$ in %) | | | | | | |
|:---:|:---:|:---:|:---:|:---:|:---:|:---:|:---:|
| | **2** | **4** | **8** | **10** | **14** | **18** | **20** |
| **1** | 0.794 | 0.910 | 0.993 | 0.989 | 0.986 | 1.000 | 0.989 |
| **5** | 0.921 | 0.964 | 1.000 | 1.000 | 1.000 | 1.000 | 0.996 |
| **10** | 0.942 | 0.971 | 1.000 | 1.000 | 1.000 | 1.000 | 0.996 |
| **20** | 0.946 | 0.978 | 1.000 | 1.000 | 1.000 | 1.000 | 0.996 |

The results show that the proposed method gives very high retrieval accuracy, with precision of 100%, when the top 18% of the words are used as the query. In case of taking the top 5 candidates of documents, the approach can always achieve a 100% of retrieval accuracy with query sizes between 8% and 18%. This fully illustrates the effectiveness of the retrieval model.

**Translation Quality**

The overall retrieval performance of the system will be affected by the quality of translation. In order to have an idea the performance of the translation model we built, we employ the commonly used evaluation metric, BLEU, for such measure. The BLEU (Bilingual Evaluation Understudy) is a classical automatic evaluation method for the translation quality of an MT system (Papineni et al., 2002). In this evaluation, the translation model is created using the parallel corpus, as described in Section 4.1.2. We use another 5,000 sentences from the TestSet1 for evaluation[46].The BLEU value, we obtained, is **32.08**. The result is higher than that of the results reported by Koehn in his work (Koehn, 2005), of which the BLEU score is **30.1** for the same language pair we used in Europarl corpora. Although we did not use exactly the same data for constructing the translation model, the value of **30.1** was presented as a baseline of the English-Spanish translation quality in Europarl corpora.

The BLEU score shows that our translation model performs very well, due to the large number of the training data we used and the pre-processing tasks we designed for cleaning the data. On the other hand, it reveals that the translation quality of our model is good.

**TQDL without Filter for CLIR**

In this section, the proposed model without length filter model is tested. Table 4-4 presents the F-measure given by TQDL system without length filter model. As illustrated, the it can only achieve up to 94.7%, counting that the desired document is returned as the most relevant document among the candidates. Although it has achieved a very good performance in the experiments, the 6.6% of documents have been discarded in the pre-processing.

Table 4-4: The F-measure of Our System without Length Filter Model

| Retrieved Documents | Query Size ($Size_q$ in %) |
|---|---|

---

[46]  See http://www.statmt.org/wmt09/baseline.html for a detailed description of MOSES evaluation options.

| (*N*-Best) | 2.0 | 4.0 | 6.0 | 8.0 | 10.0 |
|:---:|:---:|:---:|:---:|:---:|:---:|
| 1 | 0.905 | 0.943 | 0.942 | **0.947** | 0.941 |
| 2 | 0.922 | 0.949 | 0.949 | 0.953 | 0.950 |
| 5 | 0.932 | 0.950 | 0.953 | 0.963 | 0.960 |
| 10 | 0.936 | 0.954 | 0.960 | 0.968 | 0.971 |
| 20 | 0.941 | 0.958 | 0.974 | 0.979 | 0.981 |

To investigate the changes of the performance with removing abnormal documents (too lager or too small), query size $Size_q$ was set as a constant value (8.0%)**,** which can achieve the best precision as shown in Table 4-4. We believed that the abnormal document is the main obstacle to develop the performance of the system. Therefore, we removed the documents, of which length are out of a certain threshold.

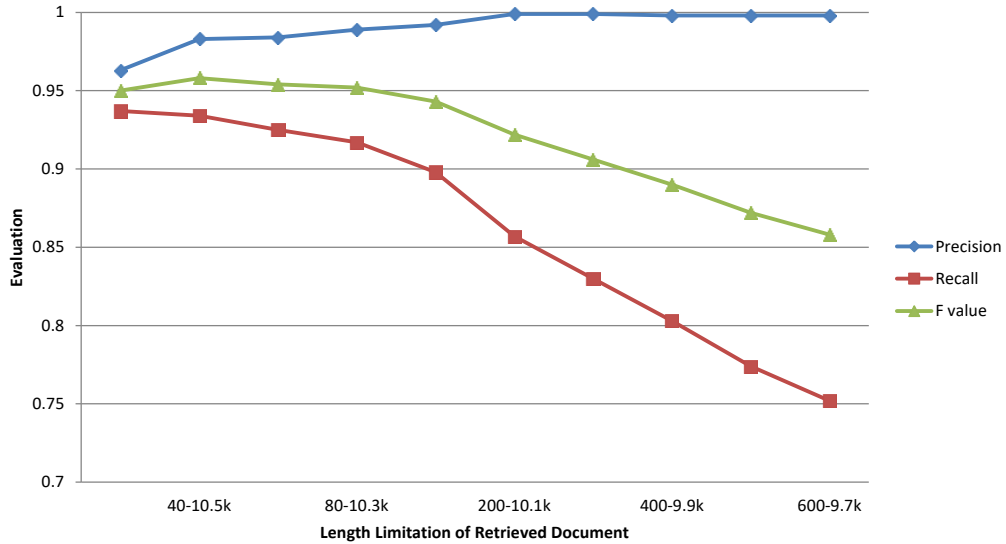

Figure 4-3: The Changes of Evaluation When Removing Data

Figure 4-3 plots the variations of *P*, *R* and *F* with the length scope increasing. As we expected, the precision increase when the more abnormal documents are discarded from the dataset. However, the recall declines sharply, which also lead to the falling of *F*-measure. When the precision is closed to **100%**, nearly **15%** documents are removed from the dataset. So the high precision is often at the cost of reducing the recall rate. *F*-measure is only 95% at its top, so it is hard to improve the performance of CLIR using traditional methods.

**TQDL with Filter for CLIR**

In order to obtain a higher retrieval rate, our model has been improved from different points. Firstly, we generate the query with dynamic size, which can do better in dealing with the problem of similar documents both in length and content. In another words, the longer the document, the more words will be used for retrieval of the target documents. So the $Size_q$ is considered as a hidden variable in our document retrieval model. Besides, all the indexed documents can be filtered with $F$ formula in Equation 5-6, and it can alleviate the scarcity of tending to select longer documents when occurring the word overlap between shorter and longer documents, because a certain source document are only searched in a subset defined by its length. It can improve the precision without discard any so-called "abnormal" documents from dataset, so the $P$, $R$ and $F$ values will always be the same. Table 4-5 presents the $F$ values given by TQDL with length filter model.

Table 4-5: The F-measure of Our System with Length Filter Model

| Retrieved Documents ($N$-Best) | Query Size ($Size_q$ in %) | | | | |
|---|---|---|---|---|---|
| | **2.0** | **4.0** | **6.0** | **8.0** | **10.0** |
| 1 | 0.958 | 0.975 | 0.983 | 0.990 | 0.992 |
| 2 | 0.967 | 0.979 | 0.986 | 0.993 | 0.996 |
| 5 | 0.971 | 0.982 | 0.987 | 0.993 | 0.996 |
| 10 | 0.974 | 0.983 | 0.988 | 0.995 | 0.996 |
| 20 | 0.974 | 0.983 | 0.990 | 0.995 | 0.996 |

Compared with the results presented in Tables 4-4 and 4-5, it shows that the length filter model is able to give a high improvement by 4.5% in F-measure and achieve more than 99% of successful rate, in the case that the desired candidate is ranked in the first place. Above all, there is no documents waste in the dataset.

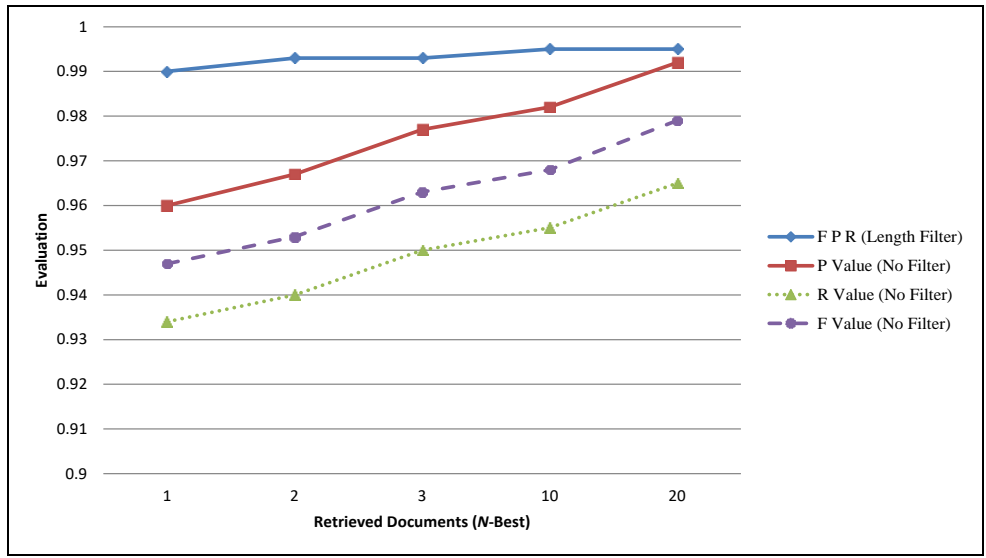

Figure 4-4: The Changes of Evaluation with N-Best

Figure 4-4 presents an ideal distribution of evaluation, of which *P* and *R* should be closed to the *F* line. In this comparison, query size $Size_q$ was still set as a constant value (8.0%). With the increasing of *N*, evaluations without filter are in a low level, while the one with this filter can achieve a good and stable performance. Finally, the precision and recall values are closed to F measure, which can all keep in a high level (99%-100%).

### 4.1.4 SECTION SUMMARY

This article presents a TQDL statistical approach for CLIR which has been explored for both large and similar documents retrieval. Different from the traditional parallel corpora-based model which relies on IBM algorithm, we divided our CLIR model into four independent parts but all work together to deal with the term disambiguation, query generation and document retrieval. The performances showed that this method can do a good job of CLIR for not only large documents but also the similar documents. This fully illustrates the discrimination power of the proposed method. It is of a great significance to both cross-language searching on the Internet and the parallel corpus producing for statistical machine translation systems. In the future work, the TQDL system will be evaluated for Chinese language, which is a big changing and more meaningful to CLIR. In the further work, we plan to make better use of the proposed models between significantly different languages such as Portuguese-Chinese. Besides, we also totally crawled 2 million Chinese-English and 500 thousand Portuguese-English sentence pairs in various domains such as news, laws, microblog etc. (Liu et al., 2018; Tian et al., 2014) as well as discourse parallel corpus (Wang et al., 2016).

## 4.2 DOMAIN ADAPTATION FOR MEDICAL TEXT TRANSLATION USING WEB RESOURCES

This section describes adapting statistical machine translation (SMT) systems to medical domain using in-domain and general-domain data as well as web-crawled in-domain resources. In order to complement the limited in-domain corpora, we apply domain focused web-crawling approaches to acquire in-domain monolingual data and bilingual lexicon from the Internet. First of all, we collect the medical terminologies from the web. This tiny but significant parallel data are helpful to reduce the out-of-vocabulary words (OOVs) in translation models. In addition, the use of larger language models during decoding is aided by more efficient storage and inference (Heafield, 2011). Thus, we crawl more in-domain monolingual data from the Internet based on domain focused web-crawling approach. In order to detect and remove out-domain data from the crawled data, we not only explore text-to-topic classifier (Pecina et al., 2011), but also propose an alternative filtering approach combined the existing one (text-to-topic classifier) with perplexity. The collected data is used for adapting the language model and translation model to boost the overall translation quality. After carefully pre-processing all the available training data, we apply language model adaptation and translation model adaptation using various kinds of training corpora. We conduct experiments on corpora of the medical summary sentence translation task of the Ninth Workshop on Statistical Machine Translation (WMT2014)[47]. Experimental results show that the presented approaches are helpful to further boost the baseline system (Lu et al., 2014).

---

[47] Six language pairs: Czech-English (cs-en), French-English (fr-en), German-English (de-en) and the reverse direction pairs (i.e., en-cs, en-fr and en-de).

### 4.2.1 PROPOSED METHODS

In this section, we introduce our domain focused web-crawling approaches on acquisition of in-domain translation terminologies and monolingual sentences.

**Bilingual Dictionary**

Terminology is a system of words used to name things in a particular discipline. The in-domain vocabulary size directly affects the performance of domain-specific SMT systems. Small size of in-domain vocabulary may result in serious OOVs problem in a translation system. Therefore, we crawl medical terminologies from some online sources such as dict.cc[48], where the vocabularies are divided into different subjects. We obtain the related bilingual entries in medicine subject by using Scala build-in XML parser and XPath. After cleaning, we collected 28,600, 37,407, and 37,600 entries in total for cs-en, de-en, and fr-en respectively.

**Monolingual Data**

The workflow for acquiring in-domain resources consists of a number of steps such as domain identification, text normalization, language identification, noise filtering, and post-processing as well as parallel sentence identification.

Firstly we use an open-source crawler, Combine[49], to crawl webpages from the Internet. In order to classify these webpages as relevant to the medical domain, we use a list of triplets *<term, relevance weight, topic class>* as the basic entries to define the topic. *Term* is a word or phrase. We select terms for each language from the following sources:

- The Wikipedia title corpus, a WMT2014 official data set consisting of titles of medical articles.
- The dict.cc dictionary.
- The DrugBank corpus, which is a WMT2014 official data set on bioinformatics and cheminformatics.

---

[48] http://www.dict.cc/.

[49] http://combine.it.lth.se/.

For the parallel data, i.e. Wikipedia and dict.cc dictionary, we separate the source and target text into individual text and use either side of them for constructing the term list for different languages. Regarding the DrugBank corpus, we directly extract the terms from the "*name*" field. The vocabulary size of collected text for each language is shown in Table 4-6.

Table 4-6: Size of Terms Used for Topic Definition

| Resources | EN | CS | DE | FR |
|---|---|---|---|---|
| Wikipedia Titles | 12,684 | 3,404 | 10,396 | 8,436 |
| dict.cc | 29,294 | 16,564 | 29,963 | 22,513 |
| DrugBank | 2,788 | | | |
| Total | 44,766 | 19,968 | 40,359 | 30,949 |

Relevance weight is the score for each occurrence of the term, which is assigned by its length, i.e., number of tokens. The *topic class* indicates the topics. In this study, we are interested in medical domain, the topic class is always marked with "MED" in our topic definition.

The topic relevance of each document is calculated[50] as follows:

$$s = \sum_{i=1}^{N}\sum_{j=1}^{4} n_{ij} w_i^I w_j^I \qquad (4\text{-}7)$$

where $N$ is the amount of terms in the topic definition; $w_i^I$ is the weight of term $i$; $w_j^I$ is the weight of term at location $j$. $n_{ij}$ is the number of occurrences of term $i$ at $j$ position. In implementation, we use the default values for setting and parameters. Another input required by the crawler is a list of seed URLs, which are web sites that related to medical topic. We limit the crawler from getting the pages within the http domain guided by the seed links. We acquired the list from the Open Directory Project[51], which is a repository maintained by volunteer editors. Totally, we collected 12,849 URLs from the medicine category.

---

[50] http://combine.it.lth.se/documentation/DocMain/node6.html.

[51] http://www.dmoz.org/Health/Medicine/.

Text normalization is to convert the text of each HTML page into UTF-8 encoding according to the content_charset of the header. In addition, HTML pages often consist of a number of irrelevant contents such as the navigation links, advertisements disclaimers, etc., which may negatively affect the performance of SMT system. Therefore, we use the Boilerpipe tool (Kohlschütter et al., 2010) to filter these noisy data and preserve the useful content that is marked by the tag, <canonicalDocument>. The resulting text is saved in an XML file, which will be further processed by the subsequent tasks. For language identification, we use the language-detection[52] toolkit to determine the possible language of the text, and discard the articles which are in the right language we are interested.

**Data Filtering**

The web-crawled documents may consist of a number of out-domain data, which would harm the domain-specific language and translation models. We explore and propose two filtering approaches for this task. The first one is to filter the documents based on their relative score, Eq. (4-7). We rank all the documents according to their relative scores and select top *K* percentage of entire collection for further processing.

Second, we use a combination method, which takes both the perplexity and relative score into account for the selection. Perplexity-based data selection has shown to be a powerful mean on SMT domain adaptation (Wang et al., 2013; Wang et al., 2014; Toral, 2013; Rubino et al., 2013; Duh et al., 2013). The combination method is carried out as follows: we first retrieve the documents based on their relative scores. The documents are then split into sentences, and ranked according to their perplexity using Eq. (4-8) (Stolcke et al., 2002). The used language model is trained on the official in-domain data. Finally, top *N* percentage of ranked sentences are considered as additional relevant in-domain data.

$$ppl1(s) = 10^{-\log\frac{P(T)}{Word}}$$

(4-8)

---

where $s$ is a input sentence or document, $P(T)$ is the probability of $n$-gram segments estimated from the training set. *Word* is the number of tokens of an input string.

### 4.2.2 EXPERIMENTS AND RESULTS

The official medical summary development sets (dev) are used for tuning and evaluating the comparative systems. The official medical summary test sets (test) are only used in our final submitted systems.

The experiments were carried out with the Moses 1.0[53] (Koehn et al., 2007). The translation and the re-ordering model utilizes the "*grow-diag-final*" symmetrized word-to-word alignments created with MGIZA++[54] (Och and Ney, 2003; Gao and Vogel, 2008) and the training scripts from Moses. A 5-gram LM was trained using the SRILM toolkit[55] (Stolcke et al., 2002), exploiting improved modified Kneser-Ney smoothing, and quantizing both probabilities and back-off weights. For the log-linear model training, we take the minimum-error-rate training (MERT) method as described in (Och, 2003).

In the following sub-sections, we describe the results of **baseline systems**, which are trained on the official corpora. We also present the **enhanced systems** that make use of the web-crawled bilingual dictionary and monolingual data as the additional training resources. Two variants of enhanced system are constructed based on different filtering criteria.

**Baseline System**

The baseline system is constructed based on the combination of TM adaptation and LM adaptation, where the corresponding selection thresholds ($M$) are manually tuned. Table 4-7 shows the BLEU scores of baseline systems as well as the threshold values of $M$ for general-domain monolingual corpora and parallel corpora selection, respectively.

---

[53] http://www.statmt.org/moses/.

[54] http://www.kyloo.net/software/doku.php/mgiza:overview.

[55] http://www.speech.sri.com/projects/srilm/.

Table 4-7: BLEU Scores of Baseline Systems for Different Language Pairs

| Lang. Pair | BLEU |
|:---:|:---:|
| **en-cs** | 17.57 |
| **cs-en** | 31.29 |
| **en-fr** | 38.36 |
| **fr-en** | 44.36 |
| **en-de** | 18.01 |
| **de-en** | 32.50 |

By looking into the results, we find that en-cs system performs poorly, because of the limited in-domain parallel and monolingual corpora (shown in Table 4-7). While the fr-en and en-fr systems achieve the best scores, due the availability of the high volume training data. We experiment with different values of $M=\{0, 25, 50, 75, 100\}$ that indicates the percentages of sentences out of the general corpus used for constructing the LM adaptation and TM adaptation. After tuning the parameter $M$, we find that BLEU scores of different systems peak at different values of $M$. LM adaptation can achieve the best translation results for cs-en, en-fr and de-en pairs when $M=25$, en-cs and en-de pairs when $M=50$, and fr-en pair when $M=75$. While TM adaptation yields the best scores for en-fr and en-de pairs at $M=25$ and cs-en and fr-en pairs at $M=50$, de-en pair when $M=75$ and en-cs pair at $M=100$.

**Based on Relevance Score Filtering**

We use the relevance score to filter out the non-in-domain documents. Once again, we evaluate different values of $K=\{0, 25, 50, 75, 100\}$ that represents the percentages of crawled documents we used for training the LMs. In Table 4-8, we show the absolute BLEU scores of the evaluated systems, listed with the optimized thresholds, and the relative improvements ($\Delta\%$) in compared to the baseline system. The size of additional training data (for LM) is displayed at the last column. The relevance score filtering approach yields an improvement of 3.08% of BLEU score for de-en pair that is the best result among the language pairs. On the other hand, en-cs pair obtains a marginal gain. The reason is very obvious that the training data is very insufficient. Empirical results of all language pairs expect fr-en indicate that data filtering is the necessity to improve the system performance.

Table 4-8: Evaluation Results for Systems That Trained on Relevance-Score-Filtered Documents

| Lang. Pair | Docs ($K$%) | BLEU | Δ (%) | Sent. |
|:---:|:---:|:---:|:---:|---:|
| **en-cs** | 50 | 17.59 | 0.11 | 31,065 |
| **en-de** | 75 | 18.52 | 2.83 | 435,547 |
| **en-fr** | 50 | 39.08 | 1.88 | 743,735 |
| **cs-en** | 75 | 32.22 | 2.97 | 7,943,931 |
| **de-en** | 25 | 33.50 | 3.08 | 4,951,189 |
| **fr-en** | 100 | 45.45 | 2.46 | 8,448,566 |

**Based on Moore-Lewis Filtering**

In this approach, we need to determine the values of two parameters, top $K$ documents and top $N$ sentences, where $K$={100, 75, 50} and $N$={75, 50, 25}, $N$<$K$. When $K$=100, it is a conventional perplexity-based data selection method, i.e. no document will be filtered. Table 4-9 shows the combination of different $K$ and $N$ that gives the best translation score for each language pair. We provide the absolute BLEU for each system, together with relative improvements (Δ%) that compared to the baseline system.

Table 4-9: Evaluation Results for Systems That Trained on Combination Filtering Approach

| Lang. Pair | Docs ($K$%) | Target Size ($N$%) | BLEU | Δ (%) |
|:---:|:---:|:---:|:---:|:---:|
| **en-cs** | 50 | 25 | 17.69 | 0.68 |
| **en-de** | 100 | 50 | 18.03 | 0.11 |
| **en-fr** | 100 | 50 | 38.73 | 0.96 |
| **cs-en** | 100 | 25 | 32.20 | 2.91 |
| **de-en** | 100 | 25 | 33.10 | 1.85 |
| **fr-en** | 100 | 25 | 45.22 | 1.94 |

In this shared task, we have a quality and quantity in-domain monolingual training data for English. All the systems that take English as the target translation always outperform the other reverse pairs. Besides, we found the systems based on the perplexity data selection method tend to achieve a better score in BLEU.

### 4.2.3 SECTION SUMMARY

We described our study on developing unconstrained systems in the medical translation task of 2014 Workshop on Statistical Machine Translation. In this work, we adopt the web crawling strategy for acquiring the in-domain monolingual data. In detection the domain data, we exploited Moore-Lewis data selection method to filter the collected data in addition to the build-in scoring model provided by the crawler toolkit. However, after investigation, we found that the two methods are very competitive to each other.

The systems we submitted to the shared task were built using the language models and translation models that yield the best results in the individual testing. The official test set is converted into the *recased* and *detokenized* SGML format. Table 4-10 presents the official results of our submissions for every language pair.

Table 4-10: BLEU Scores of The Submitted Systems for The Medical Translation Task in Six Language Pairs

| Lang. Pair | BLEU of Combined systems | Official BLEU |
|:---:|:---:|:---:|
| en-cs | 23.16 (+5.59) | 22.10 |
| cs-en | 36.8 (+5.51) | 37.40 |
| en-fr | 40.34 (+1.98) | 40.80 |
| fr-en | 45.79 (+1.43) | 43.80 |
| en-de | 19.36 (+1.35) | 18.80 |
| de-en | 34.17 (+1.67) | 32.70 |

CHAPTER 5: DOMAIN-SPECIFIC SMT ONLINE SYSTEM

To test our approaches in a real-life environment, we also develop a domain-specific SMT on-line system named **BenTu**. BenTu is an on-line SMT system for translating medical, technological domain texts, supporting English, Chinese, French, German and Czech languages. In the following sections, we take medical domain for instance to describe MT component in BenTu. We explore a number of simple and effective techniques to adapt statistical machine translation (SMT) systems in the medical domain.

## 5.1 MEDICAL DOMAIN ADAPTATION EXPERIMENTS

By comparing the medical text with common text, we discovered some interesting phenomena in medical genre. We apply domain-specific techniques in data pre-processing, language model adaptation, translation model adaptation, numeric and hyphenated words translation. Compared to the baseline systems, the results of each method show reasonable gains. We combine individual approach to further improve the performance of our systems. To validate the robustness and language-independency of individual and combined systems, we conduct experiments on the official training data in all six language pairs. We anticipate the numeric comparison (BLEU scores) on these individual and combined domain adaptation approaches that could be valuable for others on building a real-life domain-specific system (Wang et al., 2014).

### 5.1.1 EXPERIMENTAL SETUP

All available training data from both WMT2014 standard translation task [56] (general-domain data) and medical translation task[57] (in-domain data) are used in this study. The official medical summary development sets (dev) are used for tuning and evaluating all the comparative systems. The official medical summary test sets (test) are only used in our final submitted systems.

---

[56] http://www.statmt.org/wmt14/translation-task.html.

[57] http://www.statmt.org/wmt14/medical-task/.

The experiments were carried out with the Moses 1.0[58] (Koehn et al., 2007). The translation and the re-ordering model utilizes the "*grow-diag-final*" symmetrized word-to-word alignments created with MGIZA++[59] (Och and Ney, 2003; Gao and Vogel, 2008) and the training scripts from Moses. A 5-gram LM was trained using the SRILM toolkit[60] (Stolcke et al., 2002), exploiting improved modified Kneser-Ney smoothing, and quantizing both probabilities and back-off weights. For the log-linear model training, we take the minimum-error-rate training (MERT) method as described in (Och, 2003).

### 5.1.2 Task Oriented Pre-processing

A careful pre-processing on training data is significant for building a real-life SMT system. In addition to the general data preparing steps used for constructing the baseline system, we introduce some extra steps to pre-process the training data.

The first step is to remove the duplicate sentences. In data-driven methods, the more frequent a term occurs, the higher probability it biases. Duplicate data may lead to unpredicted behavior during the decoding. Therefore, we keep only the distinct sentences in monolingual corpus. By taking into account multiple translations in parallel corpus, we remove the duplicate sentence pairs. The second concern in pre-processing is symbol normalization. Due to the nature of medical genre, symbols such as numbers and punctuations are commonly-used to present chemical formula, measuring unit, terminology and expression. Figure 1 shows the examples of this case. These symbols are more frequent in medical article than that in the common texts. Besides, the punctuations of *apostrophe* and *single quotation* are interchangeably used in French text, e.g. "*l'effet de l'inhibition*". We unify it by replacing with the *apostrophe*. In addition, we observe that some monolingual training subsets (e.g., Gene Regulation Event Corpus) contain sentences of more than 3,000 words in length. To avoid the long sentences from harming the truecase model, we split them into sentences with a sentence splitter[61] (Rune et al., 2007) that is optimized for biomedical texts. On the other hand, we consider the target system is intended for summary translation, the sentences tend to be short in length. For instance, the average sentence lengths in development sets of cs, fr, de and en are around 15, 21, 17 and 18, respectively. We remove sentence pairs which are more than 80 words at length. In order to that our experiments are reproducible, we give the detailed statistics of task oriented pre-processed training data in Table 5-1.

---

[58] http://www.statmt.org/moses/.

[59] http://www.kyloo.net/software/doku.php/mgiza:overview.

[60] http://www.speech.sri.com/projects/srilm/.

| |
|---|
| **1,25**-OH<br>**47** to **80%**<br>**10-20 ml/kg**<br>A**and**E department<br>Infective endocarditis (IE) |

Figure 5-1: Examples of The Segments With Symbols in Medical Texts

To validate the effectiveness of the pre-processing, we compare the SMT systems trained on original data[62](*Baseline*1) and task-oriented-processed data (*Baseline*2), respectively. Table 5-1 shows the results of the baseline systems. We found all the *Baseline*2 systems outperform the *Baseline*1 models, showing that the systems can benefit from using the processed data. For cs-en and en-cs pairs, the BLEU scores improve quite a lot. For other language pairs, the translation quality improves slightly.

Table 5-1: BLEU Scores of Two Baseline Systems Trained On Original and Processed Corpora for Different Language Pairs

| Lang. Pair | Baseline1 | Baseline2 | Diff. |
|:---:|:---:|:---:|:---:|
| **en-cs** | 12.92 | **17.57** | +4.65 |
| **cs-en** | 20.85 | **31.29** | +10.44 |
| **en-fr** | 38.31 | **38.36** | +0.05 |
| **fr-en** | 44.27 | **44.36** | +0.09 |
| **en-de** | 17.81 | **18.01** | +0.20 |
| **de-en** | 32.34 | **32.50** | +0.16 |

---

[61]  http://www.nactem.ac.uk/y-matsu/geniass/.

[62]  Data are processed according to Moses baseline tutorial: http://www.statmt.org/moses/?n=Moses.Baseline.

By analyzing the *Baseline*2 results (in Table 5-1) and the statistics of training corpora (in Table 5-2), we can further elaborate and explain the results. The en-cs system performs poorly, because of the short average length of training sentences, as well as the limited size of in-domain parallel and monolingual corpora. On the other hand, the fr-en system achieves the best translation score, as we have sufficient training data. The translation quality of cs-en, en-fr, fr-en and de-en pairs is much higher than those in the other pairs. Hence, *Baseline*2 will be used in the subsequent comparisons with the proposed systems described in Section 4, 5, 6 and 7.

Table 5-2: Statistics Summary of Corpora after Pre-Processing

| Data Set | Lang. | Sent. | Words | Vocab. | Ave. Len. |
|---|---|---|---|---|---|
| In-domain Parallel Data | cs/en | **1,770,421** | 9,373,482/ 10,605,222 | 134,998/ 156,402 | **5.29/** **5.99** |
| | de/en | 3,894,099 | 52,211,730/ 58,544,608 | 1,146,262/ 487,850 | 13.41/ 15.03 |
| | fr/en | **4,579,533** | 77,866,237/ 68,429,649 | 495,856/ 556,587 | 17.00/ 14.94 |
| General-domain Parallel Data | cs/en | 12,426,374 | 180,349,215/ 183,841,805 | 1,614,023/ 1,661,830 | 14.51/ 14.79 |
| | de/en | 4,421,961 | 106,001,775/ 112,294,414 | 1,912,953/ 919,046 | 23.97/ 25.39 |
| | fr/en | 36,342,530 | 1,131,027,766/ 953,644,980 | 3,149,336/ 3,324,481 | 31.12/ 26.24 |
| In-domain Mono. Data | cs | **106,548** | 1,779,677 | 150,672 | 16.70 |
| | fr | 1,424,539 | 53,839,928 | 644,484 | 37.79 |
| | de | 2,222,502 | 53,840,304 | 1,415,202 | 24.23 |
| | en | **7,802,610** | 199430649 | 1,709,594 | 25.56 |
| General-domain Mono. Data | cs | 33,408,340 | 567,174,266 | 3,431,946 | 16.98 |
| | fr | 30,850,165 | 780,965,861 | 2,142,470 | 25.31 |
| | de | 84,633,641 | 1,548,187,668 | 10,726,992 | 18.29 |
| | en | 85,254,788 | 2,033,096,800 | 4,488,816 | 23.85 |

### 5.1.3 LANGUAGE MODEL ADAPTATION

The use of LMs (trained on large data) during decoding is aided by more efficient storage and inference (Heafield, 2011). Therefore, we not only use the in-domain training data, but also the selected pseudo in-domain data[63] from general-domain corpus to enhance the LMs (Toral, 2013; Rubino et al., 2013; Duh et al., 2013). Firstly, each sentence $s$ in general-domain monolingual corpus is scored using the cross-entropy difference method in (Moore and Lewis, 2010), which is calculated as Equation 3-7. Then top $N$ percentages of ranked data sentences are selected as a pseudo in-domain subset to train an additional LM. Finally, we linearly interpolate the additional LM with in-domain LM.

---

[63] Axelrod et al. (2011) names the selected data as *pseudo in-domain data*. We adopt both terminologies in this paper.

We use the top $N$% of ranked results, where $N=\{0, 25, 50, 75, 100\}$ percentages of sentences out of the general corpus. Table 5-3 shows the absolute BLEU points for *Baseline*2 ($N=0$), while the LM adapted systems are listed with values relative to the *Baseline*2. The results indicate that LM adaptation can gain a reasonable improvement if the LMs are trained on more relevant data for each pair, instead of using the whole training data. For different systems, their BLEU scores peak at different values of $N$. It gives the best results for cs-en, en-fr and de-en pairs when $N=25$, en-cs and en-de pairs when $N=50$, and fr-en pair when $N=75$. Among them, en-cs and en-fr achieve the highest BLEU scores. The reason is that their original monolingual (in-domain) data for training the LMs are not sufficient. When introducing the extra pseudo in-domain data, the systems improve the translation quality by around 2 BLEU points. While for cs-en, fr-en and de-en pairs, the gains are small. However, it can still achieve a significant improvement of 0.60 up to 1.12 BLEU points.

Table 5-3: BLEU Scores of LM Adapted Systems

| Lang. | *N*=0 | *N*=25 | *N*=50 | *N*=75 | *N*=100 |
|-------|-------|--------|--------|--------|---------|
| en-cs | 17.57 | +1.66 | **+2.08** | +1.72 | +2.04 |
| cs-en | 31.29 | **+0.94** | +0.60 | +0.66 | +0.47 |
| en-fr | 38.36 | **+1.82** | +1.66 | +1.60 | +0.08 |
| fr-en | 44.36 | +0.91 | +1.09 | **+1.12** | +0.92 |
| en-de | 18.01 | +0.57 | **+1.02** | -4.48 | -4.54 |
| de-en | 32.50 | **+0.60** | +0.50 | +0.56 | +0.38 |

### 5.1.4 Translation Model Adaptation

As shown in Table 5-2, general-domain parallel corpora are around 1 to 7 times larger than the in-domain ones. We suspect if general-domain corpus is broad enough to cover some in-domain sentences. To observe the domain-specificity of general-domain corpus, we firstly evaluate systems trained on general-domain corpora. In Table 5-4, we show the BLEU scores of general-domain systems[64] on translating the medical sentences. The BLEU scores of the compared systems are relative to the *Baseline*2 and the size of the used general-domain corpus is relative to the corresponding in-domain one. For en-cs, cs-en, en-fr and fr-en pairs, the general-domain parallel corpora we used are 6 times larger than the original ones and we obtain the improved BLEU scores by 1.72 up to 3.96 points. While for en-de and de-en pairs, the performance drops sharply due to the limited training corpus we used. Hence we can draw a conclusion: the general-domain corpus is able to aid the domain-specific translation task if the general-domain data is large and broad enough in content.

Table 5-4: The BLEU Scores of Systems Trained on General-Domain Corpora

| Lang. Pair | BLEU | Diff. | Corpus |
|------------|------|-------|--------|
| **en-cs** | 21.53 | +3.96 | +601.89% |
| **cs-en** | 33.01 | +1.72 | |
| **en-fr** | 41.57 | +3.21 | +693.59% |
| **fr-en** | 47.33 | +2.97 | |
| **en-de** | 16.54 | -1.47 | +13.63% |
| **de-en** | 27.35 | -5.15 | |

---

[64] General-domain systems are trained only on genera-domain training corpora (i.e., parallel, monolingual).

Taking into account the performance of general-domain system, we explore various data selection methods to derive the pseudo in-domain sentence pairs from general-domain parallel corpus for enhancing the TMs (Wang et al., 2013; Wang et al., 2014). Firstly, sentence pair in corresponding general-domain corpora is scored by the modified Moore-Lewis (Axelrod et al., 2011), which is calculated as Equation 3-8. Then top $N$ percentage of ranked sentence pairs are selected as a pseudo in-domain subset to train an individual translation model. The additional model is log-linearly interpolated with the in-domain model (*Baseline*2) using the multi-decoding method described in (Koehn and Schroeder, 2007).

Similar to LM adaptation, we use the top $N$% of ranked results, where $N$={0, 25, 50, 75, 100} percentages of sentences out of the general corpus. Table 5-5 shows the absolute BLEU points for *Baseline*2 ($N$=0), while for the TM adapted systems we show the values relative to the *Baseline*2. For different systems, their BLEU peak at different $N$. For en-fr and en-de pairs, it gives the best translation results at $N$=25. Regarding cs-en and fr-en pairs, the optimal performance is peaked at $N$=50. While the best results for de-en and en-cs pairs are $N$=75 and $N$=100 respectively. Besides, performance of TM adapted system heavily depends on the size and (domain) broadness of the general-domain data. For example, the improvements of en-de and de-en systems are slight due to the small general-domain corpora. While the quality of other systems improve about 3 BLEU points, because of their large and broad general-domain corpora.

Table 5-5: BLEU Scores of TM Adapted Systems

| Lang. | $N$=0 | $N$=25 | $N$=50 | $N$=75 | $N$=100 |
|-------|-------|--------|--------|--------|---------|
| en-cs | 17.57 | +0.84 | +1.53 | +1.74 | **+2.55** |
| cs-en | 31.29 | +2.03 | **+3.12** | +3.12 | +2.24 |
| en-fr | 38.36 | **+3.87** | +3.66 | +3.53 | +2.88 |
| fr-en | 44.36 | +1.29 | **+3.36** | +1.84 | +1.65 |
| en-de | 18.01 | **+0.02** | -0.13 | -0.07 | 0 |
| de-en | 32.50 | -0.12 | +0.06 | **+0.31** | +0.24 |

### 5.1.5 NUMERIC ADAPTATION

The *numeric* occurs frequently in medical texts. However, numeric expression in dates, time, measuring unit, chemical formula are often sparse, which may lead to OOV problems in phrasal translation and reordering. Replacing the sparse numbers with placeholders may produce more reliable statistics for the MT models.

Moses has support using placeholders in training and decoding. Firstly, we replace all the numbers in monolingual and parallel training corpus with a common symbol (a sample phrase is illustrated in Figure 5-2). Models are then trained on these processed data. We use the XML markup translation method for decoding.

| Original: | Vitamin D 1,25-OH |
|---|---|
| Replaced: | Vitamin D @num@, @num@-OH |

Figure 5-2: Examples of Placeholders

Table 5-6 shows the results on this number adaptation approach as well as the improvements compared to the *Baseline*2. The method improves the *Baseline*2 systems by 0.23 to 0.40 BLEU scores. Although the scores increase slightly, we still believe this adaptation method is significant for medical domain. The WMT2014 medical task only focuses on the summary of medical text, which may contain fewer chemical expression in compared with the full article. As the used of numerical instances increases, placeholder may play a more important role in domain adaptation.

Table 5-6: BLEU Scores of Numeric Adapted Systems

| Lang. Pair | BLEU (Dev) | Diff. |
|---|---|---|
| en-cs | 17.80 | +0.23 |
| cs-en | 31.52 | +0.23 |
| en-fr | 38.72 | +0.36 |
| fr-en | 44.69 | +0.33 |
| en-de | 18.41 | +0.40 |
| de-en | 32.88 | +0.38 |

### 5.1.6 HYPHENATED WORD ADAPTATION

Medical texts prefer a kind of compound words, hyphenated words, which is composed of more than one word. For instance, "*slow-growing*" and "*easy-to-use*" are composed of words and linked with hyphens. These hyphenated words occur quite frequently in medical texts. We analyze the development sets of cs, fr, en and de respectively, and observe that there are approximately 3.2%, 11.6%, 12.4% and 19.2% of sentences that contain one or more hyphenated words. The high ratio of such compound words results in Out-Of-Vocabulary words (OOV)[65], and harms the phrasal translation and reordering. However, a number of those hyphenated words still have chance to be translated, although it is not precisely, when they are tokenized into individual words. To resolve this problem, we present an *alternative-translation* method in decoding. Table 5-7 shows the proposed algorithm.

Table 5-7: Alternative-Translation Algorithm

| |
|---|
| **Algorithm:** Alternative-translation Method |
| **Input:** |
| 5.   A sentence, *s*, with *M* hyphenated words |
| 6.   Translation lexicon |
| **Run:** |
|    1.   **For** $i = 1, 2, …, M$ |
|    2.      Split the $i$th hyphenated word ($C_i$) into $P_i$ |
|    3.      Translate $P_i$ into $T_i$ |
|    4.   **If** ($T_i$ are not OOVs): |
|    5.      Put alternative translation $T_i$ in XML |
|    6.   **Else**: keep $C_i$ unchanged |
| **Output:** |
|    Sentence, *s'*, embedded with alternative translations for all $T_i$. |
| **End** |

In the implementation, we apply XML markup to record the translation (terminology) for each compound word. During the decoding, a hyphenated word delimited with markup will be replaced with its corresponding translation. Table 6-8 shows the BLEU scores of adapted systems applied to hyphenated translation. This method is effective for most language pairs. While the translation systems for en-cs and cs-en do not benefit from this adaptation, because the hyphenated words ratio in the en and cs dev are asymmetric. Thus, we only apply this method for en-fr, fr-en, de-en and en-de pairs.

---

[65] Default tokenizer does not handle the hyphenated words.

Table 5-8: BLEU Scores of Hyphenated Word Adapted Systems

| Lang. Pair | BLEU (Dev) | Diff. |
|---|---|---|
| en-cs | 16.84 | -0.73 |
| cs-en | 31.23 | -0.06 |
| en-fr | 39.12 | +0.76 |
| fr-en | 45.02 | +0.66 |
| en-de | 18.64 | +0.63 |
| de-en | 33.01 | +0.51 |

5.1.7 FINAL RESULTS AND CONCLUSIONS

According to the performance of each individual domain adaptation approach, we combined the corresponding models for each language pair. In Table 5-9, we show the BLEU scores and its increments (compared to the *Baseline*2) of combined systems in the second column. The official test set is converted into the *recased* and *detokenized* SGML format. The official results of our submissions are given in the last column of Table 5-9.

Table 5-9: BLEU Scores of the Submitted Systems for the Medical Translation Task

| Lang. Pair | BLEU of combined systems | Official BLEU |
|---|---|---|
| en-cs | 23.66 (+6.09) | 22.60 |
| cs-en | 38.05 (+6.76) | 37.60 |
| en-fr | 42.30 (+3.94) | 41.20 |
| fr-en | 48.25 (+3.89) | 47.10 |
| en-de | 21.14 (+3.13) | 20.90 |
| de-en | 36.03 (+3.53) | 35.70 |

This section presents a set of experiments conducted on all available training data for six language pairs. We explored various domain adaptation approaches for adapting medical translation systems. Compared with other methods, language model adaptation and translation model adaptation are more effective. Other adapted techniques are still necessary and important for building a real-life system. Although all individual methods are not fully additive, combining them together can further boost the performance of the overall domain-specific system. We believe these empirical approaches could be valuable for SMT development.

## 5.2 BENTU: DOMAIN-SPECIFIC SMT SYSTEM

We develop our first online translator, BenTu, which is a domain-specific multi-tire SMT system. The architecture is designed referring to PluTO project (Ceauşu et al., 2011), which addresses the flexibility of adapting to new language pairs and exploring new processing techniques, as language-specific components can be plugged in at various stages in the translation pipeline.

Our system is deployed at three levels: 1, main access point for translation; 2, translation server; 3, worker/decoder server. Communication to and between each of these levels is carried on using XML-RPC conformant messages.

The main access point for document translation offers synchronous communication to the MT server through a URL that contains the translation direction. It takes as input an JSON format as shown in Figure 5-3.

```
{domain:Technology, source:EN, target:ZH-CN,…}
    domain: "Technology"
    sentence: "we properly implement a sound automation practice.'
    source: "EN"
    target: "ZH-CN"
```

Figure 5-3: Input Format

The output is the translation of the document in the desired language. The translated document might optionally contain alignment information between source and target at both sentence and token level as shown in Figure 5-4.

```
{
    "translation":"我们 正确 地 执行 健全 的 自动化 的 做法 。 ",
    "alignment":[
        {
            "0":"0-50-0-27@0-2-0-2 3-21-3-10 22-29-11-15 30-40-16-19 41-50-20-26 @0-2-0-2 3-11-3-5 12-21-6-7 41-49-20-21"
        }
    ],
    "unks":[
        {
            "0":"0-50-0-27@0-2-0-2 3-21-3-10 22-29-11-15 30-40-16-19 41-50-20-26 @0-2-0-2 3-11-3-5 12-21-6-7 41-49-20-21"
        }
    ],
    "keywords":[
    ],
    "optionalTranslations":[
    ]
}
```

Figure 5-4: Alignment Information

The main access point for patent document translation transforms each document in a job for the XML-RPC translation servers. Each para-graph from the documents is sent as an asynchronous translation request to the server registered for the given translation direction. There are several XML-RPC methods that provide the asynchronous characteristic of the request:

- `submit_translation` sends a portion of text to be translated (usually a para-graph)
- `request_translation` returns the translation if it is ready or an estimated number of milliseconds to wait for the translation
- `request_alignment` returns the alignment information if the translation is ready or an estimated number of milli-seconds to wait for.

In order to return translations as quickly as possible, the translation server has to distribute translation tasks across several cores/machines.

The MT system diagram in Figure 5-5 shows how the system carries on translating multiple sentences simultaneously. The server is based on the multiple producers/consumers pat-tern. It has a task mapper in which, from a given input text, separate tasks are produced. In our case, the task mapper splits the input into several sentences. There can be one or more workers that pre-process, translate and post-process the translation. The task collector reorders the tasks and delivers the final translation. In-between the task mapper, the workers and the task collector, there are blocking task queues. These queues have prioritization allowing the system to provide a fair-scheduling mechanism for the documents to be translated. That means that each job (document) submitted to the translation server get approximately the same share of the server resources over time. A short document won't have to wait for the completion of a larger document – the sentences from the small document have a higher priority in the workers queue. The workers queue is also capacity-constrained allowing the system to degrade-gracefully. That means that the sys-tem won't take more jobs that it can handle in a given time-frame.

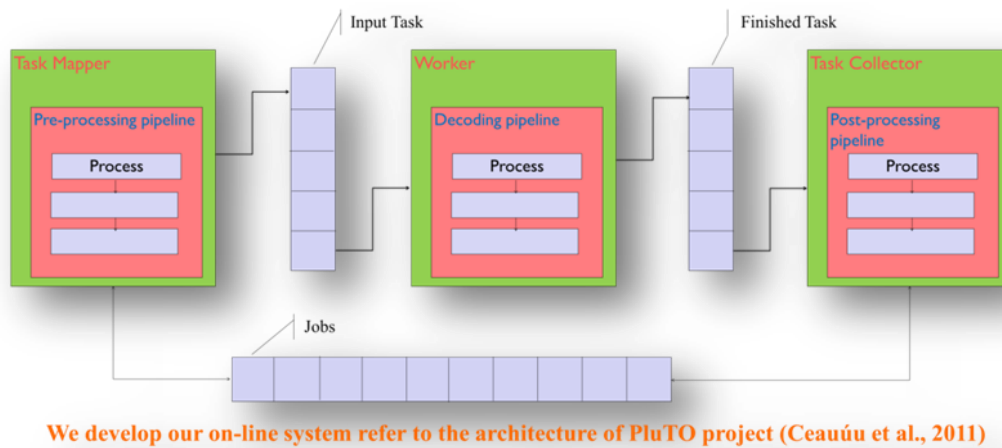

Figure 5-5: The Architecture of BenTu

All of the server modules are fully configurable through standardized XML files. The same pipelined architecture is shared among workers, task mapper and collector. In this scenario, a pipeline might consist of several processors, with each having serialized initialization and processing functions. The user interface is shown in Figure 5-6.

We also developed semantics-enhanced dialogue translation system for hotel booking (Wang et al., 2017).

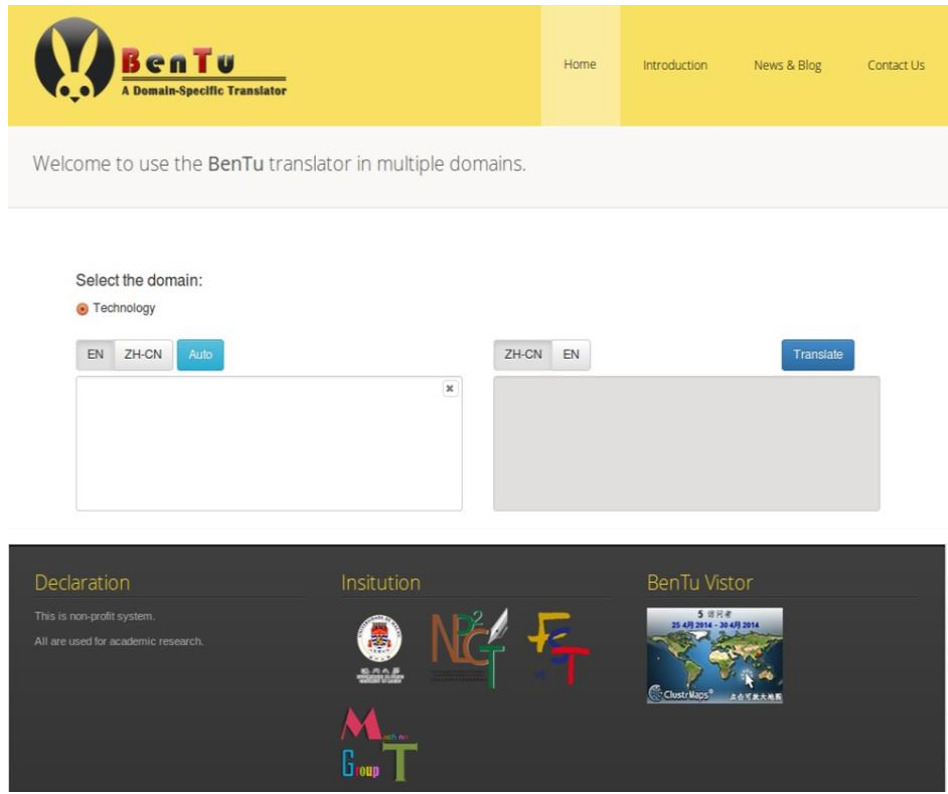

Figure 5-6: The User Interface of BenTu

# CHAPTER 6:  CONCLUSION AND FUTURE WORK

## 6.1 SUMMARY AND RESEARCH CONTRIBUTIONS

The central emphasize of our research is to explore domain adaptation approaches and techniques for improving the translation quality of domain-specific SMT systems. To address three main problems: ambiguity, language style and out-of-vocabulary words. We explore the state-of-the-art domain adaptation approaches and propose effective solutions.

Firstly, we explore intelligent data selection approaches to optimize models by selecting relevant data from general-domain corpora. As fine-grained selection model has higher ability of filtering out irrelevant data, we propose a string-difference metric as a new selection criterion. Based on this, we further explore two different approaches, at data level and model level, to combine different type of individual sources to optimize the targeted SMT models. Besides, we deeply analyze their impacts on domain-specific translation quality. We anticipate these approaches can address the ambiguity problem by transferring the data distribution of training corpora to target domain.

In order to make the models be aware the language style of sentences, we propose linguistically-augmented data selection approach to enhance perplexity-based models. This method considers various linguistic information, such as part-of-speech (POS), named entity, and so forth, instead of the surface forms. Additionally, we present two methods to combine the different types of linguistic knowledge.

In order to reduce the OOVs, we acquire additional resources to supplement in-domain training data. We apply domain-focused web-crawling methods to obtain in-domain monolingual and parallel sentences from the Internet. To further reduce the irrelevant data, we explore two domain filtering methods for this task. As crawled corpora are usually comparable, we also present an approach to improve the quality of cross-language document alignment.

To prove the robustness and language-independence of our presented methods, all the experiments were conducted on large and multi-lingual corpora. The results show a significant improvement by employing these approaches for SMT domain adaptation. Finally, we develop a domain-specific on-line SMT system named BenTu, which integrates with many useful natural language processing (NLP) toolkits and pipeline the pre-processing, hypotheses decoding and post-processing with an effective multi-tier framework.

To achieve these objectives, we explore domain adaptation approaches by conducting a lot of experiments. The main contributions of this thesis can be summarized as follows:

- Currently, there is no a uniform definition on "what is the domain?" in NLP or SMT research communities. We try to give definitions by analyzing the linguistic phenomena of corpora in different domains.

- For data selection in SMT domain adaptation, we firstly propose edit distance as a new selection criterion for this task. We systematically compare it against the state-of-the-art data selection methods. Based on the comparison, we present $i$CPE – hybrid data selection models – to combine individual models at both sentence level and model level.

- We further utilize linguistic information to improve the perplexity-based data selection for both language model adaptation and translation model adaptation. This is a novel idea compared with previous work which only considers the surface forms of words. Regarding the data selection as the task of domain labeling, we further present a graphical model to softly propagate the domain-information from labeled data to their neighboring unlabeled data.

- In order to reduce the in-domain OOVs, we firstly apply domain-focused crawling methods to mine useful resources such as monolingual corpus, parallel corpus and bilingual dictionary from the largest multilingual resource - the Internet. A combination method named TQDL to improve the cross-language document alignment performance is proposed. Besides, filtering methods are presented to reduce the "noisy" data from crawled monolingual corpora. Three kinds of bilingual dictionaries are acquired and used to improve the quality of sentence alignment. Finally, all processed crawled data are used for adapting SMT systems to a specific domain.

- A domain-specific SMT on-line system named BenTu is implemented. It covers various domains such as medicine, technology, news, and others for a number of languages. To well integrate different NLP toolkits into the system, we apply a multi-tier framework to combine the pre-processing, translation and post-processing steps.

Besides, we also explored other related CLIR and NLP tasks: cross-language document retrieval (Wang et al., 2013; Wang et al., 2012; Wang et al., 2012), Chinese word segmentation (Wang et al., 2012), named entity (Wang et al., 2012), grammar error correction (Wang et al., 2014; Xing et al., 2013), discourse parsing (Okita et al. 2015, Wang et al., 2015).

## 6.2 FUTURE WORKS

The research work in this thesis investigates and improves the performance of domain-focused machine translation system in both experiments and real-life environment. Although a number of proposed methods achieve impressive results, there are still many rooms to further improve them.

Firstly, we will explore graphical models on data selection. Regarding the problem of data selection as domain labeling, we build a graph of labeled and unlabeled data, and then apply label propagation algorithm to classify unlabeled data. We believe this model is able to capture useful from the global point of view. On the other hand, the estimated distributions of data among different domains can be used as a soft constraint to the construction of in-domain data selection algorithms.

Besides, we will explore on data selection using neural language models. We hypothesize that the continuous vector representation of words in neural language models makes them more effective than n-grams for modeling the word contexts in a more abstract way, the semantic. This can prevent the selection methods suffering from the obstacle caused by unknown words.

In the future, I will mainly focus on explore neural methods (Lu et al., 2014) and neural machine translation (Wang et al., 2016; Wang et al., 2016; Wang et al., 2017; Wang et al., 2017; Wang et al., 2018).

VITA

Longyue Wang

University of Macau

2014

**Academic Qualification**

B.Sc. Degree in Network Engineering, Shandong University of Science and Technology, Tsingtao, China, 2011

**Publications**